\documentclass[11pt]{article}

\usepackage[preprint]{acl}

\usepackage{times}
\usepackage{latexsym}

\usepackage{amsmath}
\usepackage{amssymb}
\usepackage{booktabs}
\usepackage{adjustbox}

\makeatletter
\g@addto@macro\UrlBreaks{\do\-}
\makeatother
\newcommand{\model}[1]{\path{#1}}

\usepackage[T1]{fontenc}

\usepackage[utf8]{inputenc}

\usepackage{microtype}

\usepackage{inconsolata}

\usepackage{graphicx}

\title{Embedding Models Measure in Peculiar Ways}

\author{Juri Opitz \\
  Dept.\ of Computational Linguistics \\
  University of Zurich \\
  \texttt{opitz.sci@gmail.com} \\\And
  Andrianos Michail \\
  Dept.\ of Computational Linguistics \\
  University of Zurich \\
  \texttt{andrianos.michail@cl.uzh.ch} \\}

\begin{document}

\maketitle

\begin{abstract}
Embedding spaces define notions of semantic similarity and distance. We study whether those embeddings reflect physical measurements of mass, distance, time and volume, which admit a unique, objective notion of semantic equivalence and distance. We find that physical measurement is only weakly modeled in the embedding space, and that instead quite peculiar measurement patterns can be observed. Further analysis indicates that embedding representations of physical measurements are strongly influenced by superficial string similarity, and recalibration of similarity does not substantially improve the alignment. 
\end{abstract}

\section{Introduction}

Embedding models are routinely interpreted as semantic measurement spaces. Notions like \textit{Semantic Textual Similarity} \citep[STS,][]{cer-etal-2017-semeval} concretely pose their task as assessing ``the degree to which two sentences are semantically equivalent.'' Yet, evaluation of this property has proven tricky, also because there can often be many aspects in which two items are similar \citep{tversky1977features}.

\begin{figure}[h!]
    \centering
    \includegraphics[width=0.85\linewidth]{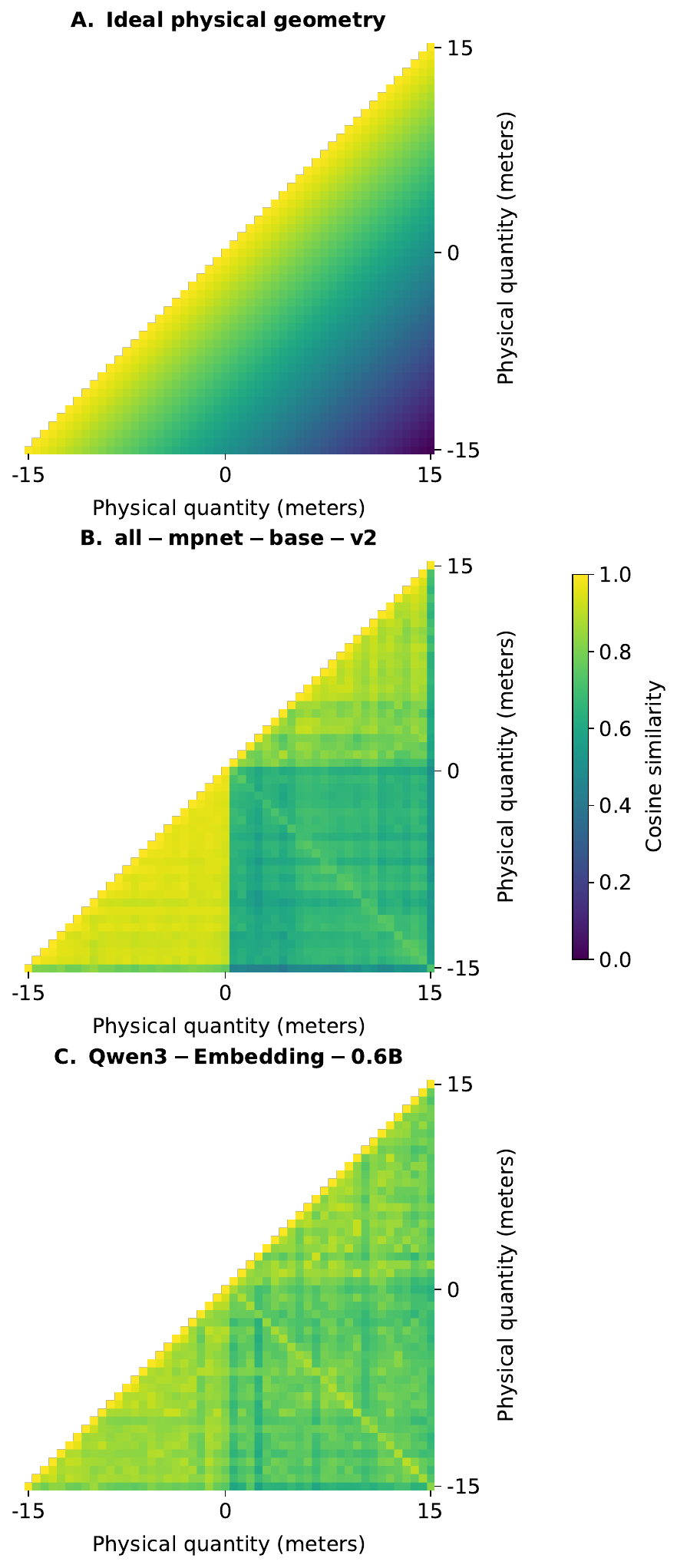}
    \caption{Ideal (A) vs.\ extracted similarities (B, C). Each square  holds the similarity of two physical measurement phrases, e.g., ``-15 meters'' vs.\ ``15 meters''.}
    \label{fig:idealized}
\end{figure}

As a focused, idealized testbed of semantic equivalence and distance, this paper examines physical measurement systems. Those systems have long enabled precise communication about the physical world, from science and engineering to everyday reasoning \cite{greaves1647discourse,d345368b-3e1e-3ffa-90ba-5afc4e854054}. In this paper, we therefore ask: 
\begin{quote}
\textit{How well are the physical metric spaces aligned with embedding metric spaces?}
\end{quote}
For instance, the expressions \textit{1 meter} and \textit{100 centimeters} clearly denote the same physical quantity and we might hope therefore that they are located in close proximity in an embedding space. Similarly, we may wonder if \textit{1 meter} is embedded closer to \textit{105 centimeters} than to, e.g., \textit{15 kilometers}, as only then would the similarity reflect the actual underlying physical relationship.

Our study finds that the physical relationships are far from faithfully reflected in embedding spaces. Instead, most peculiar patterns can be observed across all 24 models that we tested. Notably, this holds largely independently of model architecture and release date. Figure~\ref{fig:idealized} illustrates this finding: Panel A depicts an ideal measurement space in which similarity decreases smoothly with increasing physical distance. Besides the diagonal, both tested models' measurement patterns (Panels B and C) appear mostly erratic. Indeed, we can easily spot pronounced non-monotonic artifacts, including horizontal/vertical bands, as well as irregular local neighborhoods that misalign with physical distance. That said, both models seem to be able to roughly differentiate negative from positive measurements, as can be observed by the larger, darker rectangular areas on the bottom right (in B, and C).

In this work, we evaluate a set of 24 common embedding models and report three main, generalizing findings. 
\begin{enumerate}
    \item Representations of physical units and measurements are consistently weak across all 24 tested models, all showing various peculiar alignment patterns. 
    \item Models that are generally considered stronger do not clearly show better alignment in our measurement evaluation. The model family or embedding dimension also does not appear to play a major role.
    \item We find evidence that embedding similarity of physical measurements is strongly associated with lexical rather than semantic/numerical similarity, offering a plausible explanation for many of the observed non-aligned patterns. Simultaneously, via linear probes, we reject the hypothesis that the similarity is merely miscalibrated for this task.
\end{enumerate}

The remainder of this paper is structured as follows: After discussing related work in Section \ref{sec:rw}, we introduce the formal operationalization of embedding measurement and the tested embedding models in Section \ref{sec:prelim}. In Sections \ref{sec:s1} and \ref{sec:s2} we perform visual exploration, testing measurements (e.g., 2 meters, 3 meters) and unit-understanding properties of embeddings (e.g., meter, kilometer, liter, gallon, etc.), respectively. In Section \ref{sec:benchmark} we craft a focused benchmark to compare models along a single axis with regard to understanding physical measurements, testing how faithfully they reflect the physical distance relationships (The result is: Not very faithfully). Finally, in Section \ref{sec:why} we investigate potential reasons for the alignment disparities. We conclude with a discussion (Section \ref{sec:discuss}). We share code for reproducing the experiments publicly.\footnote{\url{https://github.com/flipz357/embed-and-measure}}

\section{Related Work}\label{sec:rw}

Embedding models are typically trained contrastively to minimize distances between similar texts and maximize distances between dissimilar texts. The resulting representations are used in all kinds of NLP tasks, ranging from document classification to retrieval and clustering. Early milestones of such models are based on encoders such as ``SBERT'' \cite{reimers-gurevych-2019-sentence} and ``SimCSE'' \cite{gao-etal-2021-simcse}, later ones also extend this principle of contrastive training to decoder-based LLMs \cite{zhang2025qwen3}.

Embedding models can either be evaluated through the `macro lens' via large benchmarking suites like MTEB \cite{muennighoff-etal-2023-mteb}, or a `micro lens' via focused tasks like recognizing challenging paraphrases \cite{li-etal-2025-sentence,michail2026alee}. Some works also study the interpretability of such embeddings \citep{opitz-etal-2025-interpretable}. In our paper we tighten the focus further, to a central objective phenomenon: physical measurements. While physical measurements have been explored in LLMs \cite{park-etal-2022-language} and word embeddings \cite{sundararaman-etal-2020-methods}, our exploration is conducted on the contrastively trained text representations that promise to preserve ``essential semantic and syntactic information'' \citep{Tao10.1145/3831684}. We therefore ask: To what extent do they preserve the meaning of expressions of physical measurements?

\section{General Setup}\label{sec:prelim}

\paragraph{Measurement through embeddings.} We assume that given is $\mathcal{E}$:$~T \rightarrow \mathbb{R}^n$, a function that maps from text objects to a high-dimensional vector space (i.e. `embedding model'). We further assume two quantities $(q_1, q_2)\in \mathbb{R}^2$ measured in two corresponding units $(u_1, u_2) \in T^2$. Furthermore, $str: \mathbb{R} \times T \rightarrow T$ maps a measurement in a unit to a string. Then
\begin{equation}
\label{eq:1}
\frac{\mathcal{E}\big(str(q_1, u_1)\big) \boldsymbol{\cdot} \mathcal{E}\big(str(q_2, u_2)\big)}{||\mathcal{E}\big(str(q_1, u_1)\big)|| \times ||\mathcal{E}\big(str(q_2, u_2)\big)||}
\end{equation}
 computes the standard cosine similarity score based on the vectors' dot-product ($\boldsymbol{\cdot}$) and the product of their magnitudes ($\times$).

For instance, say $q_1=10$, $q_2=1$, and $u_1=kilometer$ and $u_2=meter$. The embeddings on which the similarity is computed will be $\mathcal{E}(\text{``10 kilometers''})$ and $ \mathcal{E}(\text{``1 meter''})$.

\paragraph{Embedding models.} We select a diverse set of 24 standard and widespread embedding models, ranging from older ones based on BERT encoders, to recent larger ones based on LLM decoders, including broader model families. The full overview and discussion on the selected models is given in Appendix \ref{app:models}. When studying a specific example case in the paper, we will use \texttt{all-mpnet-base-v2}  and \texttt{Qwen3-Embedding-0.6B} as these two models are among the most downloaded embedding models and are built on encoder, and decoder, respectively. However, we will always present parallel results of other models in the Appendix. Also note that those embedding models span a range of embedding dimensionalities: from 384 (MiniLM) to 4096 (Qwen3-Embedding-8B).

\paragraph{Physical quantities and ranges.} We use the following physical quantities as measured in standard units: Length (meters), Mass (kilograms), Volume (liters), Time (seconds). We further introduce different measurement ranges as expressed by different types of ranges: Local is a range from 0 to 10; Medium from 0 to 1,000; and log for up to 100,000. Sign applies an extended scale from -100 to 100, including negative numbers. And scientific uses scientific exponential notation. The stepsizes are selected such each range is broken down into roughly 20 steps. Lastly, we express the Local scale both numerically (e.g., ``5 meters'') and as words (e.g., ``five meters''), for each integer in between zero/0 and ten/10,  including one-hundred/100. 

\section{Visual Exploration}\label{sec:s1}

\paragraph{Setup.} Given one quantity and one range, we apply the embedding model to calculate similarity for each value pair (e.g., ``5 meters'' vs. ``8 meters'', Eq.\ \ref{eq:1}). This allows us to plot and compare heatmaps according to different scales and measures, separately for each embedding model.
\begin{figure*}[h!]
    \centering
    \texttt{all-mpnet-base-v2}
    \includegraphics[width=0.86\linewidth]{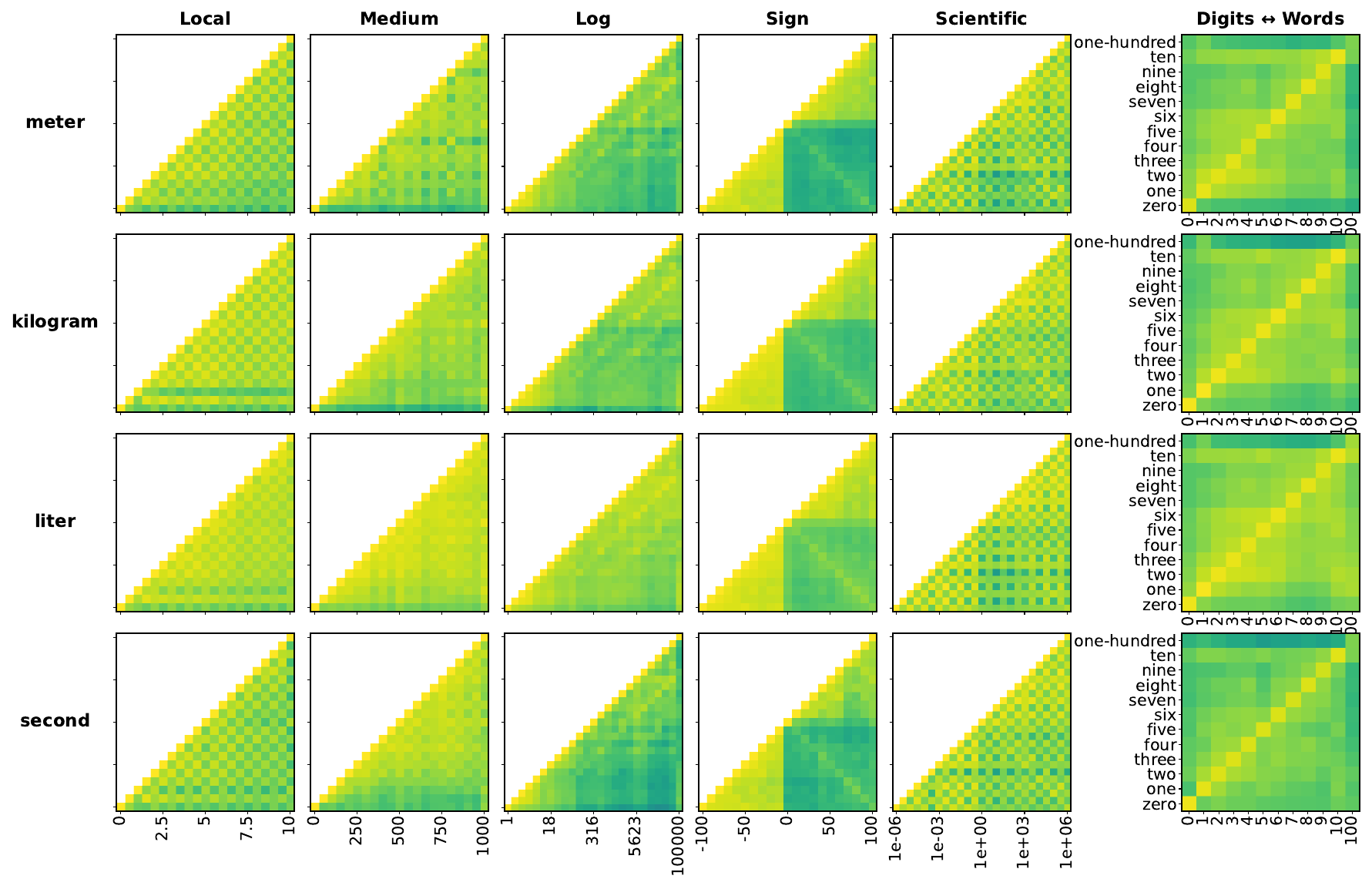}
    \texttt{Qwen3-Embedding-0.6B}
    \includegraphics[width=0.86\linewidth]{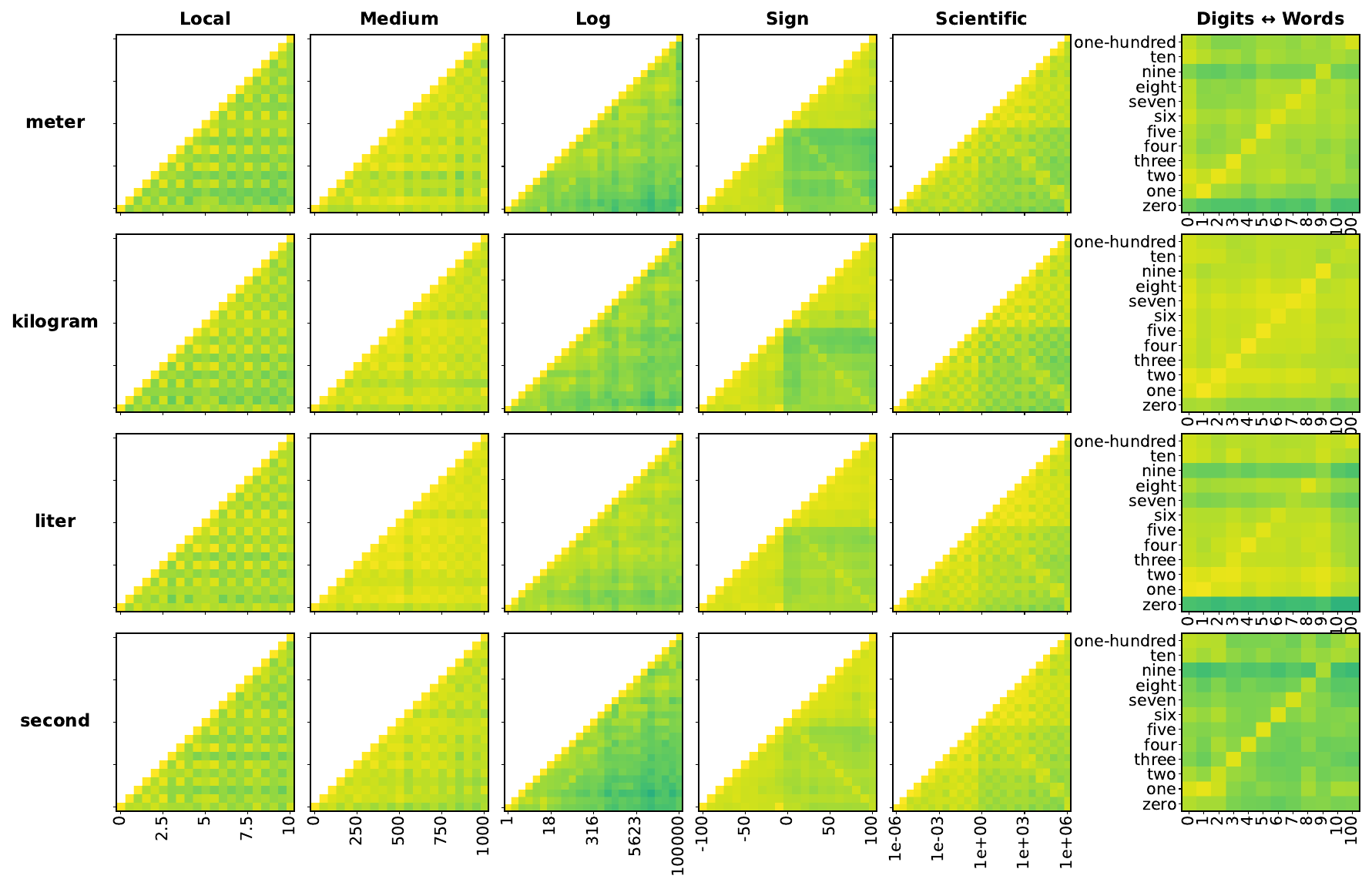}
    \caption{Large-scale measurement exploration for two example models. For most columns/plots the idealized figure is simply one that uniformly becomes more yellow towards the diagonal (Figure \ref{fig:idealized} top panel). 
    }
    \label{fig:overview-plot-mini-qwen}
\end{figure*}

\paragraph{Results.} The full results for two embedding models are visualized in Figure \ref{fig:overview-plot-mini-qwen}. Specifically, across all units and quantities, we can see a stark contrast to how a smooth idealized plot can be imagined (see the caption for description of idealized plots), observing many misaligned and peculiar patterns. 

The results for all other models are shown in Appendix \ref{app:measure plots} and show various patterns of marked misalignments. Importantly, we do not observe any marked and broader alignment difference from older embedding models that are considered ``weaker'', generally speaking, and the newer models, even if they are based on LLM-decoders. 

\paragraph{The checker pattern.} Some plots, particularly those for the \textit{Local scale} (left column, Figure \ref{fig:overview-plot-mini-qwen}) exhibit a pronounced ``checker pattern,'' dominated by aligned integer-integer pairs alternating with non-aligned integer-float pairs.  E.g., for both models 2.5 \{meters, kilograms, liters, seconds\} is more similar to 7.5 \{meters, kilograms, liters, seconds\} than to 3 \{meters, kilograms, liters, seconds\}, which misaligns with the physical measurement. The explanation for this may be intuitive and leads to a first hypothesis for the misalignment: Embedding similarity for measurements is mostly driven by surface similarity, and not the actual underlying metric distance (in the last part of this piece, we investigate this hypothesis in depth). 

\paragraph{Numerical and literal expressions.} We study the alignment of quantities expressed numerically and literally, in the right column of Figure \ref{fig:overview-plot-mini-qwen}. Strikingly, we see that embeddings barely capture this semantic equivalence, but some interesting model differences can be observed: For instance, Qwen assigns high similarities to \textit{one/1}, as well as to \textit{one/0}---most pronounced for the liter and kilogram quantity. On the other hand, mpnet assigns one and 1 a markedly higher similarity than to one and all other numbers that are not 1. 

Further, Qwen shows some interesting horizontal lines when aligning digits with words. For example, it appears that ``nine liters'' are rated as dissimilar to all other liter quantities expressed in digits, whereas this is not so for, e.g., ``one liter'' or ``two liters.'' For nine/9 and ten/10, even the self comparison is comparatively weak, showing pronounced misalignment between the numerical and literal expression of the same number. With the exception of one-hundred/100, the self-paraphrased numbers are better aligned in the mpnet model that was released a few years before Qwen.

\section{Testing Conversion Understanding}
\label{sec:s2}

\begin{figure}[ht]
    \centering
    \texttt{all-mpnet-base-v2}
    \includegraphics[width=1\linewidth]{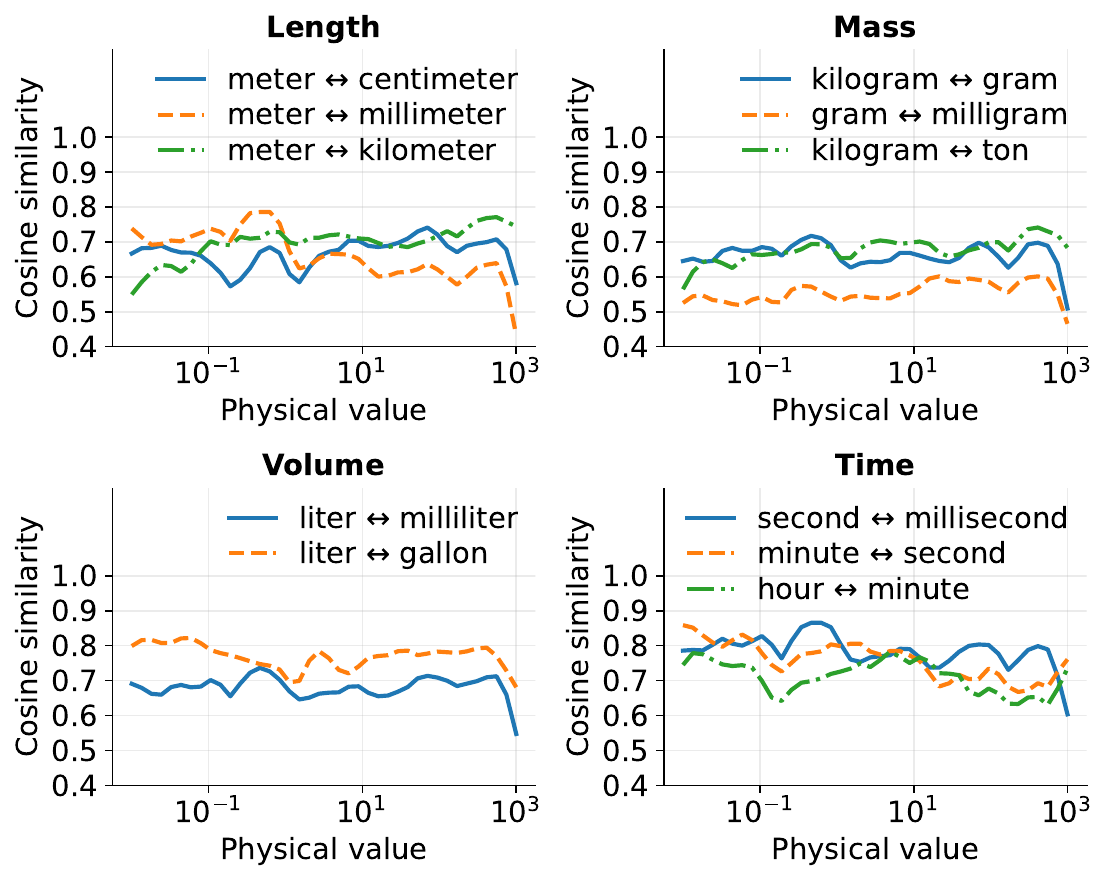}
    \texttt{Qwen3-Embedding-0.6B}
    \includegraphics[width=1\linewidth]{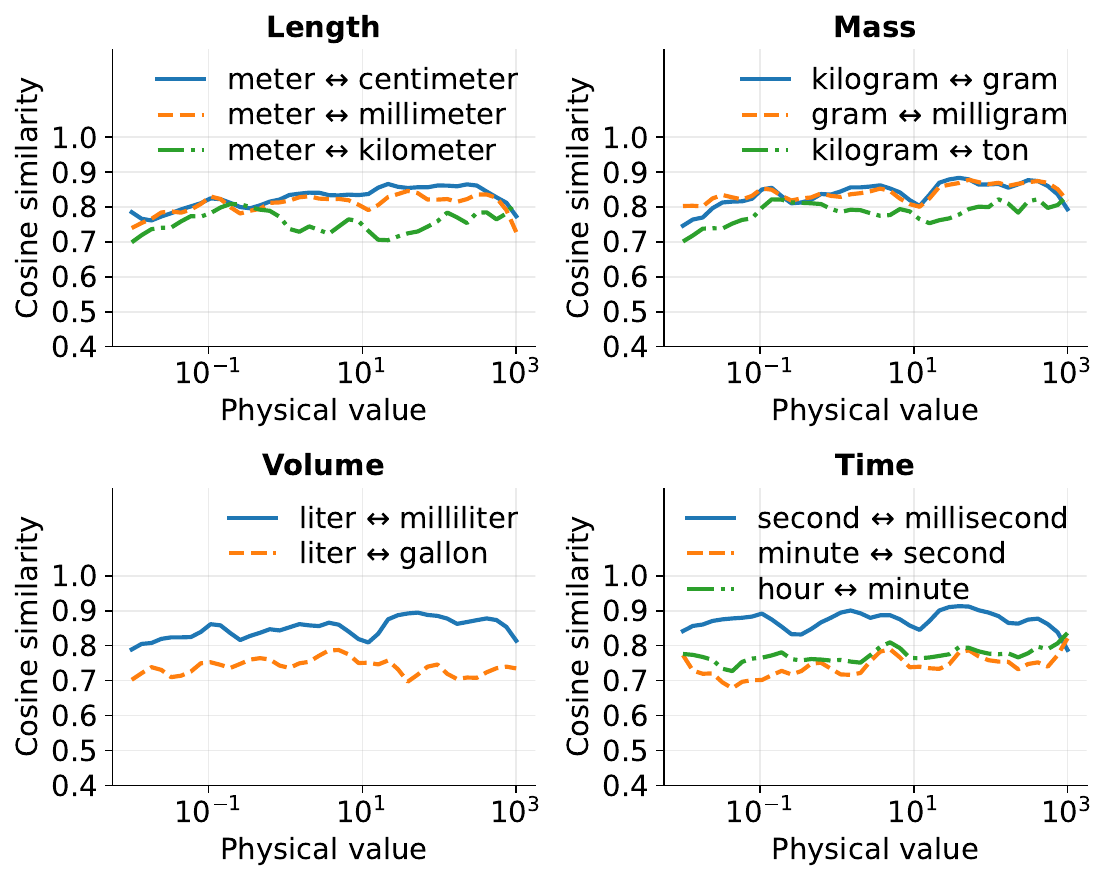}
    \caption{Conversion analysis. For each quantity and two example models there is one plot. Each line in each plot indicates a conversion-alignment. For example, in the plot on the top left, the y-value at $x\approx10^0$ denotes the embedding similarity between $\approx$1 meter and: $\approx$100 centimeters (blue line), $\approx$1000 millimeters (orange line) and $\approx$0.001 kilometers (green line). Curves are lightly smoothed over 40 log-spaced values.}
    \label{fig:alignment}
\end{figure}

For this study, we ask how well embedding spaces capture unit conversion. For instance, one hour is 60 minutes, whereas one gallon is about 3.78 liters.

Figure \ref{fig:alignment} shows how well different units align across the different types of quantities. In all of the shown plots in this figure, the idealized alignment would be a horizontal line that is very close to 1.0. However, when inspecting the alignment through embedding spaces, we again see rather diffuse patterns of alignment. This time, the lines tend to be slightly higher in the Qwen embedding space, potentially indicating better alignment across different units. Particularly, as the quantities become larger, there is a strong alignment performance drop-off observed in mpnet, but not so much in Qwen. The comparatively most pronounced alignment seems to be in Qwen when relating seconds and milliseconds - it is the highest line, reaching about 0.9. However, ultimately, no line is horizontal and close to 1.0, and so we can conclude that unit conversion is also highly fuzzy in both embedding models. 

Other embedding models are shown in Appendix \ref{app:conversion} and show similarly non-aligned patterns, with some particularly erratic ones.

\section{Crafting a Small Benchmark}\label{sec:benchmark}

We would like to assign an average degree of `task performance' to the models, in order to compare them with respect to their capacity of modeling the physical relationships, ranking the models on a single axis. 

\paragraph{Setup.} We use two performance metrics. \textbf{Kendall's $\tau$} is a classic non-parametric statistic that measures the ordinal association between two variables, well suited to this type of evaluation. We further define as an additional metric the \textbf{Pairwise Accuracy (PACC)}, which simply calculates the ratio of correctly ranked pairs (in practice, both metrics are very similar, we  include PACC primarily for easier readability). For instance, if  the similarity of \textit{2 meters} to \textit{2.5 meters} is higher than to, e.g., \textit{7 meters}, a point is added; the resulting sum is normalized by all comparisons. We calculate both performance metrics across every scale and across every unit: Given a unit, a scale, and one `anchor' item, we calculate true distances and embedding distances of this item to the other items. The Kendall's $\tau$ and the PACC are computed and averaged across the anchor items, and then again across each unit and scale. We call the resulting scalar-valued score and the benchmark in total `PhysScore'.

\begin{table}[ht]
    \centering
    \adjustbox{width=\linewidth}{
\begin{tabular}{lrr}
\toprule
model & K's $\tau$ & PACC \\
\midrule
multilingual\_e5\_large\_instruct & 50.29 & 75.37  \\
LaBSE & 44.62 & 72.53  \\
all\_mpnet\_base\_v2 & 43.24 & 71.83  \\
harrier\_oss\_v1\_270m & 41.97 & 71.16  \\
harrier\_oss\_v1\_0.6b & 40.86 & 70.61  \\
e5\_large\_v2 & 40.45 & 70.40  \\
Qwen3\_Embedding\_4B & 39.51 & 69.94  \\
bge\_m3 & 39.42 & 69.90 \\
DenseOn & 39.25 & 69.82  \\
multilingual\_e5\_large & 39.21 & 69.80   \\
all\_MiniLM\_L6\_v2 & 39.03 & 69.72  \\
Qwen3\_Embedding\_0.6B & 38.44 & 69.39  \\
Qwen3\_Embedding\_8B & 37.23 & 68.80 \\
multilingual\_e5\_base & 35.42 & 67.89  \\
bge\_large\_en\_v1.5 & 35.17 & 67.77  \\
mxbai\_embed\_large\_v1 & 34.96 & 67.66  \\
granite\_embedding\_311m\_multilingual\_r2 & 31.14  & 65.75 \\
granite\_embedding\_278m\_multilingual & 30.36 & 65.34  \\
granite\_embedding\_english\_r2 & 27.53 & 63.92 \\
granite\_embedding\_107m\_multilingual & 27.13 & 63.71  \\
nomic\_embed\_text\_v1.5 & 26.98 & 63.64  \\
granite\_embedding\_97m\_multilingual\_r2 & 26.38 & 63.35  \\
paraphrase\_multilingual\_mpnet\_base\_v2 & 25.76 & 63.01  \\
embeddinggemma\_300m & 25.41 & 62.86  \\
\midrule
all & 35.82 & 68.09  \\
random-baseline & 00.00 & 50.00 \\
\bottomrule
\end{tabular}}
    \caption{Result on PhysScore.}
    \label{tab:physcore}
\end{table}

\paragraph{Result.} Table \ref{tab:physcore} shows the  overall performance ranking of models. While all models outperform the random baseline, even the best model, an e5 variant (multilingual-e5-large-instruct), only reaches a Kendall's $\tau$ of 50.3. Hence, we conclude that all models are only weakly aligned with the physical measurements. Another observation is that model scale seems to have no clear positive effect on the alignment capacity. Sometimes even, an inverse effect is observed: The similarity of the smaller Qwen models sometimes achieves a better alignment than the larger ones.

\paragraph{Which measurements are hardest?} To explore this question, we show in Table \ref{tab:kendall_by_unit_range} the average, minimum, and maximum Kendall's $\tau$ across all embedding models, for each quantity, unit, and range.

\begin{table}[ht]
\adjustbox{width=\linewidth}{
\begin{tabular}{llccc}
\toprule
Unit & Range & Avg & Min & Max \\
\midrule
kilogram & local\_0\_10 & 31.00 & 19.04 & 46.95 \\
kilogram & log\_1\_100k & 36.38 & 14.82 & 66.53 \\
kilogram & medium\_0\_1000 & 38.59 & 20.06 & 53.39 \\
kilogram & scientific & 31.20 & 11.44 & 51.15 \\
kilogram & sign\_split & 34.56 & 3.69 & 58.35 \\
kilogram & AVG & 34.35 & 13.81 & 55.28 \\
\midrule
liter & local\_0\_10 & 31.78 & 09.81 & 48.91 \\
liter & log\_1\_100k & 39.63 & 15.34 & 65.34 \\
liter & medium\_0\_1000 & 40.30 & 17.11 & 57.08 \\
liter & scientific & 31.64 & 13.09 & 53.44 \\
liter & sign\_split & 33.40 & 3.12 & 55.26 \\
liter & AVG & 35.35 & 11.69 & 56.01 \\
\midrule
meter & local\_0\_10 & 32.99 & 25.03 & 47.99 \\
meter & log\_1\_100k & 43.60 & 14.47 & 73.55 \\
meter & medium\_0\_1000 & 38.56 & 19.00 & 56.22 \\
meter & scientific & 31.64 & 12.94 & 51.96 \\
meter & sign\_split & 36.22 & 4.16 & 56.85 \\
meter & AVG & 36.60 & 15.12 & 57.31 \\
\midrule
second & local\_0\_10 & 27.63 & 16.95 & 40.00 \\
second & log\_1\_100k & 44.62 & 13.05 & 75.69 \\
second & medium\_0\_1000 & 41.08 & 19.66 & 58.23 \\
second & scientific & 34.58 & 13.03 & 58.90 \\
second & sign\_split & 37.06 & 6.72 & 55.75 \\
second & AVG & 37.00 & 13.88 & 57.71 \\
\bottomrule
\end{tabular}}
\caption{Average, minimum, and maximum Kendall correlation across models for each unit and range.}
\label{tab:kendall_by_unit_range}
\end{table}

We see that Mass (kilogram) appears as the most challenging for all embedding models (on average), with an average Kendall's  $\tau$ of only 34.4. Overall the relatively best aligned quantities in this experiment appear to be length and time (meters and seconds). Regarding scales, the scientific notation and local scale prove more challenging for the models, on average.

\section{Analysis}
\label{sec:why}

\subsection{Is Similarity just Miscalibrated?}

Through our embedding models, we calculated similarity in the most frequent way: Using cosine similarity. However, this assigns each dimension in the embedding representation the same prior weight. Thus, it could be that the representations hold highly useful information for measuring, but it gets blurred when applying the dot-product.

To investigate this hypothesis, we perform five random train-test split, and re-evaluate the models by averaging over each test split. The baseline for this experiment is again the standard cosine similarity, and the studied model is a linear (Ridge) regression on the absolute difference of the embedding vectors of two measurement expressions (aka \textbf{linear probe}). This gives the option to potentially emphasize those dimensions which could perhaps hold the information about physical relationships.

\begin{table}[ht]
    \centering
\adjustbox{width=\linewidth}{\begin{tabular}{lcc|cc|c}
\toprule
& \multicolumn{2}{c}{Cosine} & \multicolumn{2}{c}{Linear probe} & \\
\cmidrule(lr){2-3}\cmidrule(lr){4-5}
model & K's $\tau$ & PACC & K's $\tau$ & PACC & $\Delta\tau$ \\
\midrule
multilingual\_e5\_large\_instruct & 44.7 & 72.5 & 51.4 & 75.9 & 6.7 \\
LaBSE & 43.6 & 72.0 & 50.1 & 75.3 & 6.5 \\
e5\_large\_v2 & 35.7 & 68.0 & 48.1 & 74.3 & 12.3 \\
harrier\_oss\_v1\_270m & 39.5 & 69.9 & 47.3 & 73.9 & 7.8 \\
multilingual\_e5\_large & 38.0 & 69.2 & 43.5 & 72.0 & 5.5 \\
all\_mpnet\_base\_v2 & 40.4 & 70.4 & 43.0 & 71.7 & 2.6 \\
Qwen3\_Embedding\_4B & 35.7 & 68.0 & 41.9 & 71.1 & 6.2 \\
Qwen3\_Embedding\_8B & 33.7 & 67.0 & 41.5 & 70.9 & 7.8 \\
harrier\_oss\_v1\_0.6b & 34.4 & 67.4 & 40.1 & 70.2 & 5.6 \\
bge\_m3 & 37.3 & 68.8 & 39.7 & 70.0 & 2.3 \\
multilingual\_e5\_base & 31.9 & 66.1 & 39.3 & 69.8 & 7.4 \\
all\_MiniLM\_L6\_v2 & 37.1 & 68.7 & 37.0 & 68.7 & -0.1 \\
DenseOn & 38.9 & 69.6 & 35.6 & 68.0 & -3.3 \\
granite\_embedding\_278m\_multilingual & 30.0 & 65.1 & 33.2 & 66.7 & 3.2 \\
mxbai\_embed\_large\_v1 & 31.0 & 65.6 & 32.5 & 66.4 & 1.6 \\
bge\_large\_en\_v1.5 & 31.9 & 66.1 & 32.3 & 66.3 & 0.4 \\
granite\_embedding\_311m\_multilingual\_r2 & 28.9 & 64.6 & 32.2 & 66.2 & 3.3 \\
Qwen3\_Embedding\_0.6B & 33.7 & 67.0 & 31.9 & 66.1 & -1.8 \\
nomic\_embed\_text\_v1.5 & 25.8 & 63.1 & 29.1 & 64.7 & 3.3 \\
granite\_embedding\_97m\_multilingual\_r2 & 26.8 & 63.5 & 28.7 & 64.5 & 1.9 \\
paraphrase\_multilingual\_mpnet\_base\_v2 & 20.4 & 60.3 & 26.9 & 63.6 & 6.4 \\
granite\_embedding\_english\_r2 & 25.5 & 62.9 & 26.7 & 63.5 & 1.3 \\
embeddinggemma\_300m & 28.3 & 64.3 & 21.0 & 60.6 & -7.3 \\
granite\_embedding\_107m\_multilingual & 29.3 & 64.8 & 20.4 & 60.3 & -8.9 \\
\midrule
all & 33.4 & 66.9 & 36.4 & 68.4 & 3.0 \\
\bottomrule
\end{tabular}}
    \caption{Result of recalibration experiment expressed in Kendall's $\tau \times 100$ on held-out testing data.}
    \label{tab:recalib}
\end{table}

\begin{table*}[ht]
    \centering
\adjustbox{width=\linewidth}{\begin{tabular}{lrrrr|rrrr|rrrr}
\toprule
& \multicolumn{4}{c}{Character}& \multicolumn{4}{c}{Tokenizer}& \multicolumn{4}{c}{Numeric} \\
\cmidrule(lr){2-5}\cmidrule(lr){6-9}\cmidrule(lr){10-13}
Model & All & $F{\times}F$ & $F{\times}I$ & $I{\times}I$ & All & $F{\times}F$ & $F{\times}I$ & $I{\times}I$ & All & $F{\times}F$ & $F{\times}I$ & $I{\times}I$ \\
\midrule
\texttt{all-MiniLM-L6-v2} & 21.0 & 22.0 & 17.1 & 44.2 & 15.1 & 10.4 & 10.2 & 37.1 & 28.1 & 23.0 & 28.7 & 39.0 \\
\texttt{all-mpnet-base-v2} & 27.7 & 31.7 & 21.0 & 41.9 & 20.4 & 19.9 & 8.7 & 34.0 & 26.1 & 40.9 & 26.0 & 36.2 \\
\texttt{bge-large-en-v1.5} & 10.9 & 25.2 & 29.1 & 35.2 & 7.7 & 22.9 & 31.1 & 18.5 & 8.0 & 26.5 & 6.9 & 18.6 \\
\texttt{bge-m3} & 18.1 & 31.3 & 36.8 & 49.2 & 3.6 & 25.0 & 25.7 & 31.9 & 19.5 & 34.3 & 22.7 & 29.7 \\
\texttt{DenseOn} & 23.5 & 18.1 & 21.0 & 36.9 & 27.4 & 17.0 & 25.1 & 31.6 & 20.2 & 20.2 & 18.9 & 33.3 \\
\texttt{e5-large-v2} & 29.0 & 23.9 & 27.2 & 45.1 & 22.0 & 15.2 & 14.5 & 38.7 & 16.9 & 18.2 & 15.3 & 32.8 \\
\texttt{embeddinggemma-300m} & 51.6 & 36.3 & 39.6 & 44.6 & 51.6 & 36.3 & 39.6 & 44.6 & 3.3 & 8.3 & 2.1 & 8.8 \\
\texttt{granite-embedding-107m-multilingual} & 27.0 & 25.9 & 32.4 & 43.1 & 16.7 & 10.9 & 14.1 & 31.0 & 16.7 & 14.9 & 16.4 & 20.6 \\
\texttt{granite-embedding-278m-multilingual} & 29.4 & 25.1 & 33.2 & 46.3 & 16.3 & 11.8 & 14.5 & 25.2 & 14.8 & 11.7 & 14.6 & 21.1 \\
\texttt{granite-embedding-311m-multilingual-r2} & 14.2 & 32.5 & 33.3 & 53.6 & 14.2 & 32.5 & 33.3 & 53.6 & 9.1 & 17.0 & 10.4 & 11.8 \\
\texttt{granite-embedding-97m-multilingual-r2} & 4.9 & 17.5 & 11.9 & 27.9 & 8.5 & 18.9 & 2.5 & 29.3 & 12.5 & 12.5 & 12.7 & 21.2 \\
\texttt{granite-embedding-english-r2} & 16.7 & 29.9 & 29.6 & 44.6 & 9.1 & 14.7 & 4.6 & 21.1 & 14.4 & 20.5 & 10.8 & 22.6 \\
\texttt{harrier-oss-v1-0.6b} & 27.0 & 42.1 & 32.0 & 43.0 & 27.0 & 42.1 & 32.0 & 43.0 & 21.7 & 46.1 & 29.7 & 32.6 \\
\texttt{harrier-oss-v1-270m} & 28.4 & 42.5 & 30.5 & 43.1 & 28.4 & 42.5 & 30.5 & 43.1 & 23.3 & 48.5 & 32.8 & 38.0 \\
\texttt{LaBSE} & 39.6 & 32.7 & 28.5 & 48.7 & 42.7 & 42.0 & 13.0 & 35.7 & 18.1 & 29.7 & 18.8 & 24.4 \\
\texttt{multilingual-e5-base} & 23.5 & 32.6 & 34.3 & 50.3 & 6.8 & 27.3 & 13.6 & 27.7 & 17.2 & 24.1 & 23.3 & 25.4 \\
\texttt{multilingual-e5-large} & 25.5 & 32.4 & 29.4 & 50.1 & 6.8 & 24.9 & 0.9 & 31.2 & 18.4 & 37.4 & 26.9 & 24.9 \\
\texttt{multilingual-e5-large-instruct} & 39.7 & 28.1 & 29.4 & 40.4 & 27.5 & 25.3 & 15.9 & 22.4 & 26.9 & 49.1 & 29.9 & 30.7 \\
\texttt{mxbai-embed-large-v1} & 10.0 & 24.2 & 28.7 & 33.9 & 5.4 & 16.0 & 26.6 & 17.1 & 6.9 & 18.0 & 6.1 & 16.5 \\
\texttt{nomic-embed-text-v1.5} & -5.6 & 16.9 & 28.9 & 32.8 & -16.9 & -3.7 & 5.7 & 17.5 & 6.5 & 5.8 & 5.9 & 15.2 \\
\texttt{paraphrase-multilingual-mpnet-base-v2} & 5.0 & 17.8 & 19.8 & 29.7 & -11.5 & -5.0 & 3.7 & 6.4 & 8.1 & 14.4 & 7.0 & 8.2 \\
\texttt{Qwen3-Embedding-0.6B} & 47.6 & 23.6 & 30.7 & 50.8 & 47.6 & 23.6 & 30.7 & 50.8 & 11.7 & 13.7 & 13.7 & 25.2 \\
\texttt{Qwen3-Embedding-4B} & 30.1 & 26.3 & 38.8 & 53.8 & 30.1 & 26.3 & 38.8 & 53.8 & 13.1 & 12.6 & 8.4 & 27.3 \\
\texttt{Qwen3-Embedding-8B} & 36.7 & 33.9 & 39.2 & 49.5 & 36.7 & 33.9 & 39.2 & 49.5 & 18.3 & 24.2 & 15.8 & 25.1 \\
\bottomrule
\end{tabular}}
\caption{Kendall's $\tau$ ($\times 100$; range $-100$ to $100$, higher = stronger association) between embedding similarity and three reference similarities: character-level Levenshtein (Character), token-level Levenshtein (Tokenizer), and true numerical proximity (Numeric). Columns distinguish float--float ($F{\times}F$), float--integer ($F{\times}I$), integer--integer ($I{\times}I$), and all pairs (All). The identical character and tokenizer columns for the three Qwen3 models, both harrier models, embeddinggemma, and granite-311m-r2 appear due to their tokenizer splitting digits into characters.}
\label{tab:superficial-similarity}
\end{table*}

The results on the test split are shown in Table \ref{tab:recalib}. Overall, changes by recalibration are small, only +3 points on average (from 33 to 36 $\tau$). The largest increase in Kendall's $\tau$ points is +12.33, but the absolute value for this model is still fairly small (48 $\tau$). Furthermore, \textit{no model} exceeds 52 $\tau$, even after recalibration.

Regarding the Qwen model family, this time the larger ones appear to profit more from recalibration, so their representation may hold the information in better ways, but not aligned with the similarity. Still, overall alignment is low.

In sum, we find no convincing support for the hypothesis that the misalignment with physical relationships is just due to miscalibration of the similarity function.\footnote{This is in line with  \citet{huber2026exploring} who find that the original similarity function of the contrastively trained models is already fairly optimal.} 

\subsection{The Effect of Superficial String Overlap}

Based on the results from our visual explorations, we had already speculated that one major driver of the misalignments is an over-focus on superficial string similarity. Because of this, embedding models might tend to neglect the actual underlying physical relationships. In this section, we empirically investigate this hypothesis.

\paragraph{Setup.} We sample 50 random floats (5 decimals) and 50 random integers, i.e., one series of floating-point numbers ($F$) and another series of integers ($I$), both $\in [-1000,+1000]$. We convert these numbers to strings. For every unique pair of these number strings (joining floating and integer series), we calculate 1.\ The Levenshtein distances based on the characters in the strings; 2.\ The Levenshtein distances based on the generated sub-word sequences of the embedding models themselves (`tokens'); and 3.\ The `ground-truth' so to speak, i.e., the actual distance between the numbers. We negate these distances and compute Kendall's $\tau$ between embedding similarity and the (inverse) distances. Put simply, a higher correlation implies a stronger link between embedding similarity and another similarity (either character, sub-word, or ground truth). We report correlations separately for float--float ($F{\times}F$), float--integer ($F{\times}I$), and integer--integer ($I{\times}I$) pairs, as well as over all pairs jointly (All).

\paragraph{Result.} The results in Table \ref{tab:superficial-similarity} reveal a clear distinction between lexical similarity and numerical semantics. Embedding similarity tends to be more strongly associated with the textual representation for 19 of 24 models. Only for 5 models (MiniLM, bge-m3, granite-97m-r2, nomic and paraphrase-mpnet) the numeric correlation is higher than both lexical ones. In other words, numbers that \emph{look} similar---for example, because they share digits or token fragments---often receive more similar embeddings than numerically close numbers.

For example, the Qwen3 embeddings exhibit a fair correlation with both character-level and tokenizer-level Levenshtein similarity. At the same time, their correlation with the true numerical distance is comparatively weaker. This suggests that Qwen3 primarily encodes the numbers as lexical objects rather than as quantities with an inherent ordering or magnitude. 

Overall, these findings support our initial hypothesis that modern embedding models may largely process numbers according to their textual representation rather than their quantitative meaning. If embeddings primarily organize numbers according to shared digits, prefixes, or tokenizer fragments, then semantic operations over numerical quantities become unreliable and highly dependent on the accidental textual form of the numbers. 

This effect persists even for recent, large-scale embedding models, suggesting that increased model capacity alone does not necessarily lead to more faithful representations of physical statements. Indeed, one of the strongest correlations with string similarity and one of the weakest correlations with numerical distance is observed for Google's LLM-based gemma embedding (\texttt{embeddinggemma-300m}).

\section{Discussion}
\label{sec:discuss}

From our study there is one central conclusion:
\begin{quote}
\textit{Embedding model similarity is only weakly aligned with physical measurements.} 
\end{quote}
While our range of tested embedding models is necessarily limited, we did not observe a systematic improvement between older to more recent models, suggesting that it is not simply a matter of scale or recency. This raises a central question: 
\begin{quote}
\textit{What drives these misalignments?}
\end{quote}
Our final experiment provides one potential explanation. Embedding similarity is more strongly associated with superficial lexical similarity---measured either at the character level or by the models' own tokenization---than with the actual numerical relationships between values. In particular, even LLM-based embedding models like Qwen3 exhibit moderate dependence on string similarity while showing only weak correspondence to the underlying quantitative semantics. Recalibration of the similarity function did not alleviate this issue. This suggests that at least part of the observed misalignment stems from the models organizing representations according to surface form rather than meaning.

While the work of \citet{weller2026on} might suggest that the misalignments reflect a theoretical limitation of embedding models by their dimensionality, we do not observe a corresponding trend in alignment quality that would correspond to their dimensionality. 

Instead, we believe that our results point toward the learning objective itself. Contrastive training may simply provide insufficient pressure to organize the embedding space according to physical or quantitative relationships. For such alignment to emerge without explicit supervision, it would have to constitute an emergent capability of embedding models (or be an explicit objective in benchmarks). This naturally raises another question: 
\begin{quote}
\textit{Should similarity be aligned with physical measurements?} 
\end{quote}
For many retrieval and clustering applications, understanding quantitative relationships may play only a minor role. And lexical similarity can often be a useful inductive bias, particularly for identifiers, version numbers, dates, or product codes. Nevertheless, it seems also fair to expect that any embedding model that promises to measure \textit{semantic equivalence}, or \textit{semantic similarity}, should clearly judge ``2 kilometers'' and ``2000 meters'' as more similar than ``2 kilometers'' and ``2 meters''. 

More generally, the inability to capture such elementary relationships may become increasingly problematic as embeddings are employed beyond conventional retrieval tasks, for example as semantic representations in scientific, engineering, or agentic systems that require reasoning about quantities and measurements. Taken together, our findings provide further evidence for the conclusion of \citet{fodor-etal-2025-compositionality} that ``state-of-the-art transformers poorly capture the pattern of human semantic similarity judgments'' and agree with \citet{sun2026position} who argue that embeddings should capture implicit semantics. 

Our evaluation through physical measurements may thus provide an additional quality metric in future model development, or a focused test of whether an embedding model can produce representations with a deeper understanding of the content. To conclude, we wish to emphasize what \citet{thawani-etal-2021-representing} say: In NLP, ``numbers are important'', and ``numbers are neglected''; Our work provides evidence that numbers, and, more broadly, physical measurements, are misaligned in contrastively trained text representation models.

\section*{Acknowledgements}

We thank Halil Talha Ertan for their useful feedback. This work has been supported by the Swiss National Science Foundation (grant no.\ CRSII5\_213585) and by the Luxembourg National Research Fund (grant no.\ 17498891).

\bibliography{custom}

@article{zhang2025qwen3, 
    title={Qwen3 embedding: {Advancing} text embedding and reranking through foundation models},
    author={Zhang, Yanzhao and Li, Mingxin and Long, Dingkun and Zhang, Xin and Lin, Huan and Yang, Baosong and Xie, Pengjun and Yang, An and Liu, Dayiheng and Lin, Junyang and others},
    journal={arXiv preprint arXiv:2506.05176},
    year={2025},
    url={https://arxiv.org/abs/2506.05176},
}

@inproceedings{park-etal-2022-language,
    title = "Do Language Models Understand Measurements?",
    author = "Park, Sungjin  and
      Ryu, Seungwoo  and
      Choi, Edward",
    editor = "Goldberg, Yoav  and
      Kozareva, Zornitsa  and
      Zhang, Yue",
    booktitle = "Findings of the Association for Computational Linguistics: EMNLP 2022",
    month = dec,
    year = "2022",
    address = "Abu Dhabi, United Arab Emirates",
    publisher = "Association for Computational Linguistics",
    url = "https://aclanthology.org/2022.findings-emnlp.128/",
    doi = "10.18653/v1/2022.findings-emnlp.128",
    pages = "1782--1792"
}

@article{huber2026exploring,
  title={Exploring Dowker Homology for Sentence Similarity},
  author={Huber, Marius and Opitz, Juri},
  journal={arXiv preprint arXiv:2608.22909},
  year={2026}
}

@inproceedings{thawani-etal-2021-representing,
    title = "Representing Numbers in {NLP}: a Survey and a Vision",
    author = "Thawani, Avijit  and
      Pujara, Jay  and
      Ilievski, Filip  and
      Szekely, Pedro",
    editor = "Toutanova, Kristina  and
      Rumshisky, Anna  and
      Zettlemoyer, Luke  and
      Hakkani-Tur, Dilek  and
      Beltagy, Iz  and
      Bethard, Steven  and
      Cotterell, Ryan  and
      Chakraborty, Tanmoy  and
      Zhou, Yichao",
    booktitle = "Proceedings of the 2021 Conference of the North American Chapter of the Association for Computational Linguistics: Human Language Technologies",
    month = jun,
    year = "2021",
    address = "Online",
    publisher = "Association for Computational Linguistics",
    url = "https://aclanthology.org/2021.naacl-main.53/",
    doi = "10.18653/v1/2021.naacl-main.53",
    pages = "644--656"
}

@article{Tao10.1145/3831684,
author = {Tao, Chongyang and Shen, Tao and Gao, Shen and Zhang, Junshuo and Li, Zhen and Hua, Kai and Hu, Wenpen and Tao, Zhangwei and Ma, Shuai},
title = {LLMs are Also Effective Embedding Models: An In-depth Overview},
year = {2026},
publisher = {Association for Computing Machinery},
address = {New York, NY, USA},
issn = {1046-8188},
url = {https://doi.org/10.1145/3831684},
doi = {10.1145/3831684},
note = {Just Accepted},
journal = {ACM Trans. Inf. Syst.},
month = jul
}

@inproceedings{
sun2026position,
title={Position: Text Embeddings Should Capture Implicit Semantics, Not Just Surface Meaning},
author={Yiqun Sun and Qiang Huang and Anthony Kum Hoe Tung and Jun Yu},
booktitle={Forty-third International Conference on Machine Learning Position Paper Track},
year={2026},
url={https://openreview.net/forum?id=7NL8On4v6m}
}

@inproceedings{sundararaman-etal-2020-methods,
    title = "Methods for Numeracy-Preserving Word Embeddings",
    author = "Sundararaman, Dhanasekar  and
      Si, Shijing  and
      Subramanian, Vivek  and
      Wang, Guoyin  and
      Hazarika, Devamanyu  and
      Carin, Lawrence",
    editor = "Webber, Bonnie  and
      Cohn, Trevor  and
      He, Yulan  and
      Liu, Yang",
    booktitle = "Proceedings of the 2020 Conference on Empirical Methods in Natural Language Processing (EMNLP)",
    month = nov,
    year = "2020",
    address = "Online",
    publisher = "Association for Computational Linguistics",
    url = "https://aclanthology.org/2020.emnlp-main.384/",
    doi = "10.18653/v1/2020.emnlp-main.384",
    pages = "4742--4753"
}

@inproceedings{li-etal-2025-sentence,
    title = "Sentence Smith: Controllable Edits for Evaluating Text Embeddings",
    author = "Li, Hongji  and
      Michail, Andrianos  and
      Gubelmann, Reto  and
      Clematide, Simon  and
      Opitz, Juri",
    editor = "Christodoulopoulos, Christos  and
      Chakraborty, Tanmoy  and
      Rose, Carolyn  and
      Peng, Violet",
    booktitle = "Proceedings of the 2025 Conference on Empirical Methods in Natural Language Processing",
    month = nov,
    year = "2025",
    address = "Suzhou, China",
    publisher = "Association for Computational Linguistics",
    url = "https://aclanthology.org/2025.emnlp-main.1343/",
    doi = "10.18653/v1/2025.emnlp-main.1343",
    pages = "26428--26445",
    ISBN = "979-8-89176-332-6"
}

@article{michail2026alee,
  title={ALEE: Any-Language Evaluation of Embeddings via English-Centric Minimal Pairs},
  author={Michail, Andrianos and Psychias, Stylianos and Wastl, Michelle and Clematide, Simon and Sennrich, Rico and Opitz, Juri},
  journal={arXiv preprint arXiv:2607.00171},
  year={2026}
}

@inproceedings{gao-etal-2021-simcse,
    title = "{S}im{CSE}: Simple Contrastive Learning of Sentence Embeddings",
    author = "Gao, Tianyu  and
      Yao, Xingcheng  and
      Chen, Danqi",
    editor = "Moens, Marie-Francine  and
      Huang, Xuanjing  and
      Specia, Lucia  and
      Yih, Scott Wen-tau",
    booktitle = "Proceedings of the 2021 Conference on Empirical Methods in Natural Language Processing",
    month = nov,
    year = "2021",
    address = "Online and Punta Cana, Dominican Republic",
    publisher = "Association for Computational Linguistics",
    url = "https://aclanthology.org/2021.emnlp-main.552/",
    doi = "10.18653/v1/2021.emnlp-main.552",
    pages = "6894--6910"
}

@article{fodor-etal-2025-compositionality, 
    title={Compositionality and Sentence Meaning: {Comparing} Semantic Parsing and Transformers on a Challenging Sentence Similarity Dataset},
    author={Fodor, James  and De Deyne, Simon  and Suzuki, Shinsuke},
    journal={Computational Linguistics},
    volume={51},
    number={1},
    month={mar},
    year={2025},
    address={Cambridge, MA},
    url={https://aclanthology.org/2025.cl-1.5/},
    doi={10.1162/coli_a_00536},
    pages={139--190},
}

@article{wang2022text, 
    title={Text embeddings by weakly-supervised contrastive pre-training},
    author={Wang, Liang and Yang, Nan and Huang, Xiaolong and Jiao, Binxing and Yang, Linjun and Jiang, Daxin and Majumder, Rangan and Wei, Furu},
    journal={arXiv preprint arXiv:2212.03533},
    year={2022},
    url={https://arxiv.org/abs/2212.03533},
}

@article{nussbaum2025nomic, 
    title={Nomic Embed: {Training} a Reproducible Long Context Text Embedder},
    author={Zach Nussbaum and John Xavier Morris and Andriy Mulyar and Brandon Duderstadt},
    journal={Transactions on Machine Learning Research},
    issn={2835-8856},
    year={2025},
    url={https://openreview.net/forum?id=IPmzyQSiQE},
    note={Reproducibility Certification},
}

@inproceedings{chen-etal-2024-m3, 
    title={{M}3-Embedding: {Multi-Linguality,} Multi-Functionality, Multi-Granularity Text Embeddings Through Self-Knowledge Distillation},
    author={Chen, Jianlyu  and Xiao, Shitao  and Zhang, Peitian  and Luo, Kun  and Lian, Defu  and Liu, Zheng},
    editor={Ku, Lun-Wei  and Martins, Andre  and Srikumar, Vivek},
    booktitle={Findings of the Association for Computational Linguistics: ACL 2024},
    month={aug},
    year={2024},
    address={Bangkok, Thailand},
    publisher={Association for Computational Linguistics},
    url={https://aclanthology.org/2024.findings-acl.137/},
    doi={10.18653/v1/2024.findings-acl.137},
    pages={2318--2335},
}

@inproceedings{cer-etal-2017-semeval, 
    title={{S}em{E}val-2017 Task 1: {Semantic} Textual Similarity Multilingual and Crosslingual Focused Evaluation},
    author={Cer, Daniel  and Diab, Mona  and Agirre, Eneko  and Lopez-Gazpio, I{\~n}igo  and Specia, Lucia},
    editor={Bethard, Steven  and Carpuat, Marine  and Apidianaki, Marianna  and Mohammad, Saif M.  and Cer, Daniel  and Jurgens, David},
    booktitle={Proceedings of the 11th International Workshop on Semantic Evaluation ({S}em{E}val-2017)},
    month={aug},
    year={2017},
    address={Vancouver, Canada},
    publisher={Association for Computational Linguistics},
    url={https://aclanthology.org/S17-2001/},
    doi={10.18653/v1/S17-2001},
    pages={1--14},
}

@article{tversky1977features,
  title={Features of similarity.},
  author={Tversky, Amos},
  journal={Psychological review},
  volume={84},
  number={4},
  pages={327},
  year={1977},
  publisher={American Psychological Association}
}

@inproceedings{weller2026on, 
    title={On the Theoretical Limitations of Embedding-Based Retrieval},
    author={Orion Weller and Michael Boratko and Iftekhar Naim and Jinhyuk Lee},
    booktitle={The Fourteenth International Conference on Learning Representations},
    year={2026},
    url={https://openreview.net/forum?id=k9CzIvzfaA},
}

@inproceedings{muennighoff-etal-2023-mteb, 
    title={{MTEB}: {Massive} Text Embedding Benchmark},
    author={Muennighoff, Niklas  and Tazi, Nouamane  and Magne, Loic  and Reimers, Nils},
    editor={Vlachos, Andreas  and Augenstein, Isabelle},
    booktitle={Proceedings of the 17th Conference of the European Chapter of the Association for Computational Linguistics},
    month={may},
    year={2023},
    address={Dubrovnik, Croatia},
    publisher={Association for Computational Linguistics},
    url={https://aclanthology.org/2023.eacl-main.148/},
    doi={10.18653/v1/2023.eacl-main.148},
    pages={2014--2037},
}

@inproceedings{opitz-etal-2025-interpretable, 
    title={Interpretable Text Embeddings and Text Similarity Explanation: {A} Survey},
    author={Opitz, Juri  and Moeller, Lucas  and Michail, Andrianos  and Pad{\'o}, Sebastian  and Clematide, Simon},
    editor={Christodoulopoulos, Christos  and Chakraborty, Tanmoy  and Rose, Carolyn  and Peng, Violet},
    booktitle={Proceedings of the 2025 Conference on Empirical Methods in Natural Language Processing},
    month={nov},
    year={2025},
    address={Suzhou, China},
    publisher={Association for Computational Linguistics},
    url={https://aclanthology.org/2025.emnlp-main.1135/},
    doi={10.18653/v1/2025.emnlp-main.1135},
    pages={22303--22319},
    isbn={979-8-89176-332-6},
}

@article{d345368b-3e1e-3ffa-90ba-5afc4e854054, 
    issn={00278424, 10916490},
    url={http://www.jstor.org/stable/86426},
    author={Arthur E. Kennelly},
    journal={Proceedings of the National Academy of Sciences of the United States of America},
    number={10},
    pages={579--583},
    title={Adoption of the Meter-Kilogram-Mass-Second (M.K.S.) Absolute System of Practical Units by the International Electrotechnical Commission (I.E.C.), Bruxelles, June, 1935},
    urldate={2026-07-06},
    volume={21},
    year={1935},
}

@inproceedings{reimers-gurevych-2019-sentence, 
    title={Sentence-{BERT}: {Sentence} Embeddings using {S}iamese {BERT}-Networks},
    author={Reimers, Nils  and Gurevych, Iryna},
    editor={Inui, Kentaro  and Jiang, Jing  and Ng, Vincent  and Wan, Xiaojun},
    booktitle={Proceedings of the 2019 Conference on Empirical Methods in Natural Language Processing and the 9th International Joint Conference on Natural Language Processing (EMNLP-IJCNLP)},
    month={nov},
    year={2019},
    address={Hong Kong, China},
    publisher={Association for Computational Linguistics},
    url={https://aclanthology.org/D19-1410},
    doi={10.18653/v1/D19-1410},
    pages={3982--3992},
}

@book{greaves1647discourse, 
    author={John Greaves},
    title={A Discourse of the Romane Foot and Denarius; from whence, as from two principles, the measures and weights used by the ancients may be deduced},
    year={1647},
    address={London},
    publisher={William Lee},
    url={https://archive.org/search.php?query=external-identifier%3A%22urn%3Alcp%3Adiscourseofroman00grea%3Aepub%3A69f1c272-c650-4c74-bc4f-6acfd9418b8e%22},
}

@inproceedings{song2020mpnet,
 author = {Song, Kaitao and Tan, Xu and Qin, Tao and Lu, Jianfeng and Liu, Tie-Yan},
 booktitle = {Advances in Neural Information Processing Systems},
 editor = {H. Larochelle and M. Ranzato and R. Hadsell and M.F. Balcan and H. Lin},
 pages = {16857--16867},
 publisher = {Curran Associates, Inc.},
 title = {MPNet: Masked and Permuted Pre-training for Language Understanding},
 url = {https://proceedings.neurips.cc/paper_files/paper/2020/file/c3a690be93aa602ee2dc0ccab5b7b67e-Paper.pdf},
 volume = {33},
 year = {2020}
}

@inproceedings{wang2020minilm,
 author = {Wang, Wenhui and Wei, Furu and Dong, Li and Bao, Hangbo and Yang, Nan and Zhou, Ming},
 booktitle = {Advances in Neural Information Processing Systems},
 editor = {H. Larochelle and M. Ranzato and R. Hadsell and M.F. Balcan and H. Lin},
 pages = {5776--5788},
 publisher = {Curran Associates, Inc.},
 title = {MiniLM: Deep Self-Attention Distillation for Task-Agnostic Compression of Pre-Trained Transformers},
 url = {https://proceedings.neurips.cc/paper_files/paper/2020/file/3f5ee243547dee91fbd053c1c4a845aa-Paper.pdf},
 volume = {33},
 year = {2020}
}

@inproceedings{reimers-gurevych-2020-making,
    title = "Making Monolingual Sentence Embeddings Multilingual using Knowledge Distillation",
    author = "Reimers, Nils  and
      Gurevych, Iryna",
    editor = "Webber, Bonnie  and
      Cohn, Trevor  and
      He, Yulan  and
      Liu, Yang",
    booktitle = "Proceedings of the 2020 Conference on Empirical Methods in Natural Language Processing (EMNLP)",
    month = nov,
    year = "2020",
    address = "Online",
    publisher = "Association for Computational Linguistics",
    url = "https://aclanthology.org/2020.emnlp-main.365/",
    doi = "10.18653/v1/2020.emnlp-main.365",
    pages = "4512--4525"
}

@inproceedings{feng-etal-2022-language,
    title = "Language-agnostic {BERT} Sentence Embedding",
    author = "Feng, Fangxiaoyu  and
      Yang, Yinfei  and
      Cer, Daniel  and
      Arivazhagan, Naveen  and
      Wang, Wei",
    editor = "Muresan, Smaranda  and
      Nakov, Preslav  and
      Villavicencio, Aline",
    booktitle = "Proceedings of the 60th Annual Meeting of the Association for Computational Linguistics (Volume 1: Long Papers)",
    month = may,
    year = "2022",
    address = "Dublin, Ireland",
    publisher = "Association for Computational Linguistics",
    url = "https://aclanthology.org/2022.acl-long.62/",
    doi = "10.18653/v1/2022.acl-long.62",
    pages = "878--891"
}

@article{wang2024multilinguale5textembeddings,
  title={Multilingual e5 text embeddings: A technical report},
  author={Wang, Liang and Yang, Nan and Huang, Xiaolong and Yang, Linjun and Majumder, Rangan and Wei, Furu},
  journal={arXiv preprint arXiv:2402.05672},
  year={2024}
}

@inproceedings{10.1145/3626772.3657878,
author = {Xiao, Shitao and Liu, Zheng and Zhang, Peitian and Muennighoff, Niklas and Lian, Defu and Nie, Jian-Yun},
title = {C-Pack: Packed Resources For General Chinese Embeddings},
year = {2024},
isbn = {9798400704314},
publisher = {Association for Computing Machinery},
address = {New York, NY, USA},
url = {https://doi.org/10.1145/3626772.3657878},
doi = {10.1145/3626772.3657878},
booktitle = {Proceedings of the 47th International ACM SIGIR Conference on Research and Development in Information Retrieval},
pages = {641–649},
numpages = {9},
location = {Washington DC, USA},
series = {SIGIR '24}
}

@online{lee2024mxbai,
  title={Open Source Strikes Bread - New Fluffy Embedding Model},
  author={Sean Lee and Aamir Shakir and Darius Koenig and Julius Lipp},
  year={2024},
  url={https://www.mixedbread.com/blog/mxbai-embed-large-v1},
}

@inproceedings{li2023angle,
    title = "{A}o{E}: Angle-optimized Embeddings for Semantic Textual Similarity",
    author = "Li, Xianming  and
      Li, Jing",
    editor = "Ku, Lun-Wei  and
      Martins, Andre  and
      Srikumar, Vivek",
    booktitle = "Proceedings of the 62nd Annual Meeting of the Association for Computational Linguistics (Volume 1: Long Papers)",
    month = aug,
    year = "2024",
    address = "Bangkok, Thailand",
    publisher = "Association for Computational Linguistics",
    url = "https://aclanthology.org/2024.acl-long.101/",
    doi = "10.18653/v1/2024.acl-long.101",
    pages = "1825--1839"
}

@article{awasthy2025granite,
  title={Granite embedding r2 models},
  author={Awasthy, Parul and Trivedi, Aashka and Li, Yulong and Doshi, Meet and Bhat, Riyaz and Kumar, Vishwajeet and Yang, Yushu and Iyer, Bhavani and Daniels, Abraham and Murthy, Rudra and others},
  journal={arXiv preprint arXiv:2508.21085},
  year={2025}
}

@inproceedings{warner-etal-2025-smarter,
    title = "Smarter, Better, Faster, Longer: A Modern Bidirectional Encoder for Fast, Memory Efficient, and Long Context Finetuning and Inference",
    author = {Warner, Benjamin  and
      Chaffin, Antoine  and
      Clavi{\'e}, Benjamin  and
      Weller, Orion  and
      Hallstr{\"o}m, Oskar  and
      Taghadouini, Said  and
      Gallagher, Alexis  and
      Biswas, Raja  and
      Ladhak, Faisal  and
      Aarsen, Tom  and
      Adams, Griffin Thomas  and
      Howard, Jeremy  and
      Poli, Iacopo},
    editor = "Che, Wanxiang  and
      Nabende, Joyce  and
      Shutova, Ekaterina  and
      Pilehvar, Mohammad Taher",
    booktitle = "Proceedings of the 63rd Annual Meeting of the Association for Computational Linguistics (Volume 1: Long Papers)",
    month = jul,
    year = "2025",
    address = "Vienna, Austria",
    publisher = "Association for Computational Linguistics",
    url = "https://aclanthology.org/2025.acl-long.127/",
    doi = "10.18653/v1/2025.acl-long.127",
    pages = "2526--2547",
    ISBN = "979-8-89176-251-0"
}

@article{awasthy2026graniteembeddingmultilingualr2,
  title={Granite Embedding Multilingual R2 Models},
  author={Awasthy, Parul and Trivedi, Aashka and Yang, Yushu and Barker, Ken and Li, Yulong and Iyer, Bhavani and Franz, Martin and Bross, Juergen and Doshi, Meet and Kumar, Vishwajeet and others},
  journal={arXiv preprint arXiv:2605.13521},
  year={2026}
}

@article{sourty2026denseonlateonfullyopen,
  title={DenseOn with the LateOn: Fully Open Dense and Late-Interaction Models for Multilingual, Long-Context, and Code Search},
  author={Sourty, Rapha{\"e}l and Chaffin, Antoine and Junior, Paulo Roberto Moura and Chatelain, Am{\'e}lie},
  journal={arXiv preprint arXiv:2607.27178},
  year={2026}
}

@misc{schechtervera2025embeddinggemma,
      title={EmbeddingGemma: Powerful and Lightweight Text Representations}, 
      author={Henrique Schechter Vera and Sahil Dua and Biao Zhang and Daniel Salz and Ryan Mullins and Sindhu Raghuram Panyam and Sara Smoot and Iftekhar Naim and Joe Zou and Feiyang Chen and Daniel Cer and Alice Lisak and Min Choi and Lucas Gonzalez and Omar Sanseviero and Glenn Cameron and Ian Ballantyne and Kat Black and Kaifeng Chen and Weiyi Wang and Zhe Li and Gus Martins and Jinhyuk Lee and Mark Sherwood and Juyeong Ji and Renjie Wu and Jingxiao Zheng and Jyotinder Singh and Abheesht Sharma and Divyashree Sreepathihalli and Aashi Jain and Adham Elarabawy and AJ Co and Andreas Doumanoglou and Babak Samari and Ben Hora and Brian Potetz and Dahun Kim and Enrique Alfonseca and Fedor Moiseev and Feng Han and Frank Palma Gomez and Gustavo Hernández Ábrego and Hesen Zhang and Hui Hui and Jay Han and Karan Gill and Ke Chen and Koert Chen and Madhuri Shanbhogue and Michael Boratko and Paul Suganthan and Sai Meher Karthik Duddu and Sandeep Mariserla and Setareh Ariafar and Shanfeng Zhang and Shijie Zhang and Simon Baumgartner and Sonam Goenka and Steve Qiu and Tanmaya Dabral and Trevor Walker and Vikram Rao and Waleed Khawaja and Wenlei Zhou and Xiaoqi Ren and Ye Xia and Yichang Chen and Yi-Ting Chen and Zhe Dong and Zhongli Ding and Francesco Visin and Gaël Liu and Jiageng Zhang and Kathleen Kenealy and Michelle Casbon and Ravin Kumar and Thomas Mesnard and Zach Gleicher and Cormac Brick and Olivier Lacombe and Adam Roberts and Qin Yin and Yunhsuan Sung and Raphael Hoffmann and Tris Warkentin and Armand Joulin and Tom Duerig and Mojtaba Seyedhosseini},
      year={2025},
      eprint={2509.20354},
      archivePrefix={arXiv},
      primaryClass={cs.CL},
      url={https://arxiv.org/abs/2509.20354}, 
}

@misc{microsoft2026harrier,
  title        = {Microsoft Open-Sources Industry-Leading Embedding Model},
  author       = {Huang, Xiaolong and Wang, Liang and Wei, Furu and Lu, Jingwen and Risvik, Knut and Li, Jason},
  year         = {2026},
  month        = apr,
  howpublished = {Bing Blogs},
  url          = {https://blogs.bing.com/search/April-2026/Microsoft-Open-Sources-Industry-Leading-Embedding-Model},
  note         = {Models: \url{https://huggingface.co/microsoft/harrier-oss-v1-0.6b}}
}

\appendix

\section{Appendix}
\label{sec:appendix}

\begin{table*}
\centering
\adjustbox{max width=\textwidth}{
\begin{tabular}{lll}
\toprule
Model & Measure plot & Conversion plot \\
\midrule
\multicolumn{3}{l}{\textit{Sentence-Transformers baselines}} \\
\texttt{sentence-transformers/all-mpnet-base-v2} & Section~\ref{sec:s1} & Section~\ref{sec:s2} \\
\texttt{sentence-transformers/all-MiniLM-L6-v2} & Figure~\ref{fig:all_MiniLM_L6_v2-eo} & Figure~\ref{fig:all_MiniLM_L6_v2-al} \\
\texttt{sentence-transformers/paraphrase-multilingual-mpnet-base-v2} & Figure~\ref{fig:paraphrase_multilingual_mpnet_base_v2-eo} & Figure~\ref{fig:paraphrase_multilingual_mpnet_base_v2-al} \\
\texttt{sentence-transformers/LaBSE} & Figure~\ref{fig:LaBSE-eo} & Figure~\ref{fig:LaBSE-al} \\
\midrule
\multicolumn{3}{l}{\textit{Contrastive BERT-style encoders}} \\
\texttt{intfloat/e5-large-v2} & Figure~\ref{fig:e5_large_v2-eo} & Figure~\ref{fig:e5_large_v2-al} \\
\texttt{intfloat/multilingual-e5-base} & Figure~\ref{fig:multilingual_e5_base-eo} & Figure~\ref{fig:multilingual_e5_base-al} \\
\texttt{intfloat/multilingual-e5-large} & Figure~\ref{fig:multilingual_e5_large-eo} & Figure~\ref{fig:multilingual_e5_large-al} \\
\texttt{intfloat/multilingual-e5-large-instruct} & Figure~\ref{fig:multilingual_e5_large_instruct-eo} & Figure~\ref{fig:multilingual_e5_large_instruct-al} \\
\texttt{BAAI/bge-large-en-v1.5} & Figure~\ref{fig:bge_large_en_v1.5-eo} & Figure~\ref{fig:bge_large_en_v1.5-al} \\
\texttt{BAAI/bge-m3} & Figure~\ref{fig:bge_m3-eo} & Figure~\ref{fig:bge_m3-al} \\
\texttt{mixedbread-ai/mxbai-embed-large-v1} & Figure~\ref{fig:mxbai_embed_large_v1-eo} & Figure~\ref{fig:mxbai_embed_large_v1-al} \\
\texttt{ibm-granite/granite-embedding-107m-multilingual} & Figure~\ref{fig:granite_embedding_107m_multilingual-eo} & Figure~\ref{fig:granite_embedding_107m_multilingual-al} \\
\texttt{ibm-granite/granite-embedding-278m-multilingual} & Figure~\ref{fig:granite_embedding_278m_multilingual-eo} & Figure~\ref{fig:granite_embedding_278m_multilingual-al} \\
\texttt{nomic-ai/nomic-embed-text-v1.5} & Figure~\ref{fig:nomic_embed_text_v1.5-eo} & Figure~\ref{fig:nomic_embed_text_v1.5-al} \\
\midrule
\multicolumn{3}{l}{\textit{ModernBERT-based encoders}} \\
\texttt{ibm-granite/granite-embedding-english-r2} & Figure~\ref{fig:granite_embedding_english_r2-eo} & Figure~\ref{fig:granite_embedding_english_r2-al} \\
\texttt{ibm-granite/granite-embedding-97m-multilingual-r2} & Figure~\ref{fig:granite_embedding_97m_multilingual_r2-eo} & Figure~\ref{fig:granite_embedding_97m_multilingual_r2-al} \\
\texttt{ibm-granite/granite-embedding-311m-multilingual-r2} & Figure~\ref{fig:granite_embedding_311m_multilingual_r2-eo} & Figure~\ref{fig:granite_embedding_311m_multilingual_r2-al} \\
\texttt{lightonai/DenseOn} & Figure~\ref{fig:DenseOn-eo} & Figure~\ref{fig:DenseOn-al} \\
\midrule
\multicolumn{3}{l}{\textit{Models derived from LLMs}} \\
\texttt{Qwen/Qwen3-Embedding-0.6B} & Section~\ref{sec:s1} & Section~\ref{sec:s2} \\
\texttt{Qwen/Qwen3-Embedding-4B} & Figure~\ref{fig:Qwen3_Embedding_4B-eo} & Figure~\ref{fig:Qwen3_Embedding_4B-al} \\
\texttt{Qwen/Qwen3-Embedding-8B} & Figure~\ref{fig:Qwen3_Embedding_8B-eo} & Figure~\ref{fig:Qwen3_Embedding_8B-al} \\
\texttt{microsoft/harrier-oss-v1-270m} & Figure~\ref{fig:harrier_oss_v1_270m-eo} & Figure~\ref{fig:harrier_oss_v1_270m-al} \\
\texttt{microsoft/harrier-oss-v1-0.6b} & Figure~\ref{fig:harrier_oss_v1_0.6b-eo} & Figure~\ref{fig:harrier_oss_v1_0.6b-al} \\
\texttt{google/embeddinggemma-300m} & Figure~\ref{fig:embeddinggemma_300m-eo} & Figure~\ref{fig:embeddinggemma_300m-al} \\
\bottomrule
\end{tabular}}
\caption{Hugging Face identifiers of all evaluated models, grouped as in the text, with links to their measurement and conversion plots.}
\label{tab:models}
\end{table*}

\subsection{Embedding Models}
\label{app:models}

We evaluate 24 embedding models. Table~\ref{tab:models} lists their Hugging Face identifiers. We group them into four families by backbone and training approach.

\paragraph{Sentence-Transformers baselines.}
\model{all-mpnet-base-v2} and \model{all-MiniLM-L6-v2} are English models from the Sentence-Transformers library \citep{reimers-gurevych-2019-sentence}. They are based on MPNet \citep{song2020mpnet} and MiniLM \citep{wang2020minilm} and are fine-tuned on a large collection of sentence pairs. Both are widely used and serve as classic baselines. We also include two multilingual models that are distributed through the same library. \model{paraphrase-multilingual-mpnet-base-v2} is trained with knowledge distillation: a multilingual student model learns to place a sentence and its translation at the same position as an English teacher model \citep{reimers-gurevych-2020-making}. \model{LaBSE} is a multilingual BERT model trained on translation pairs, originally for finding parallel sentences \citep{feng-etal-2022-language}.

\paragraph{Contrastive BERT-style encoders.}
Most models in our set follow a similar recipe: a BERT-like encoder is trained contrastively on large amounts of weakly paired text and then fine-tuned on labeled data. This group includes \model{e5-large-v2} \citep{wang2022text} and its multilingual versions \model{multilingual-e5-base}, \model{multilingual-e5-large}, and \model{multilingual-e5-large-instruct}, where the last one takes a task instruction as part of the input \citep{wang2024multilinguale5textembeddings}. It further includes \model{bge-large-en-v1.5} \citep{10.1145/3626772.3657878} and its multilingual successor \model{bge-m3} \citep{chen-etal-2024-m3}; \model{mxbai-embed-large-v1} \citep{lee2024mxbai}, which is trained with the AnglE objective \citep{li2023angle}; and the first release of the multilingual Granite embedding models (\model{granite-embedding-107m-multilingual} and \model{granite-embedding-278m-multilingual}), which are based on XLM-RoBERTa \citep{awasthy2025granite}. The model \model{nomic-embed-text-v1.5} \citep{nussbaum2025nomic} also belongs to this group, but uses a modified BERT architecture that supports long inputs.

\paragraph{ModernBERT-based encoders.}
Several recent encoders are built on ModernBERT \citep{warner-etal-2025-smarter}, an updated BERT architecture with a longer context window. These are \model{granite-embedding-english-r2} \citep{awasthy2025granite}, the multilingual \model{granite-embedding-97m-multilingual-r2} and \model{granite-embedding-311m-multilingual-r2} \citep{awasthy2026graniteembeddingmultilingualr2}, and \model{DenseOn} \citep{sourty2026denseonlateonfullyopen}, an English retrieval model trained on openly released data.

\paragraph{Models derived from LLMs.}
The last group starts from pre-trained decoder language models. \model{Qwen3-Embedding} \citep{zhang2025qwen3} is built on the Qwen3 LLMs. To test the effect of model size in a focused way, we include the 0.6B, 4B, and 8B variants. \model{harrier-oss-v1} \citep{microsoft2026harrier} is a multilingual decoder-only model family from Microsoft. The two variants we use (\model{harrier-oss-v1-270m} and \model{harrier-oss-v1-0.6b}, based on the Gemma 3 and Qwen3 architectures, respectively) are trained contrastively and with knowledge distillation from larger embedding models. The model \model{embeddinggemma-300m} \citep{schechtervera2025embeddinggemma} is also based on Gemma 3, but it first turns the LLM into an encoder-decoder model and then uses only the encoder. It therefore reads the input text in both directions, like the BERT-style models above.

\subsection{Measure Plots}
\label{app:measure plots}
\begin{figure*}
    \centering
    \includegraphics[width=1.0\linewidth]{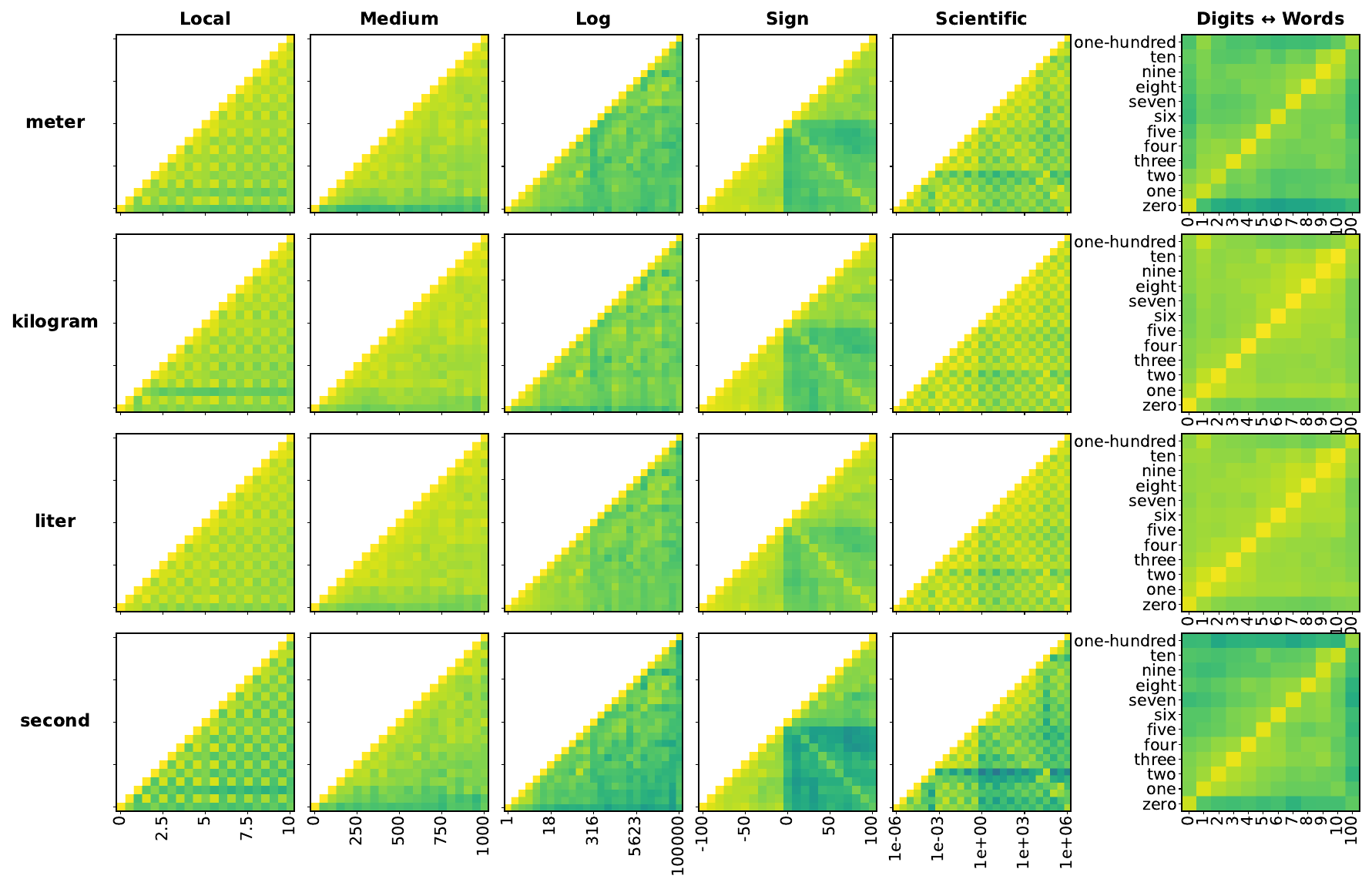}
    \caption{\texttt{all-MiniLM-L6-v2}. For more information see caption in Figure~\ref{fig:overview-plot-mini-qwen}.}
    \label{fig:all_MiniLM_L6_v2-eo}
\end{figure*}

\begin{figure*}
    \centering
    \includegraphics[width=1.0\linewidth]{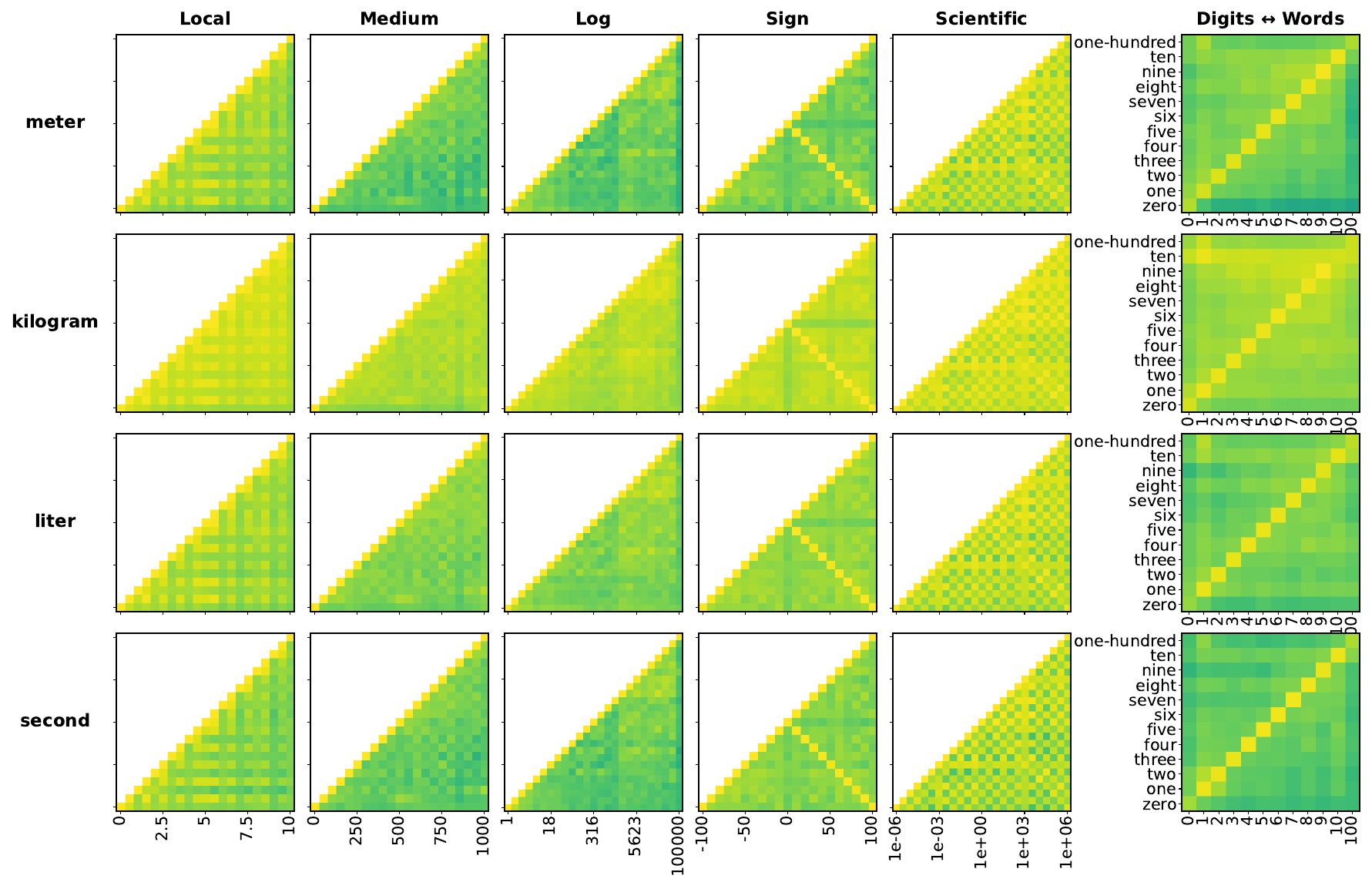}
    \caption{\texttt{paraphrase-multilingual-mpnet-base-v2}. For more information see caption in Figure~\ref{fig:overview-plot-mini-qwen}.}
    \label{fig:paraphrase_multilingual_mpnet_base_v2-eo}
\end{figure*}

\begin{figure*}
    \centering
    \includegraphics[width=1.0\linewidth]{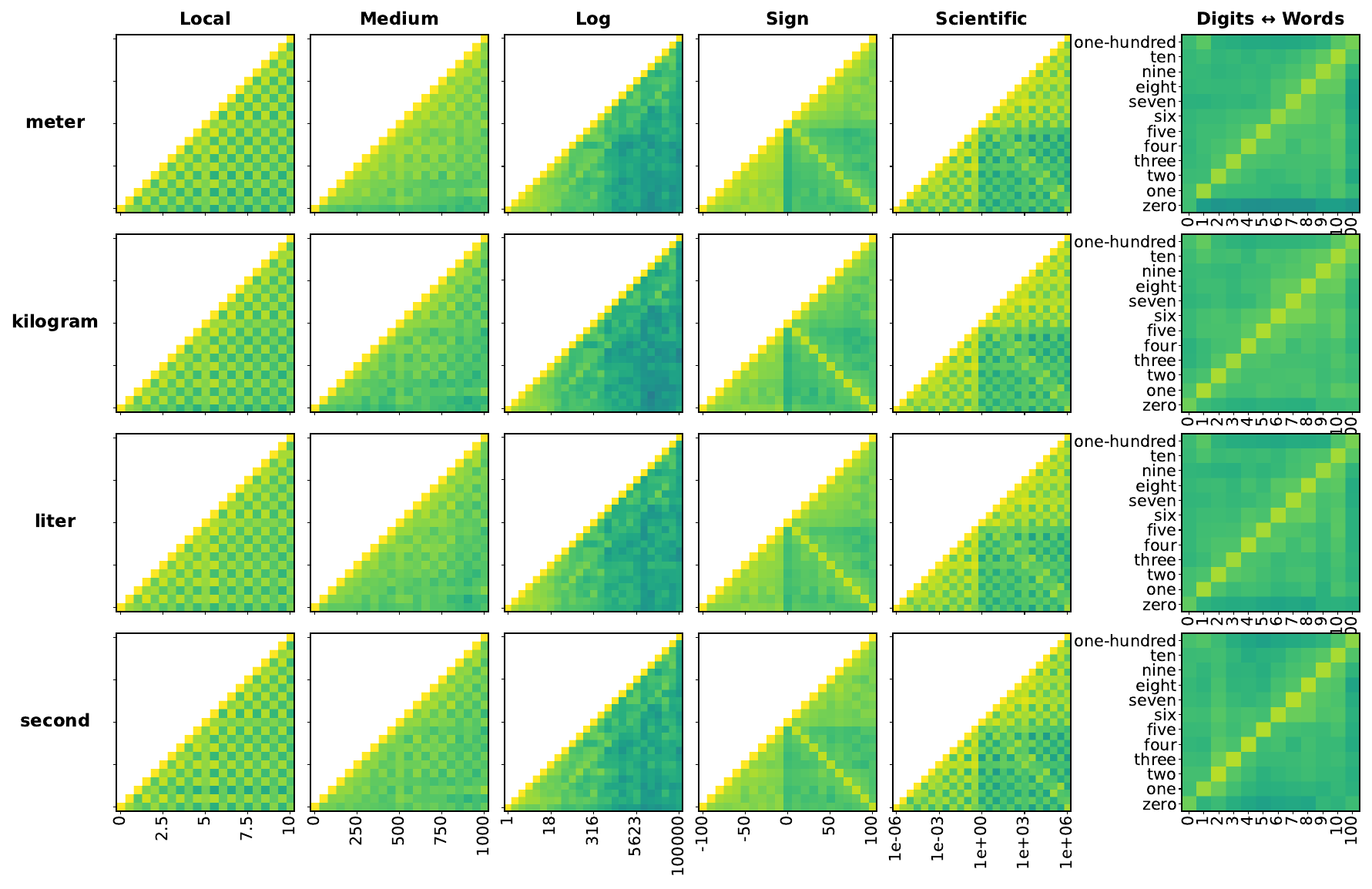}
    \caption{\texttt{LaBSE}. For more information see caption in Figure~\ref{fig:overview-plot-mini-qwen}.}
    \label{fig:LaBSE-eo}
\end{figure*}

\begin{figure*}
    \centering
    \includegraphics[width=1.0\linewidth]{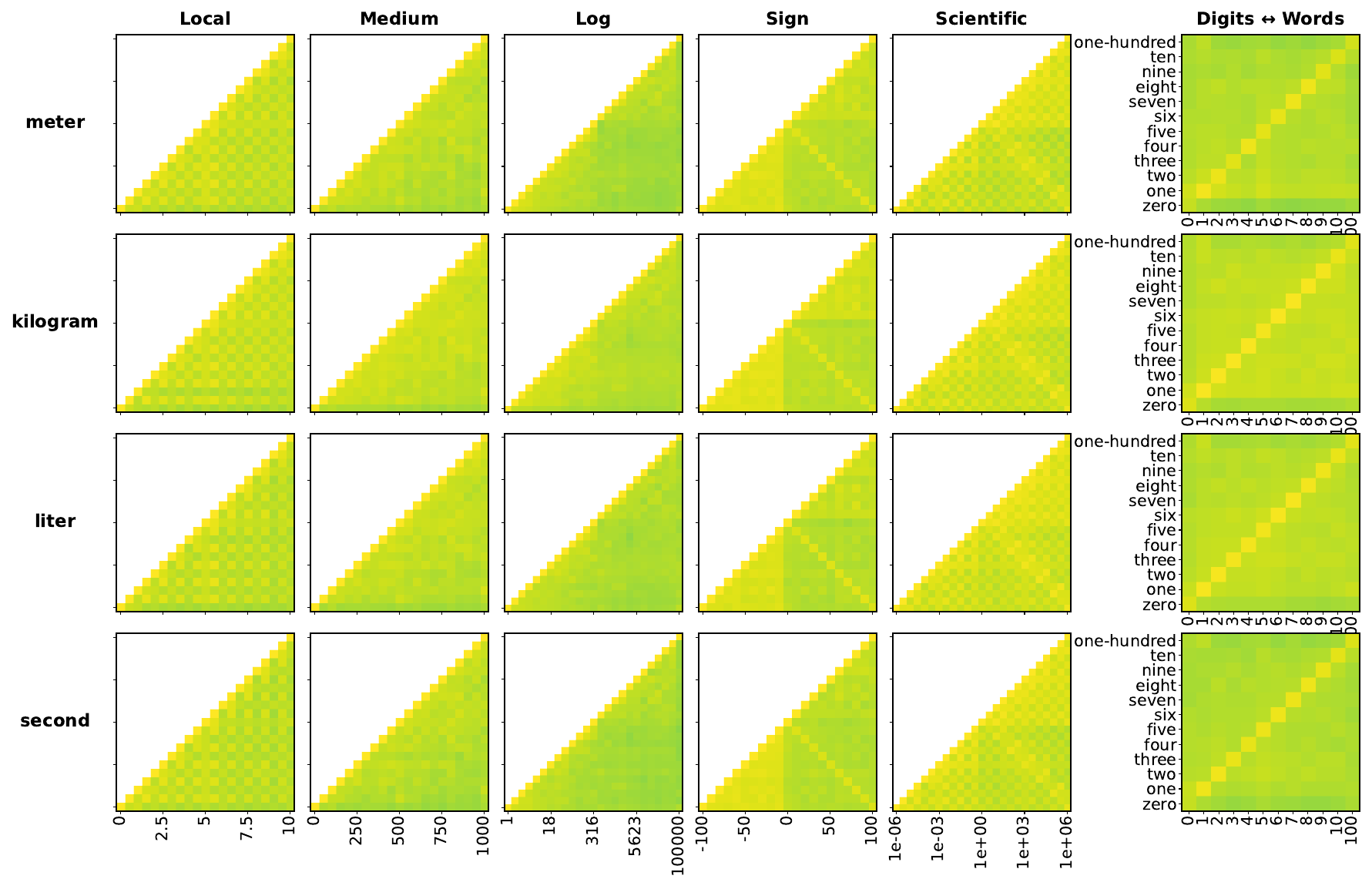}
    \caption{\texttt{e5-large-v2}. For more information see caption in Figure~\ref{fig:overview-plot-mini-qwen}.}
    \label{fig:e5_large_v2-eo}
\end{figure*}

\begin{figure*}
    \centering
    \includegraphics[width=1.0\linewidth]{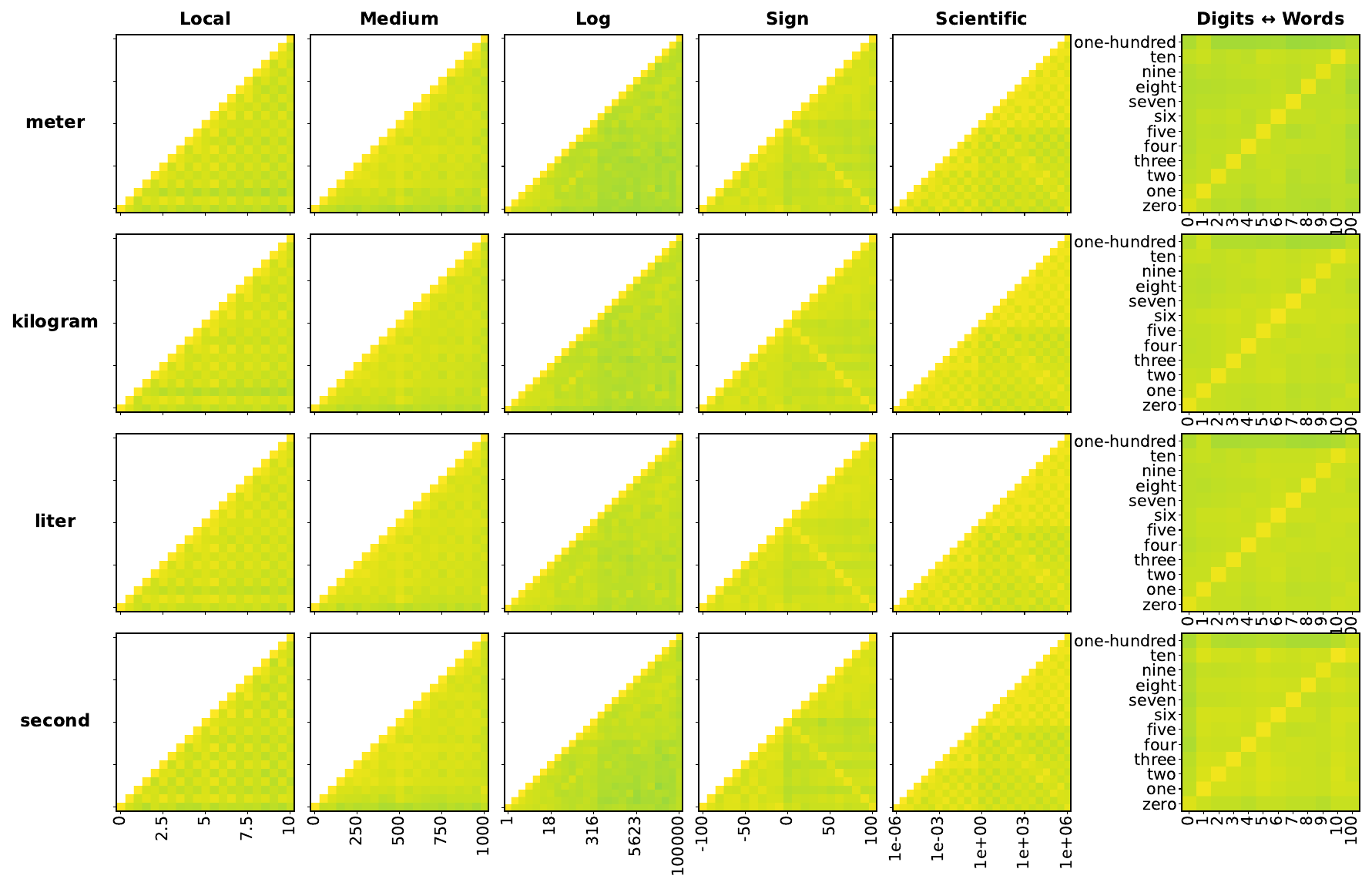}
    \caption{\texttt{multilingual-e5-base}. For more information see caption in Figure~\ref{fig:overview-plot-mini-qwen}.}
    \label{fig:multilingual_e5_base-eo}
\end{figure*}

\begin{figure*}
    \centering
    \includegraphics[width=1.0\linewidth]{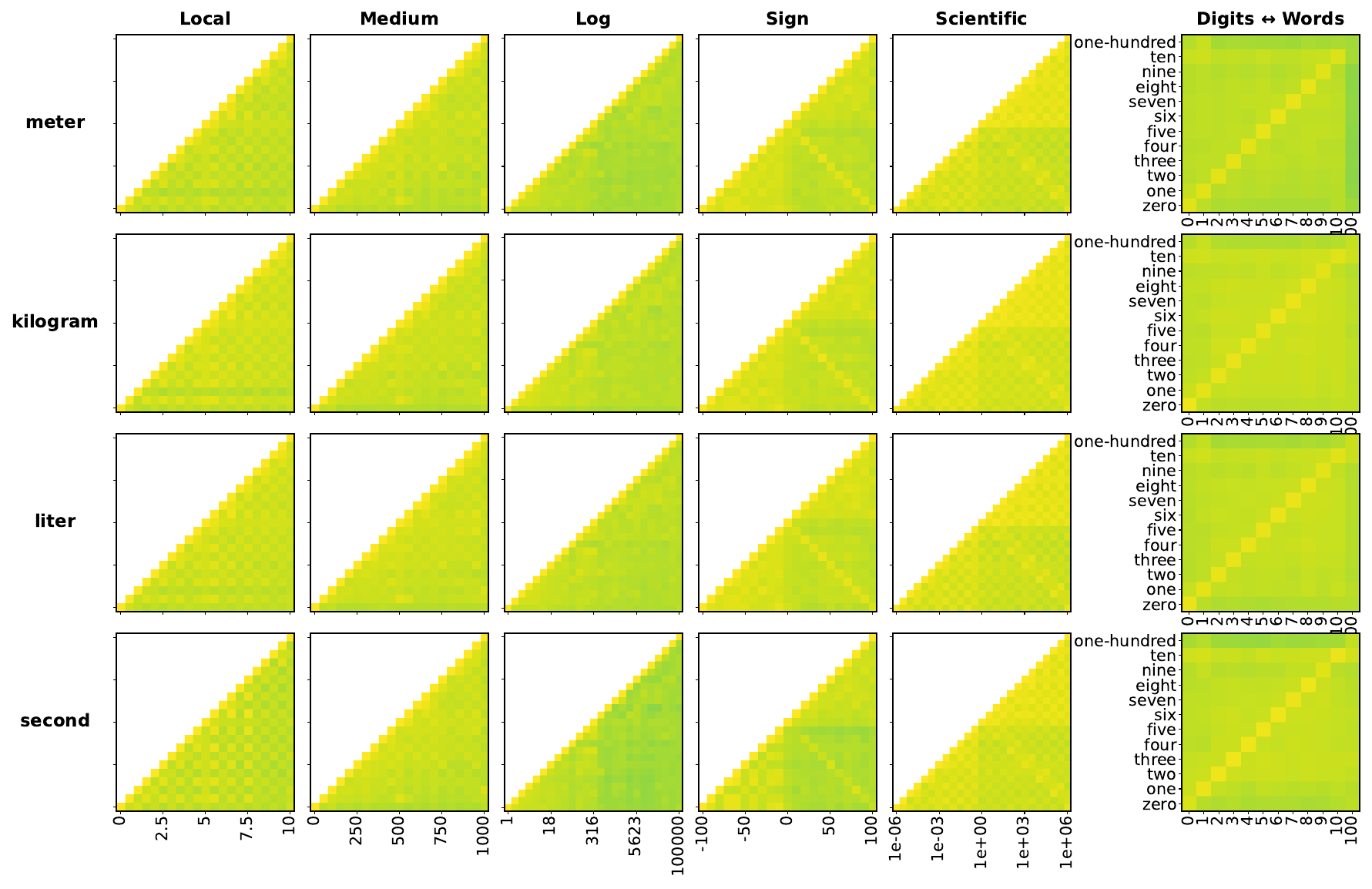}
    \caption{\texttt{multilingual-e5-large}. For more information see caption in Figure~\ref{fig:overview-plot-mini-qwen}.}
    \label{fig:multilingual_e5_large-eo}
\end{figure*}

\begin{figure*}
    \centering
    \includegraphics[width=1.0\linewidth]{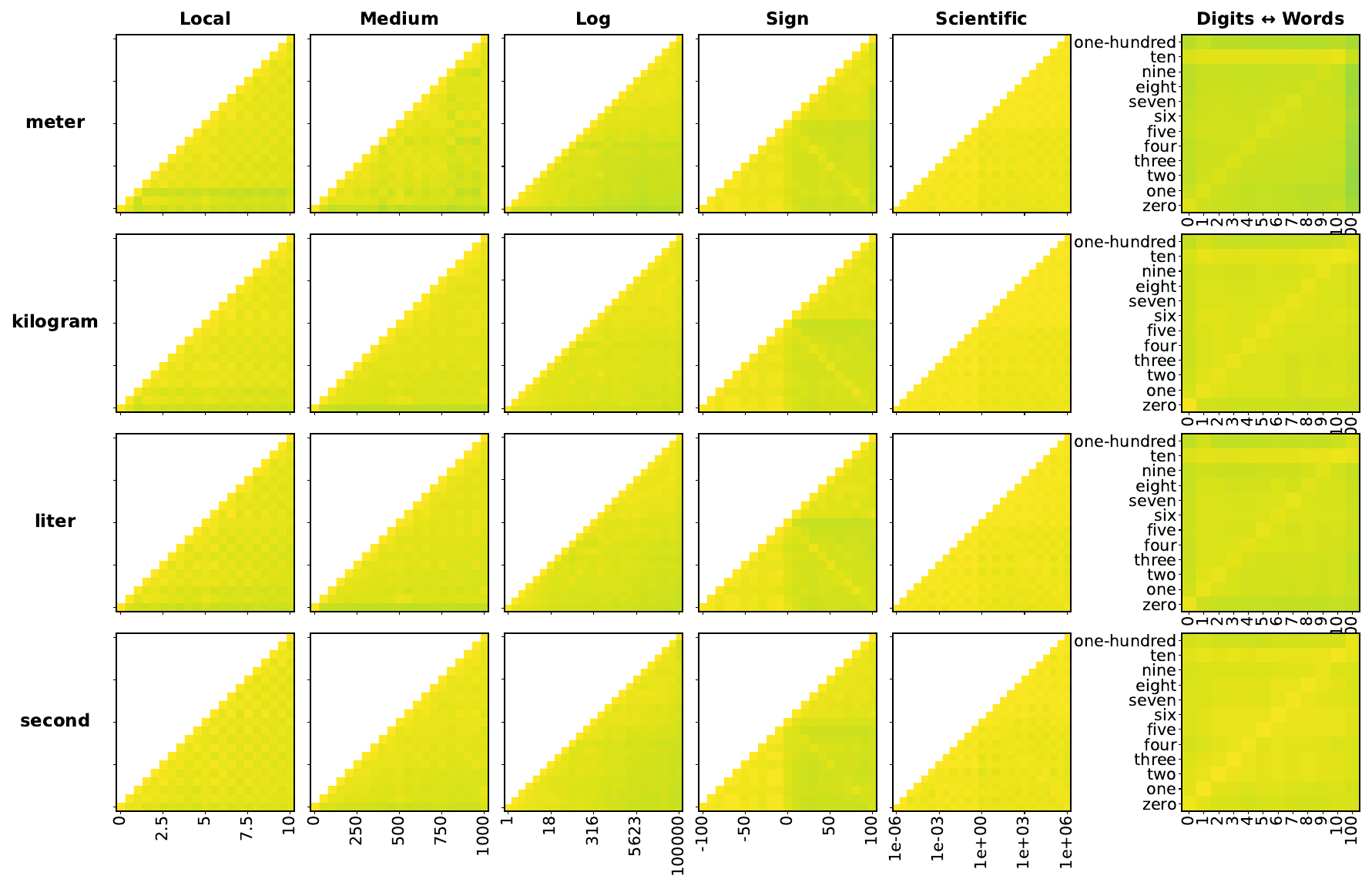}
    \caption{\texttt{multilingual-e5-large-instruct}. For more information see caption in Figure~\ref{fig:overview-plot-mini-qwen}.}
    \label{fig:multilingual_e5_large_instruct-eo}
\end{figure*}

\begin{figure*}
    \centering
    \includegraphics[width=1.0\linewidth]{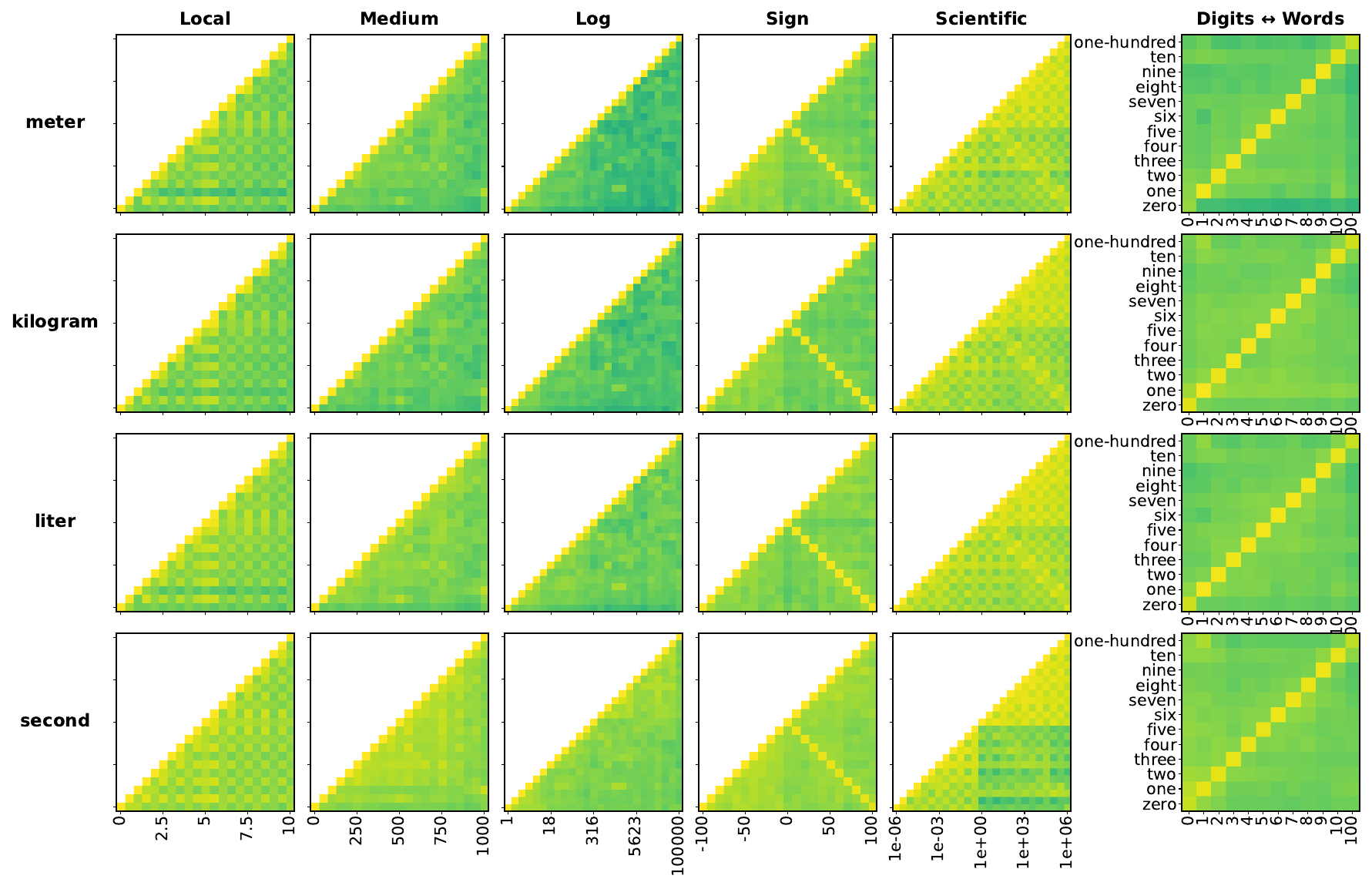}
    \caption{\texttt{bge-large-en-v1.5}. For more information see caption in Figure~\ref{fig:overview-plot-mini-qwen}.}
    \label{fig:bge_large_en_v1.5-eo}
\end{figure*}

\begin{figure*}
    \centering
    \includegraphics[width=1.0\linewidth]{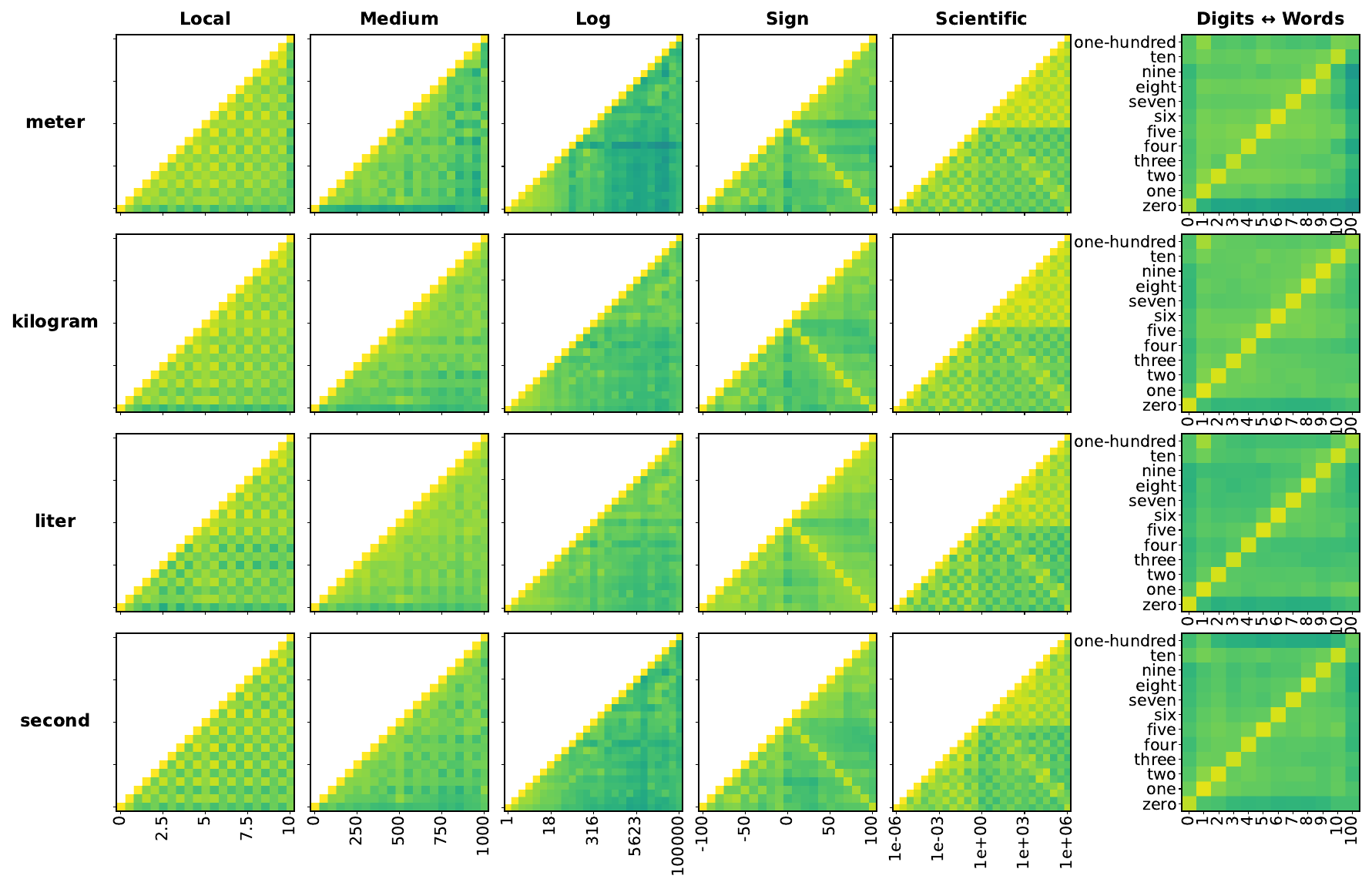}
    \caption{\texttt{bge-m3}. For more information see caption in Figure~\ref{fig:overview-plot-mini-qwen}.}
    \label{fig:bge_m3-eo}
\end{figure*}

\begin{figure*}
    \centering
    \includegraphics[width=1.0\linewidth]{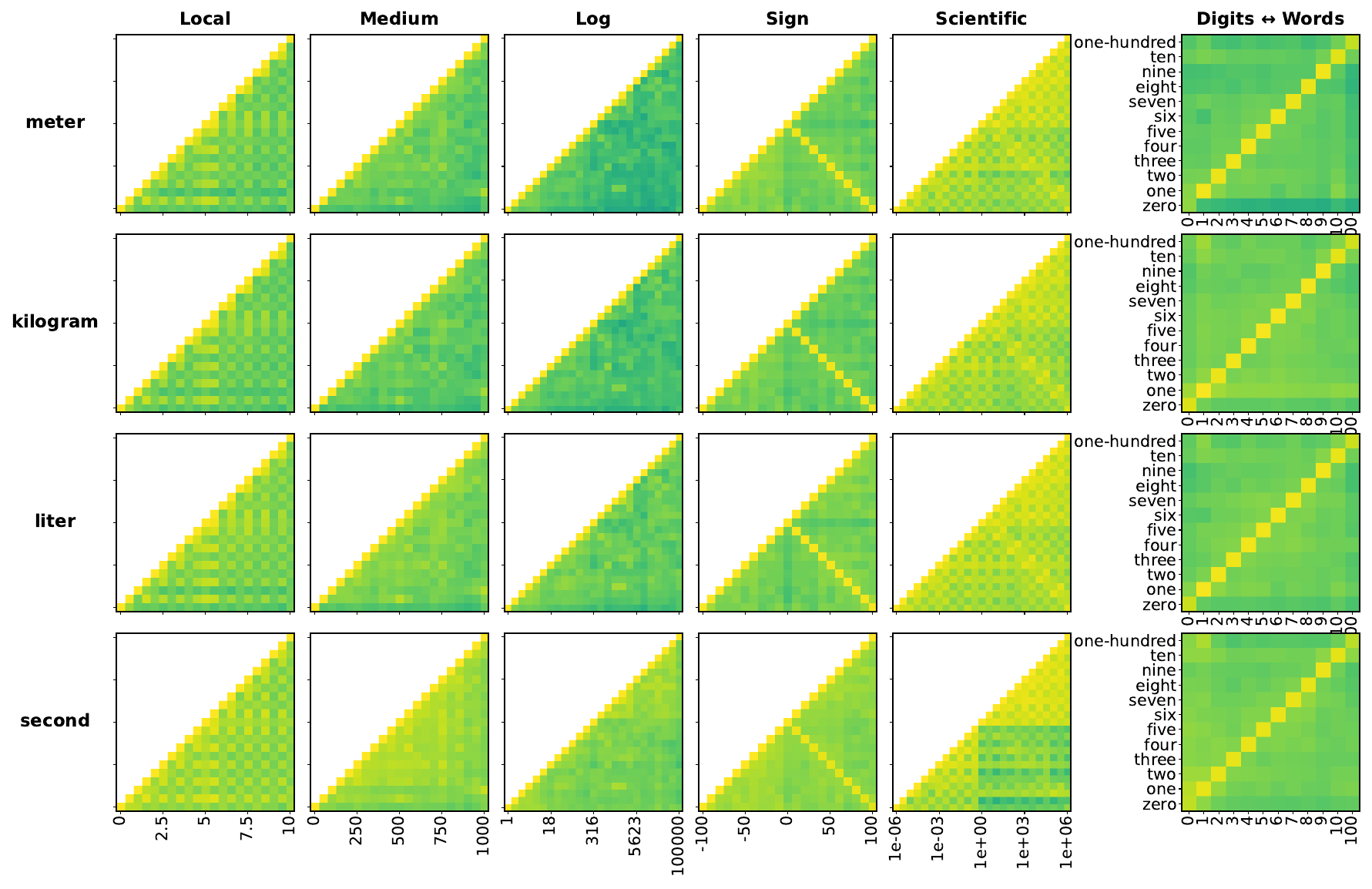}
    \caption{\texttt{mxbai-embed-large-v1}. For more information see caption in Figure~\ref{fig:overview-plot-mini-qwen}.}
    \label{fig:mxbai_embed_large_v1-eo}
\end{figure*}

\begin{figure*}
    \centering
    \includegraphics[width=1.0\linewidth]{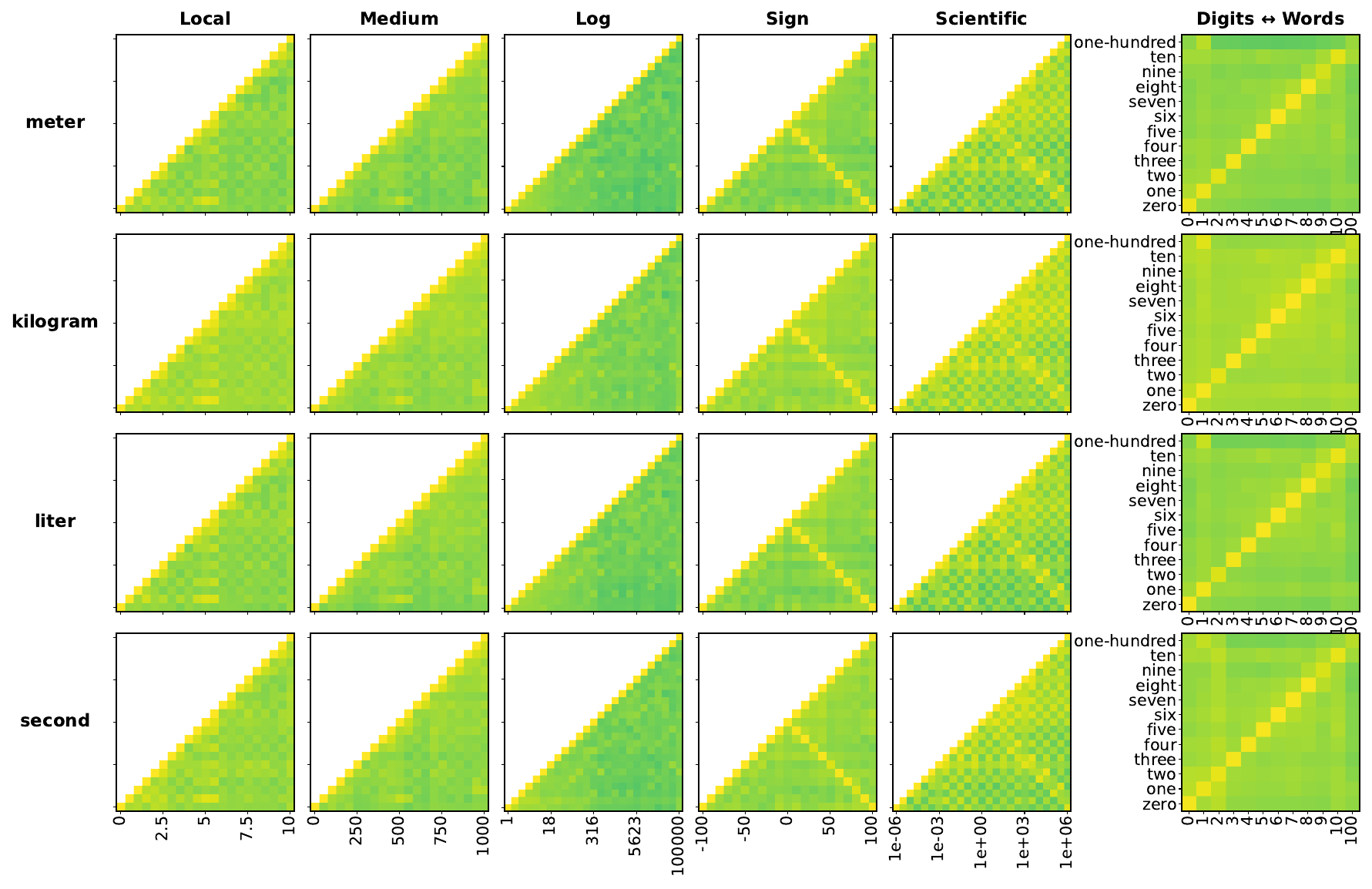}
    \caption{\texttt{granite-embedding-107m-multilingual}. For more information see caption in Figure~\ref{fig:overview-plot-mini-qwen}.}
    \label{fig:granite_embedding_107m_multilingual-eo}
\end{figure*}

\begin{figure*}
    \centering
    \includegraphics[width=1.0\linewidth]{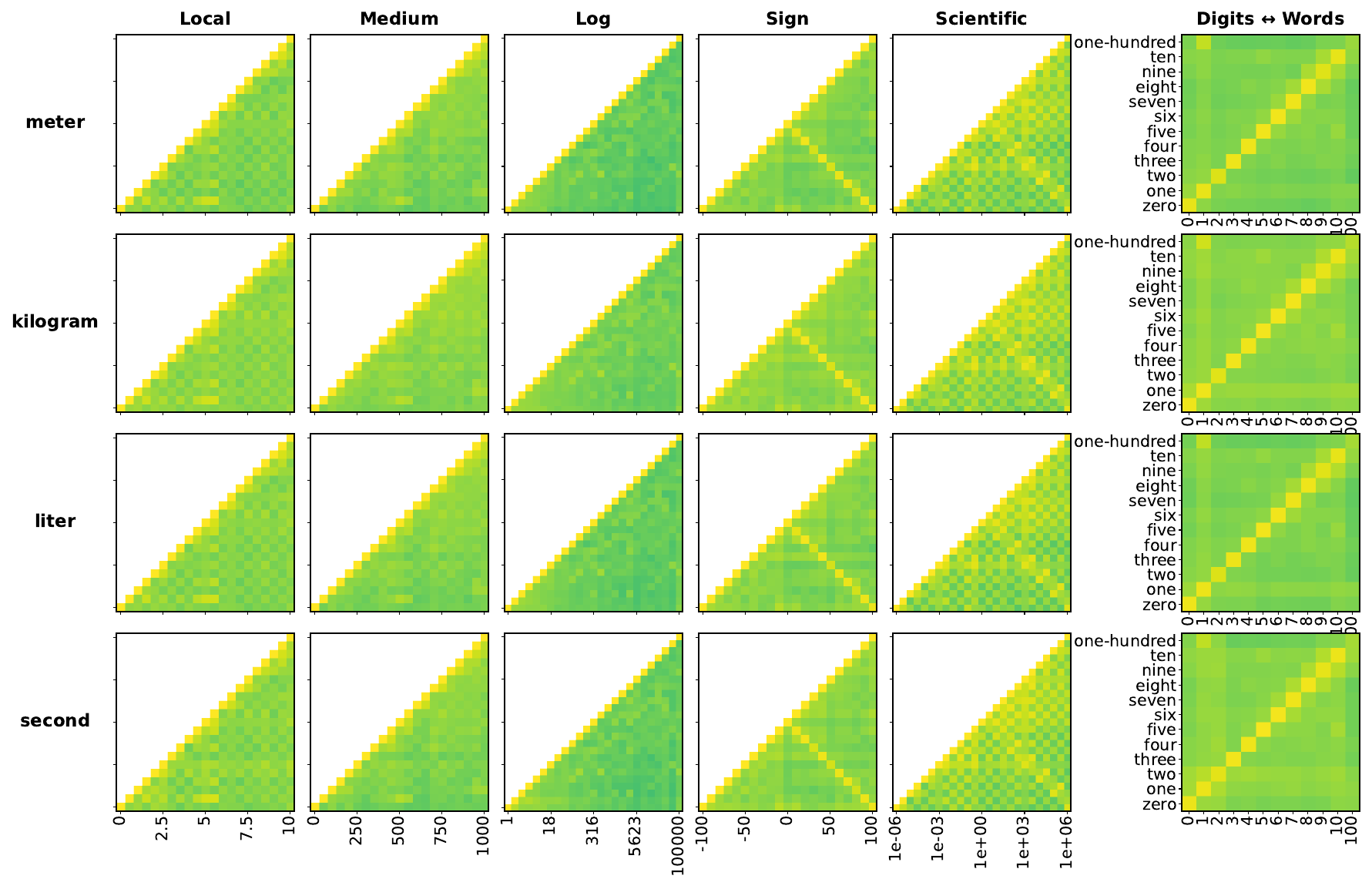}
    \caption{\texttt{granite-embedding-278m-multilingual}. For more information see caption in Figure~\ref{fig:overview-plot-mini-qwen}.}
    \label{fig:granite_embedding_278m_multilingual-eo}
\end{figure*}

\begin{figure*}
    \centering
    \includegraphics[width=1.0\linewidth]{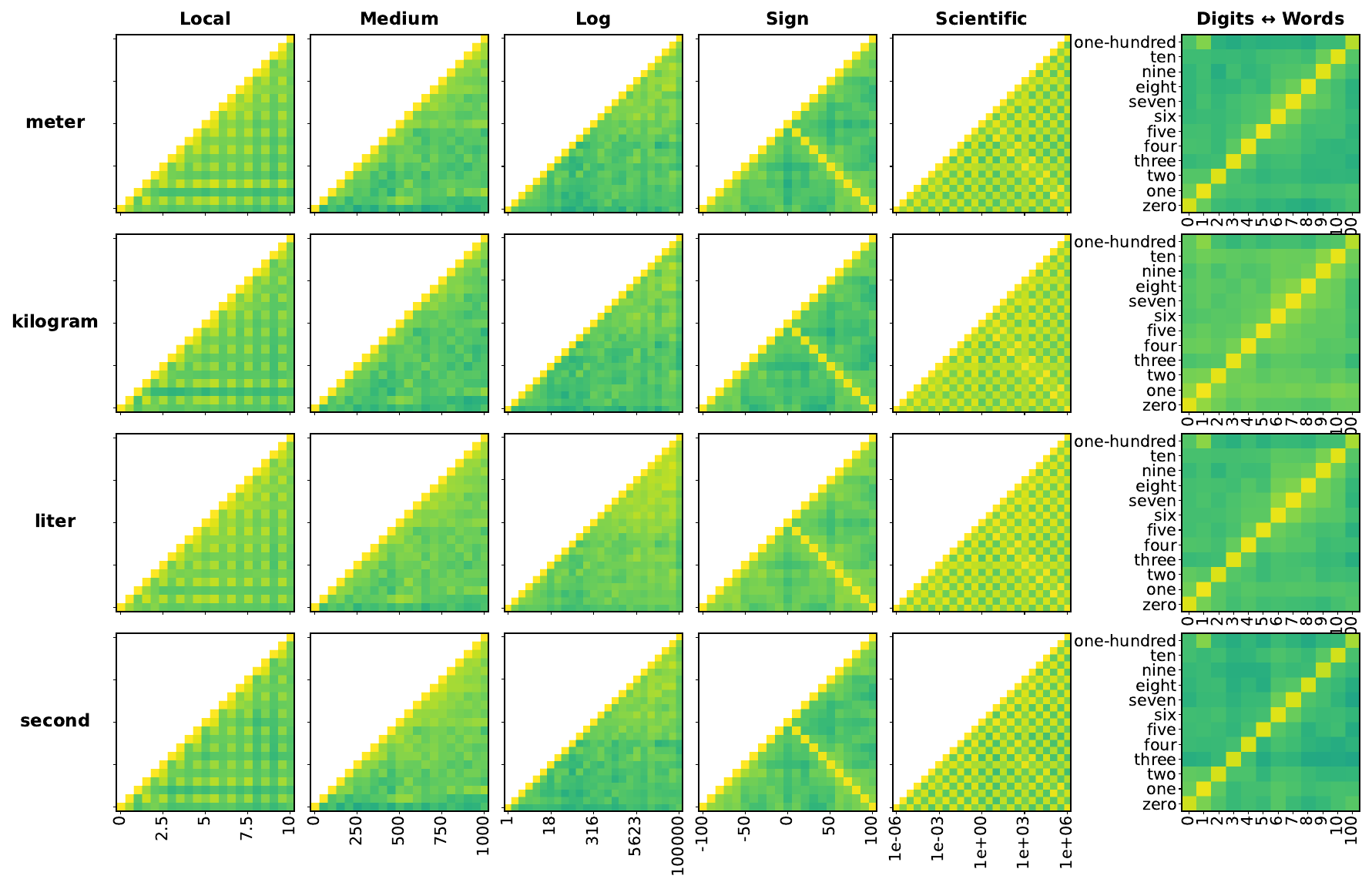}
    \caption{\texttt{nomic-embed-text-v1.5}. For more information see caption in Figure~\ref{fig:overview-plot-mini-qwen}.}
    \label{fig:nomic_embed_text_v1.5-eo}
\end{figure*}

\begin{figure*}
    \centering
    \includegraphics[width=1.0\linewidth]{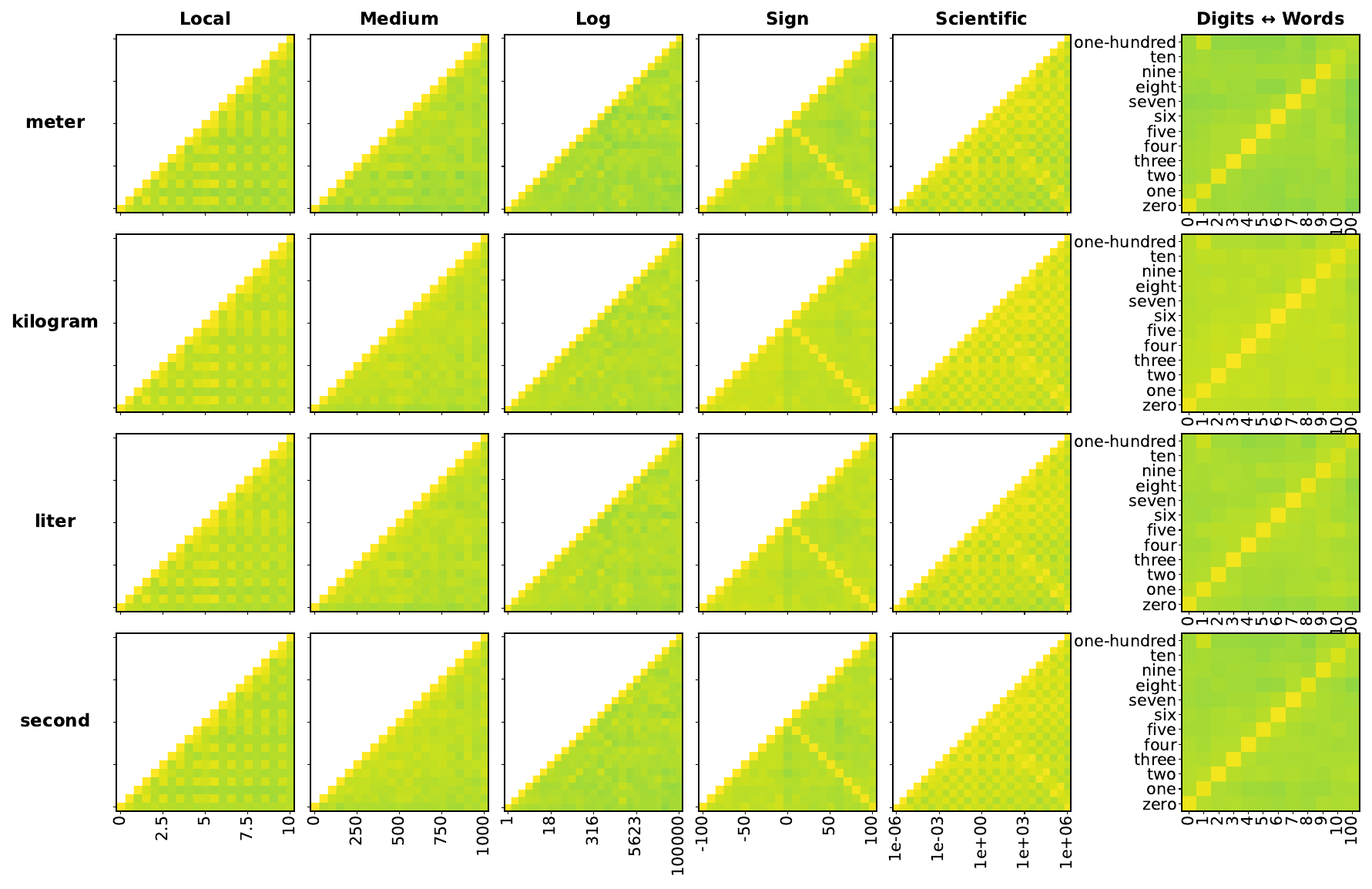}
    \caption{\texttt{granite-embedding-english-r2}. For more information see caption in Figure~\ref{fig:overview-plot-mini-qwen}.}
    \label{fig:granite_embedding_english_r2-eo}
\end{figure*}

\begin{figure*}
    \centering
    \includegraphics[width=1.0\linewidth]{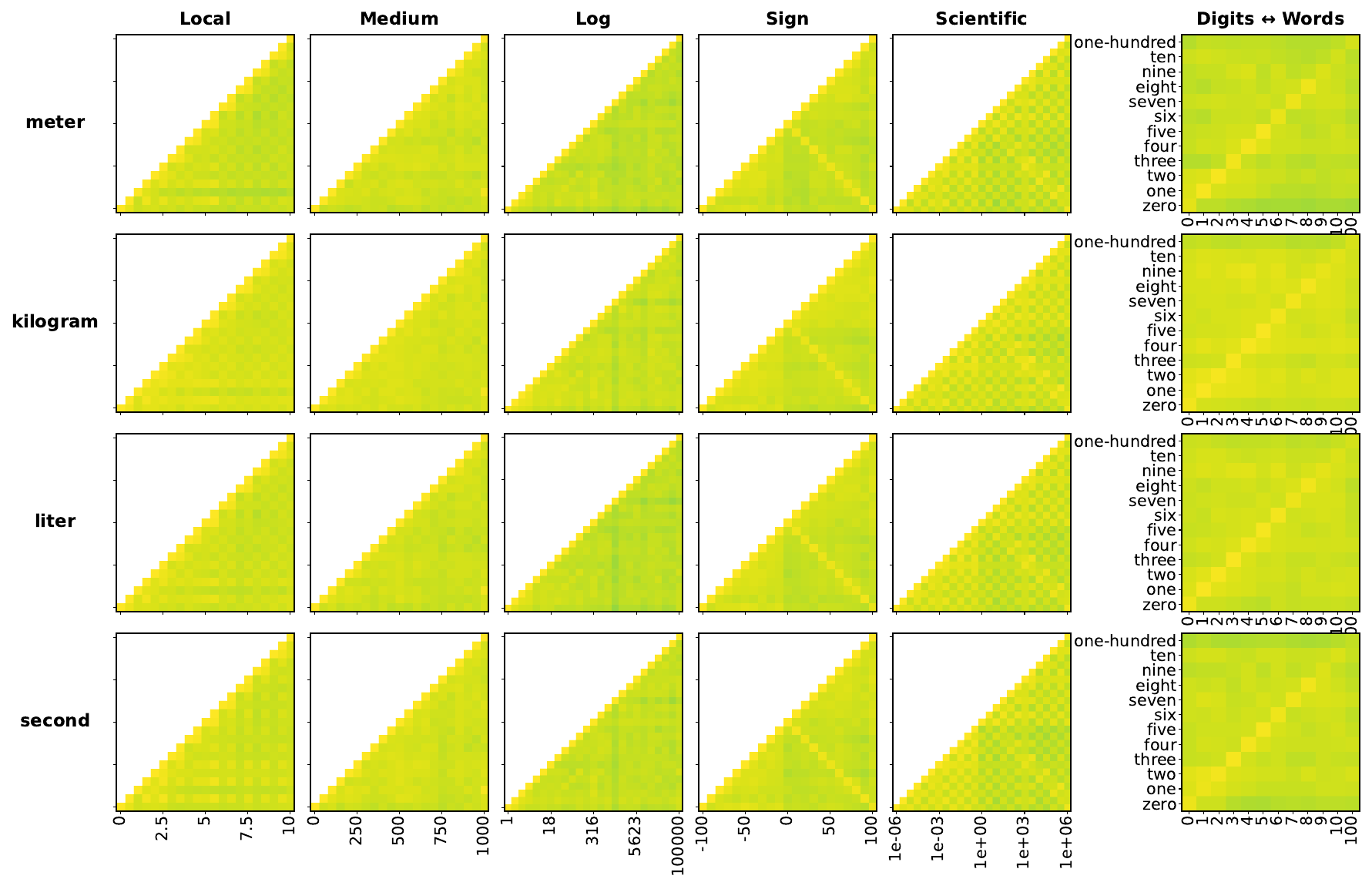}
    \caption{\texttt{granite-embedding-97m-multilingual-r2}. For more information see caption in Figure~\ref{fig:overview-plot-mini-qwen}.}
    \label{fig:granite_embedding_97m_multilingual_r2-eo}
\end{figure*}

\begin{figure*}
    \centering
    \includegraphics[width=1.0\linewidth]{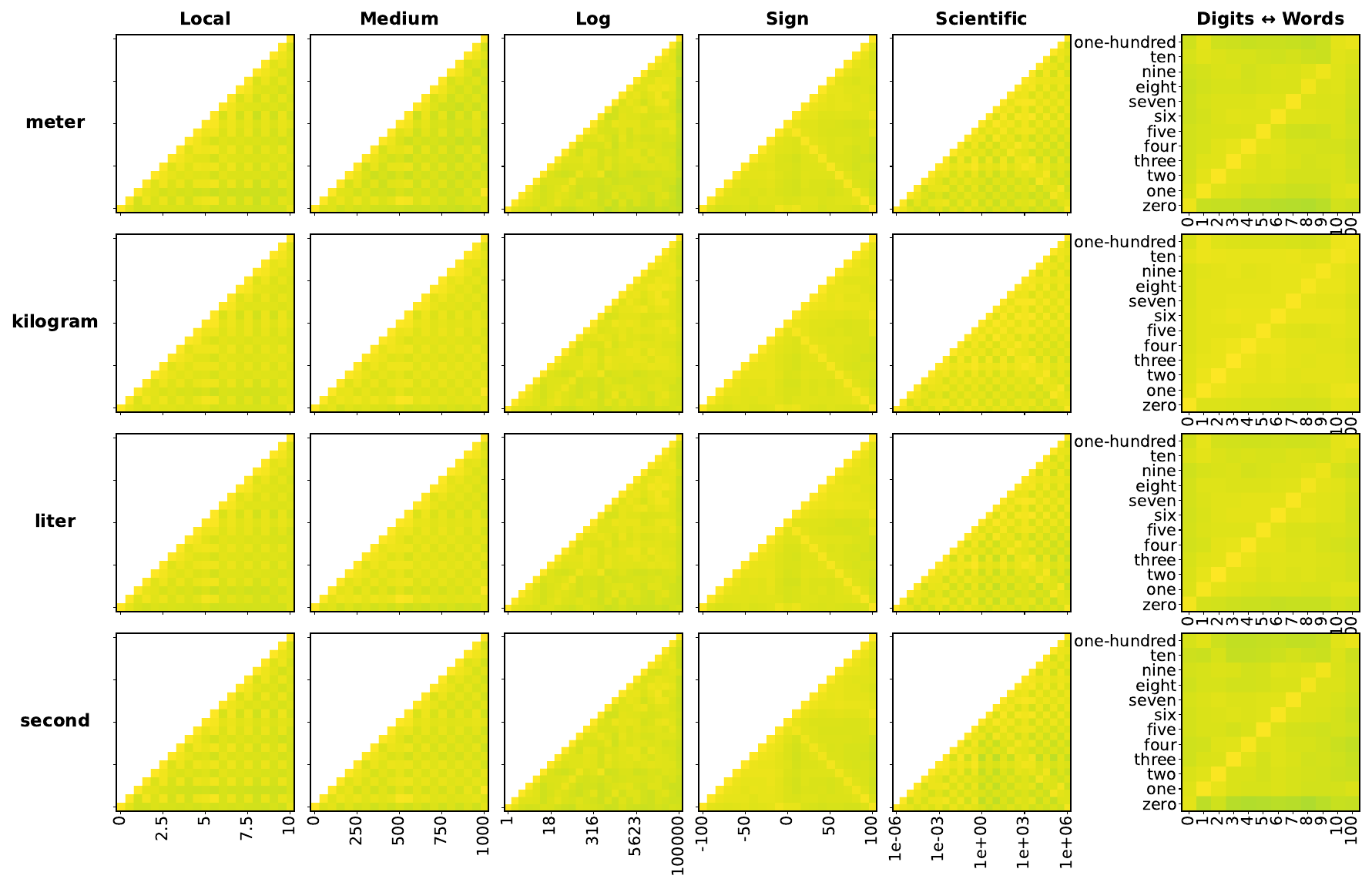}
    \caption{\texttt{granite-embedding-311m-multilingual-r2}. For more information see caption in Figure~\ref{fig:overview-plot-mini-qwen}.}
    \label{fig:granite_embedding_311m_multilingual_r2-eo}
\end{figure*}

\begin{figure*}
    \centering
    \includegraphics[width=1.0\linewidth]{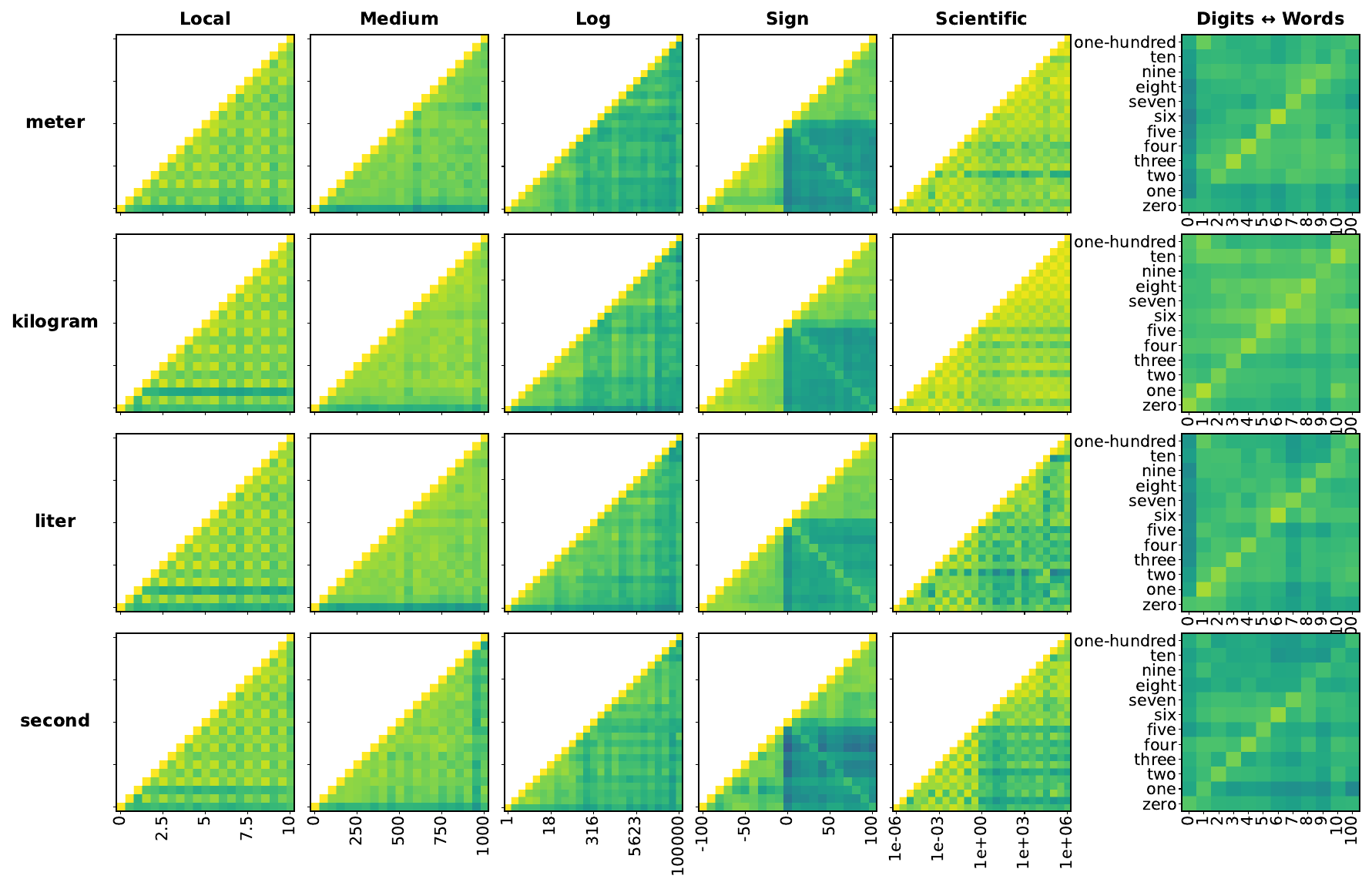}
    \caption{\texttt{DenseOn}. For more information see caption in Figure~\ref{fig:overview-plot-mini-qwen}.}
    \label{fig:DenseOn-eo}
\end{figure*}

\begin{figure*}
    \centering
    \includegraphics[width=1.0\linewidth]{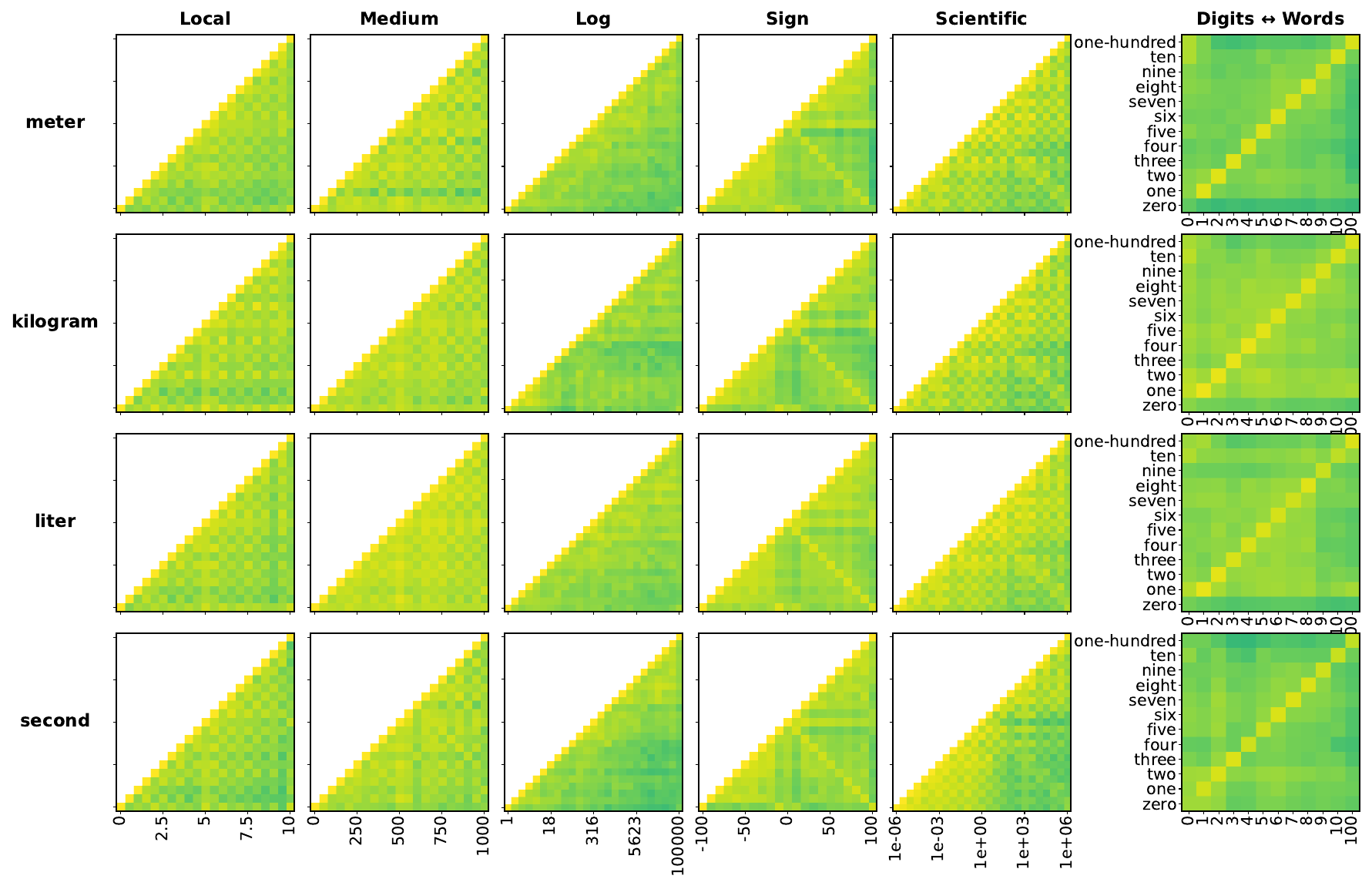}
    \caption{\texttt{Qwen3-Embedding-4B}. For more information see caption in Figure~\ref{fig:overview-plot-mini-qwen}.}
    \label{fig:Qwen3_Embedding_4B-eo}
\end{figure*}

\begin{figure*}
    \centering
    \includegraphics[width=1.0\linewidth]{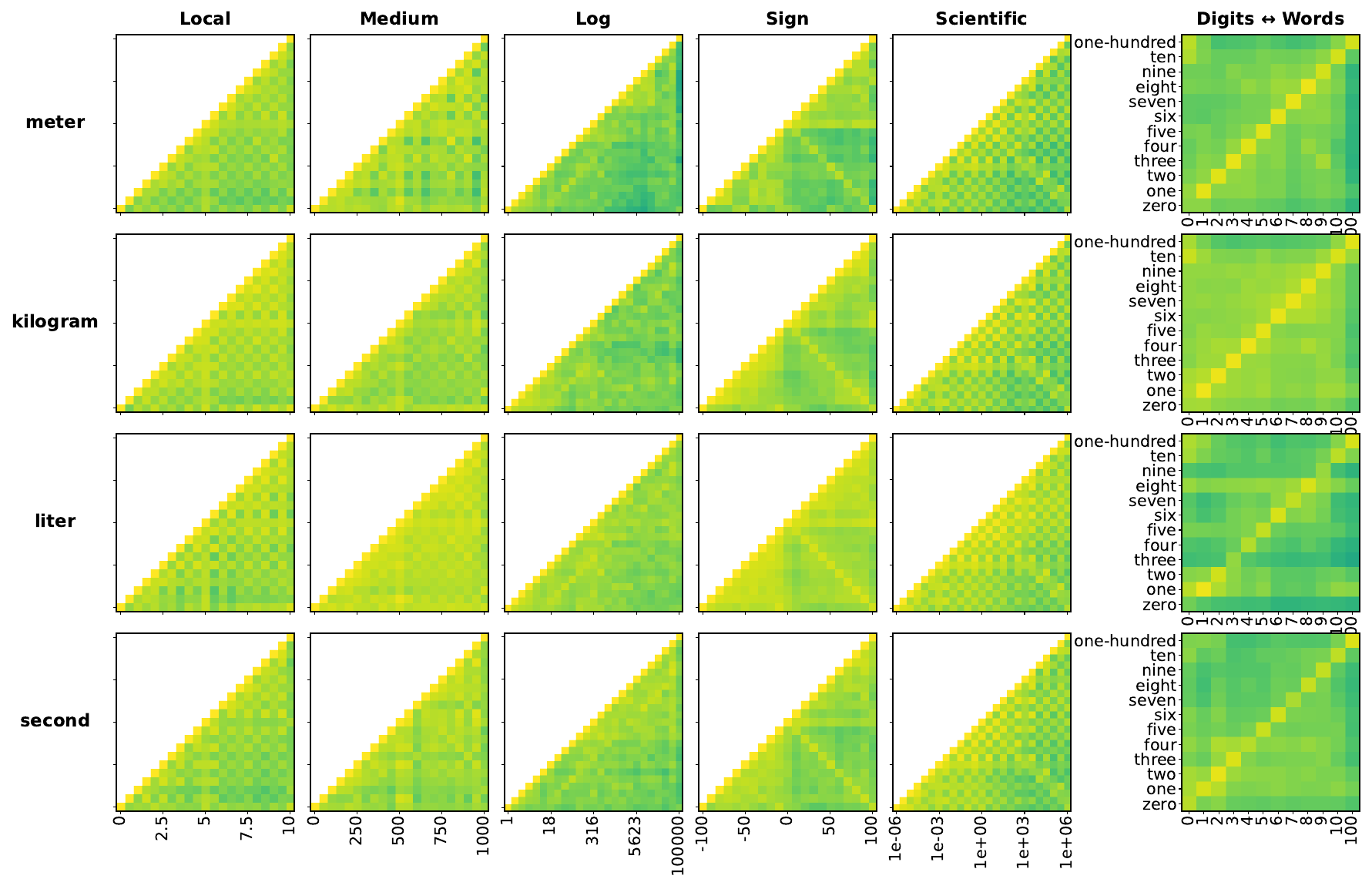}
    \caption{\texttt{Qwen3-Embedding-8B}. For more information see caption in Figure~\ref{fig:overview-plot-mini-qwen}.}
    \label{fig:Qwen3_Embedding_8B-eo}
\end{figure*}

\begin{figure*}
    \centering
    \includegraphics[width=1.0\linewidth]{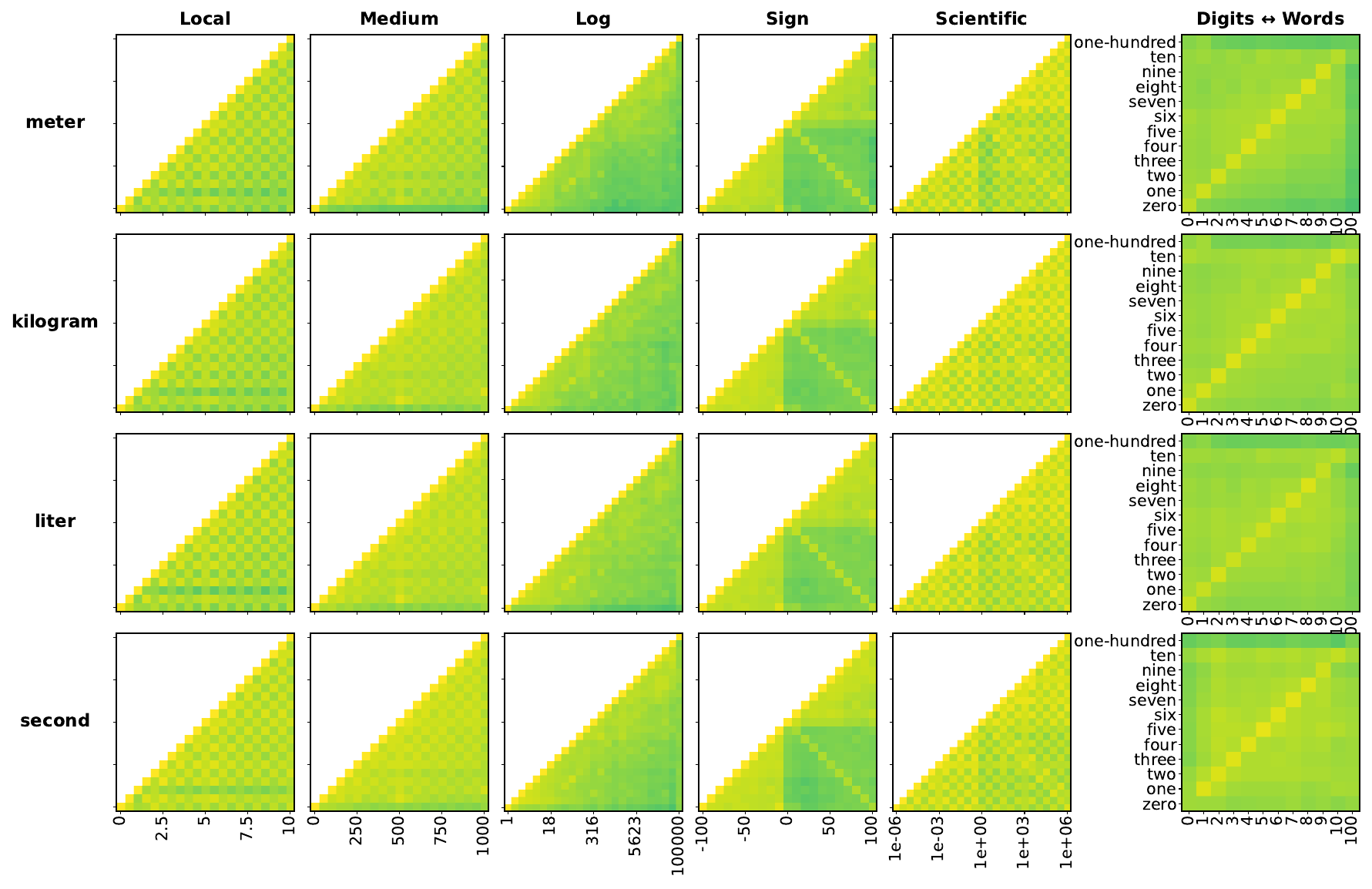}
    \caption{\texttt{harrier-oss-v1-270m}. For more information see caption in Figure~\ref{fig:overview-plot-mini-qwen}.}
    \label{fig:harrier_oss_v1_270m-eo}
\end{figure*}

\begin{figure*}
    \centering
    \includegraphics[width=1.0\linewidth]{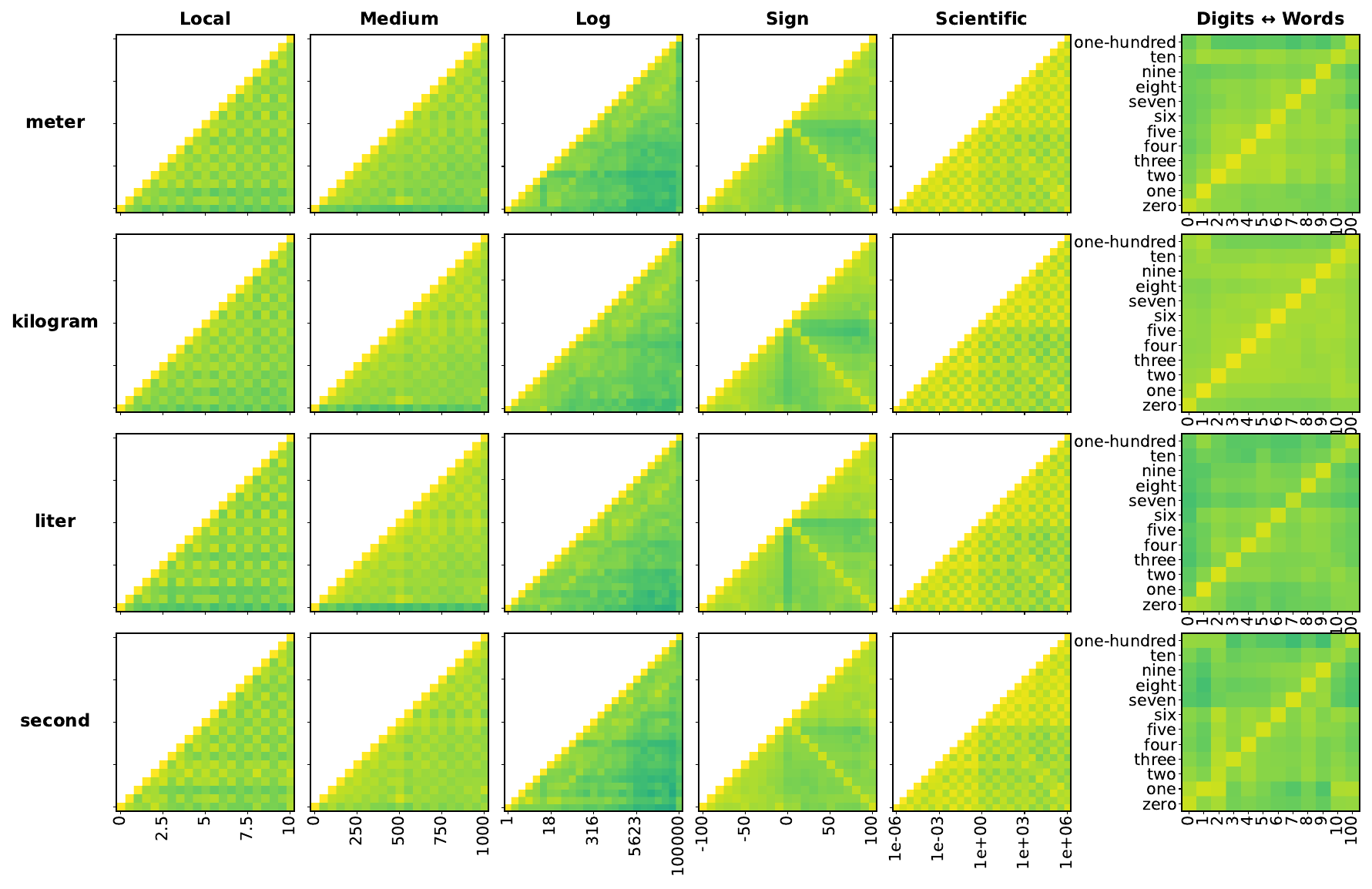}
    \caption{\texttt{harrier-oss-v1-0.6b}. For more information see caption in Figure~\ref{fig:overview-plot-mini-qwen}.}
    \label{fig:harrier_oss_v1_0.6b-eo}
\end{figure*}

\begin{figure*}
    \centering
    \includegraphics[width=1.0\linewidth]{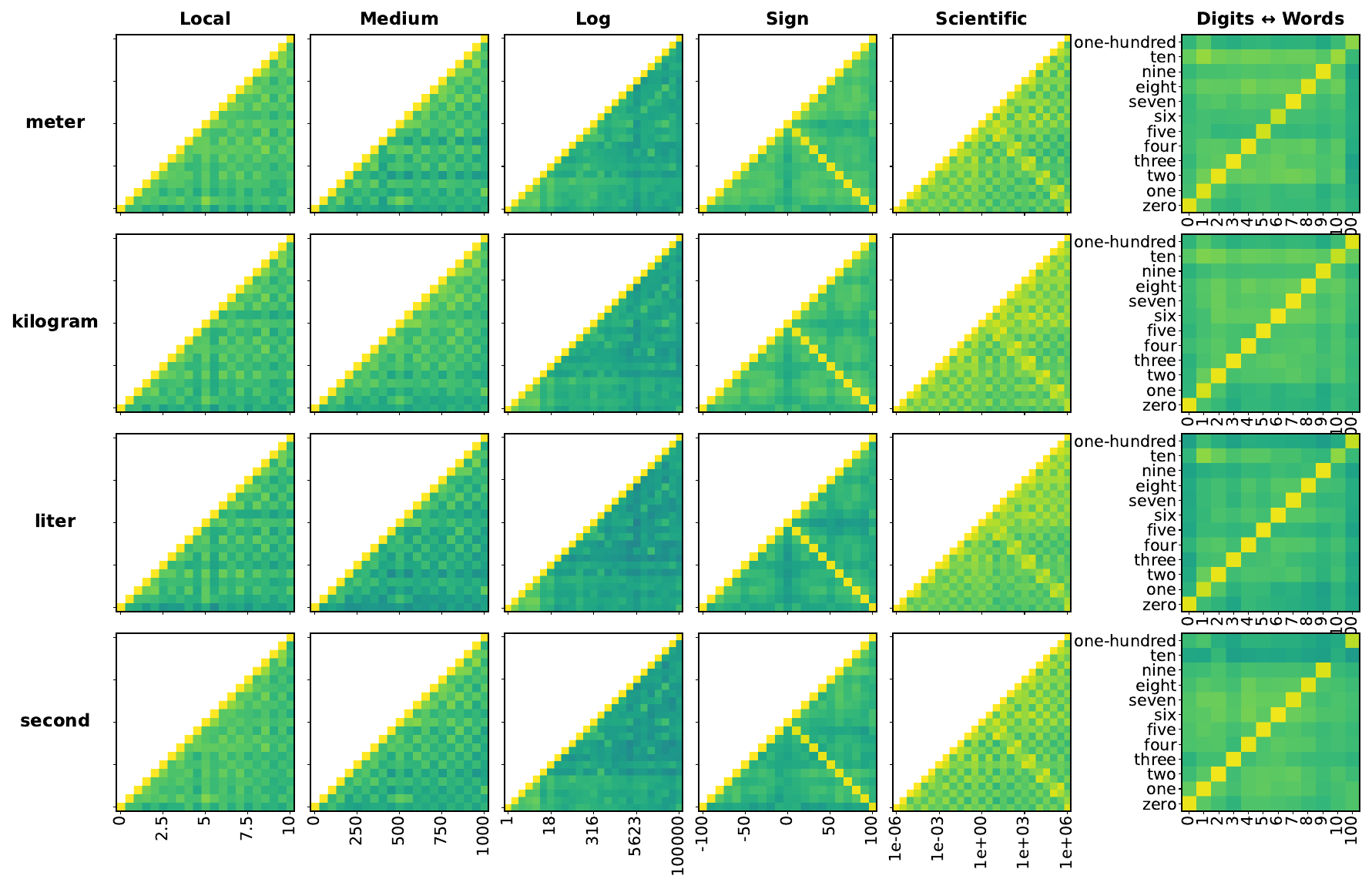}
    \caption{\texttt{embeddinggemma-300m}. For more information see caption in Figure~\ref{fig:overview-plot-mini-qwen}.}
    \label{fig:embeddinggemma_300m-eo}
\end{figure*}

\subsection{Conversion Plots}
\label{app:conversion}

\begin{figure}
    \centering
    \includegraphics[width=1.0\linewidth]{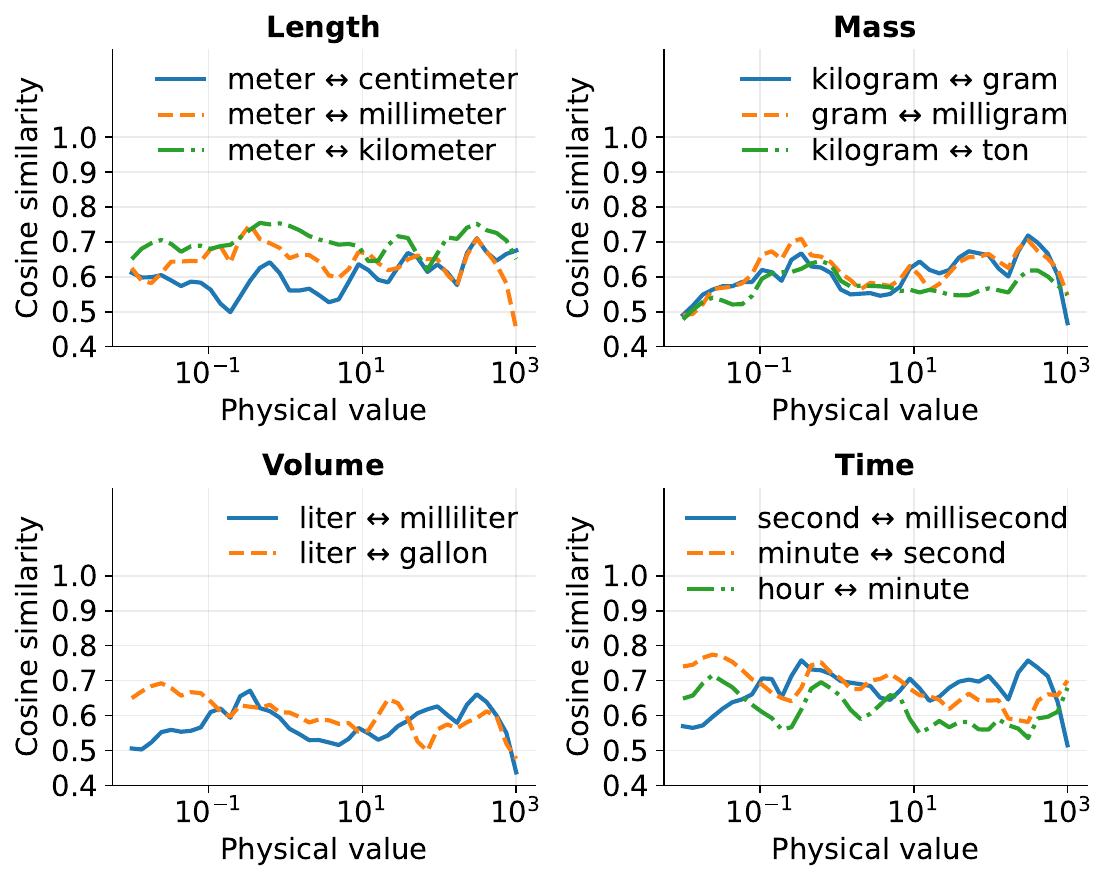}
    \caption{\texttt{all-MiniLM-L6-v2}. For more information see caption in Figure~\ref{fig:alignment}.}
    \label{fig:all_MiniLM_L6_v2-al}
\end{figure}

\begin{figure}
    \centering
    \includegraphics[width=1.0\linewidth]{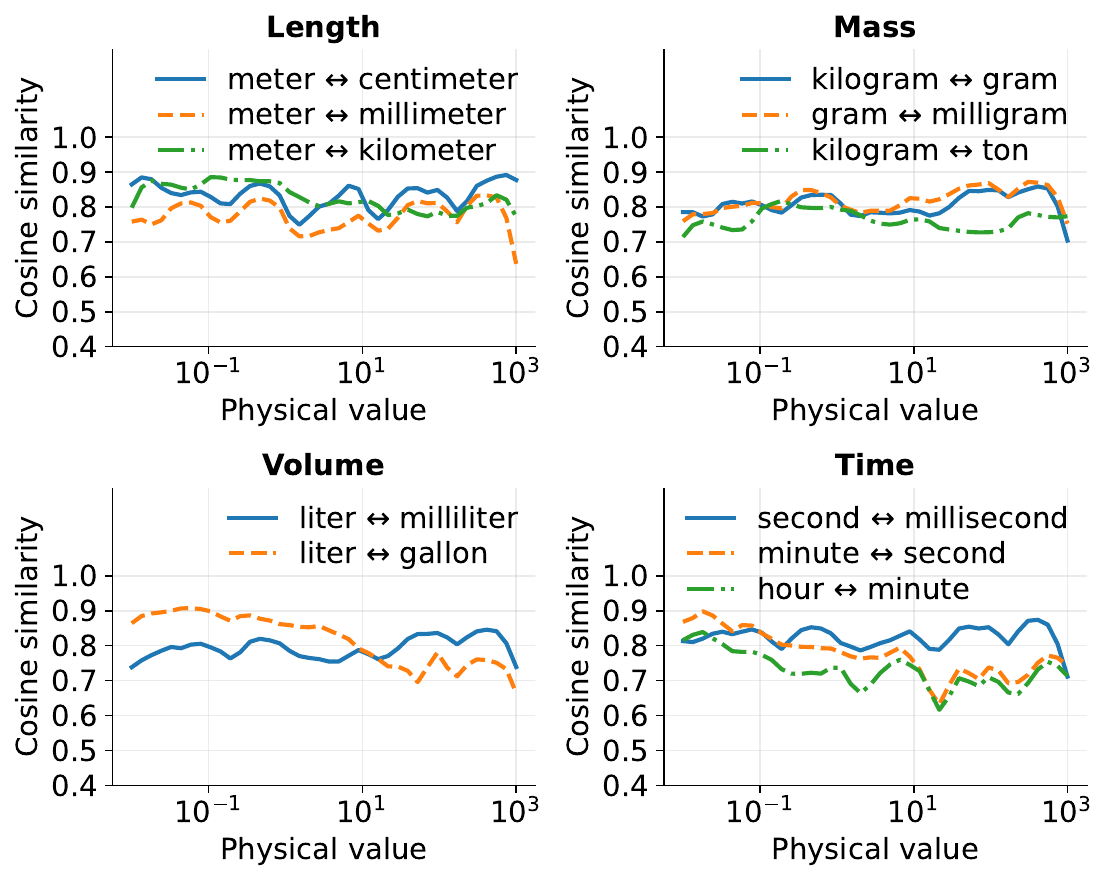}
    \caption{\texttt{paraphrase-multilingual-mpnet-base-v2}. For more information see caption in Figure~\ref{fig:alignment}.}
    \label{fig:paraphrase_multilingual_mpnet_base_v2-al}
\end{figure}

\begin{figure}
    \centering
    \includegraphics[width=1.0\linewidth]{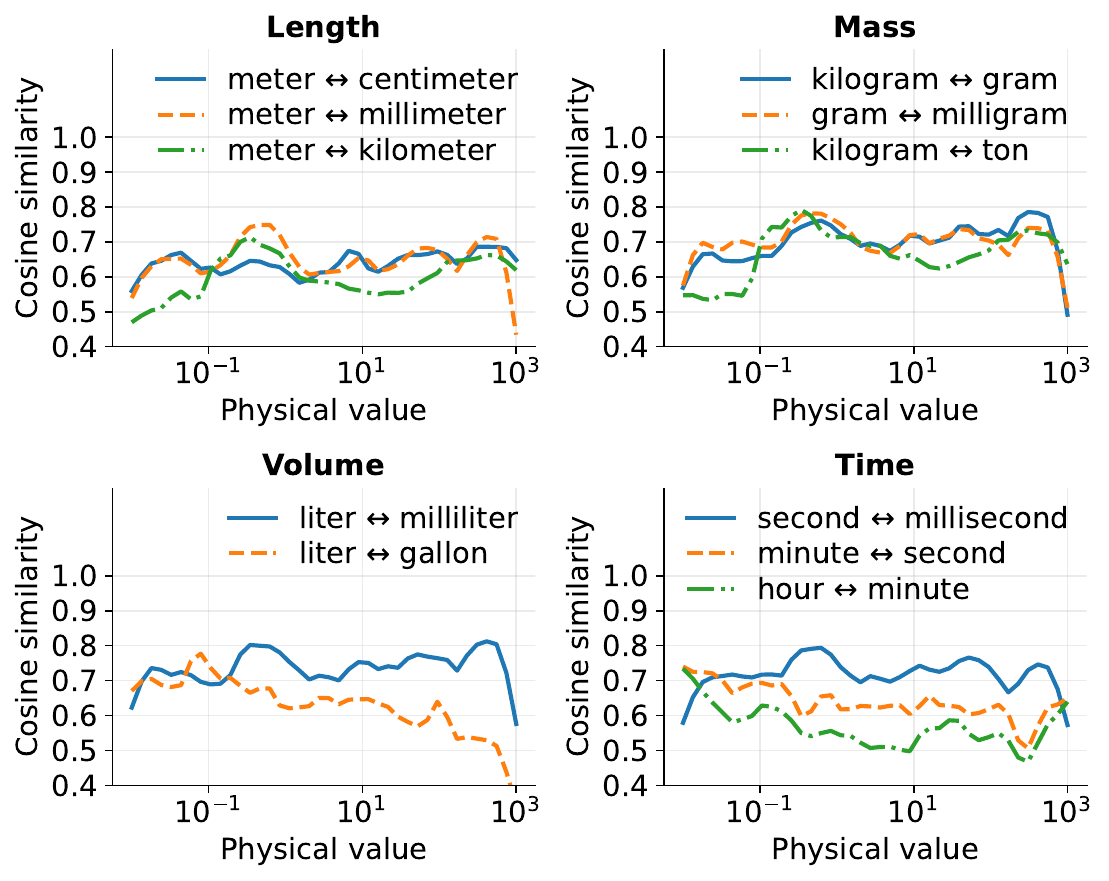}
    \caption{\texttt{LaBSE}. For more information see caption in Figure~\ref{fig:alignment}.}
    \label{fig:LaBSE-al}
\end{figure}

\begin{figure}
    \centering
    \includegraphics[width=1.0\linewidth]{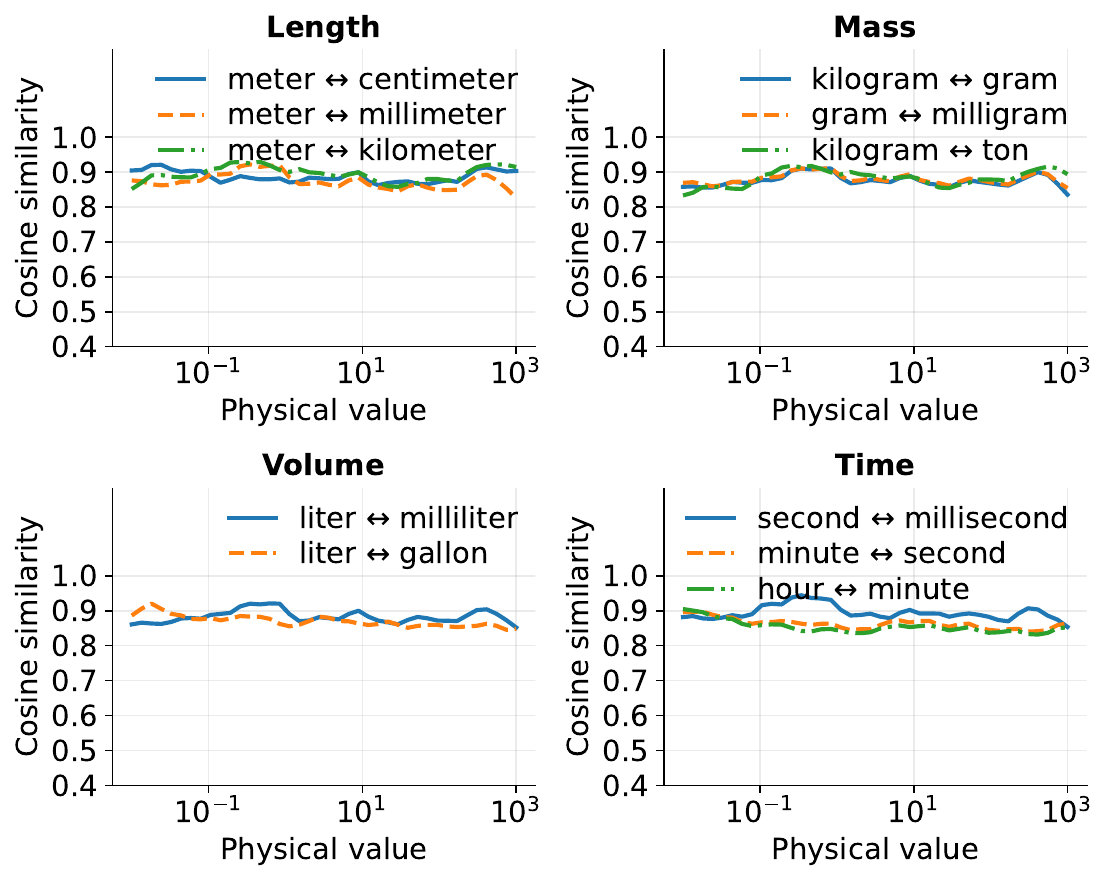}
    \caption{\texttt{e5-large-v2}. For more information see caption in Figure~\ref{fig:alignment}.}
    \label{fig:e5_large_v2-al}
\end{figure}

\begin{figure}
    \centering
    \includegraphics[width=1.0\linewidth]{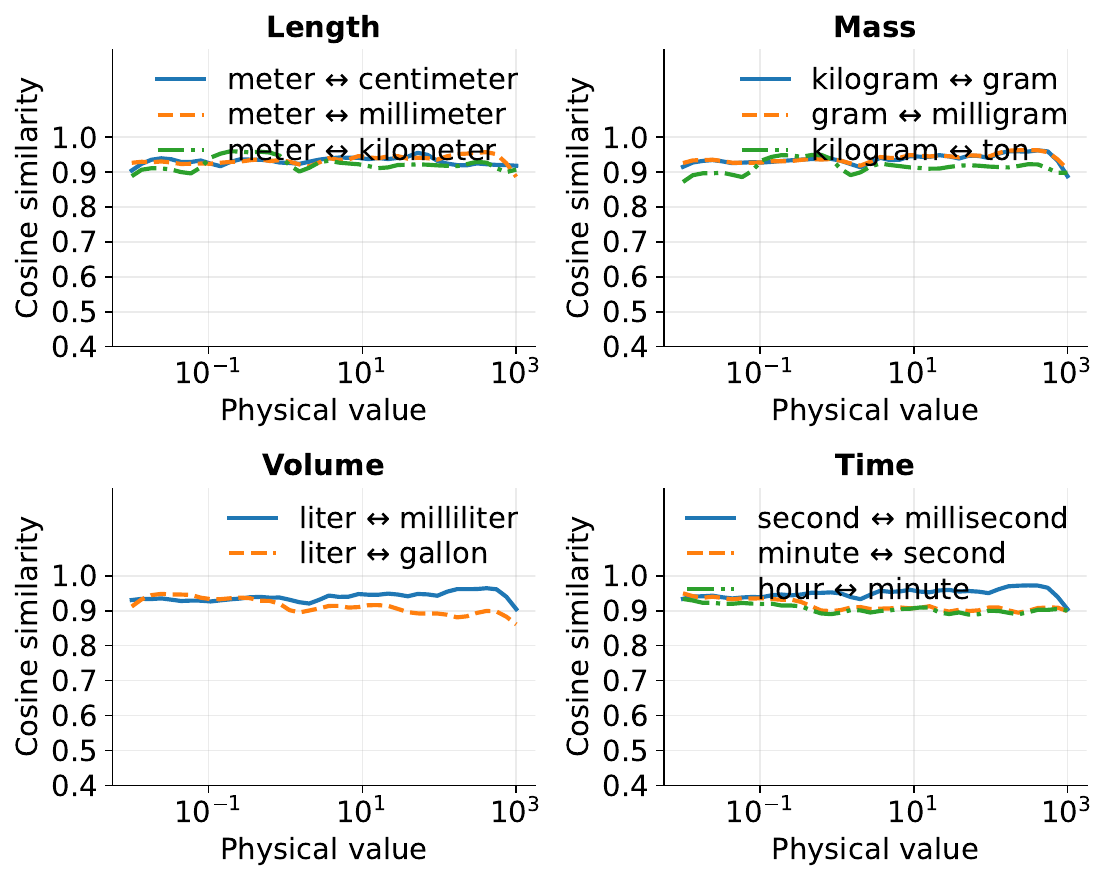}
    \caption{\texttt{multilingual-e5-base}. For more information see caption in Figure~\ref{fig:alignment}.}
    \label{fig:multilingual_e5_base-al}
\end{figure}

\begin{figure}
    \centering
    \includegraphics[width=1.0\linewidth]{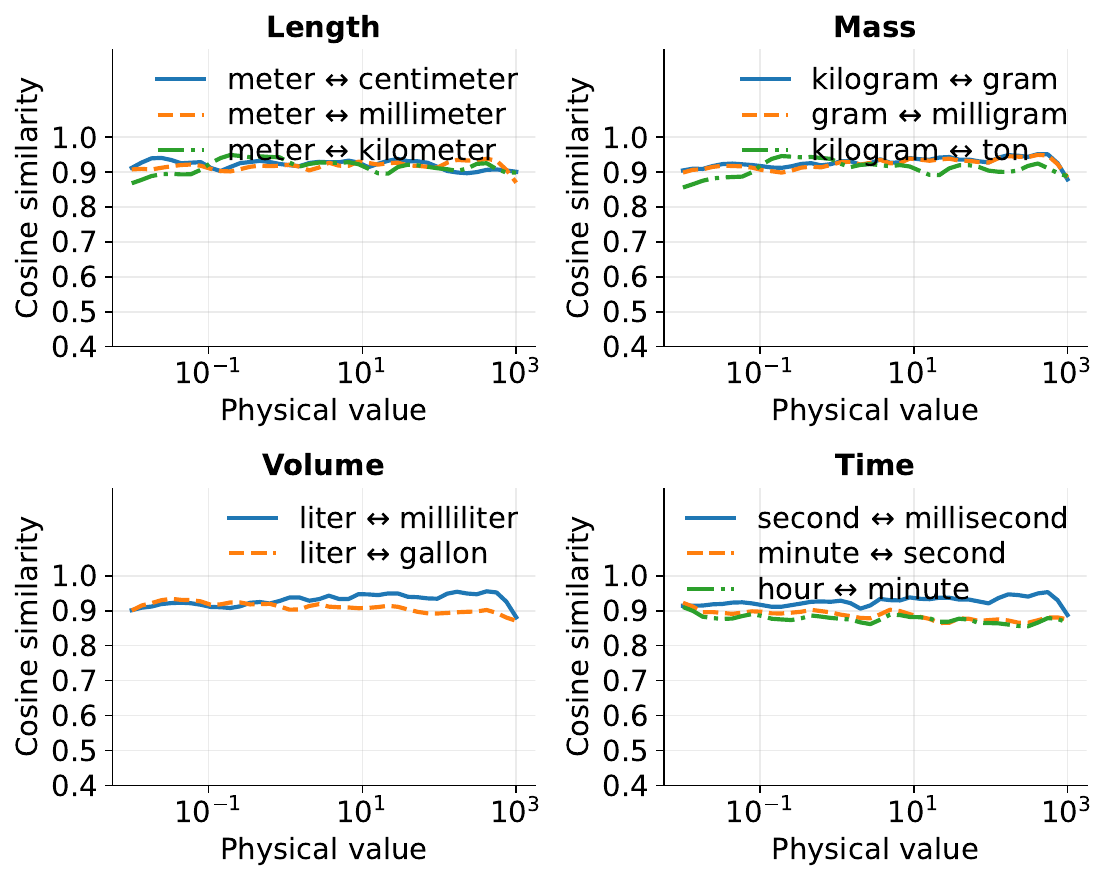}
    \caption{\texttt{multilingual-e5-large}. For more information see caption in Figure~\ref{fig:alignment}.}
    \label{fig:multilingual_e5_large-al}
\end{figure}

\begin{figure}
    \centering
    \includegraphics[width=1.0\linewidth]{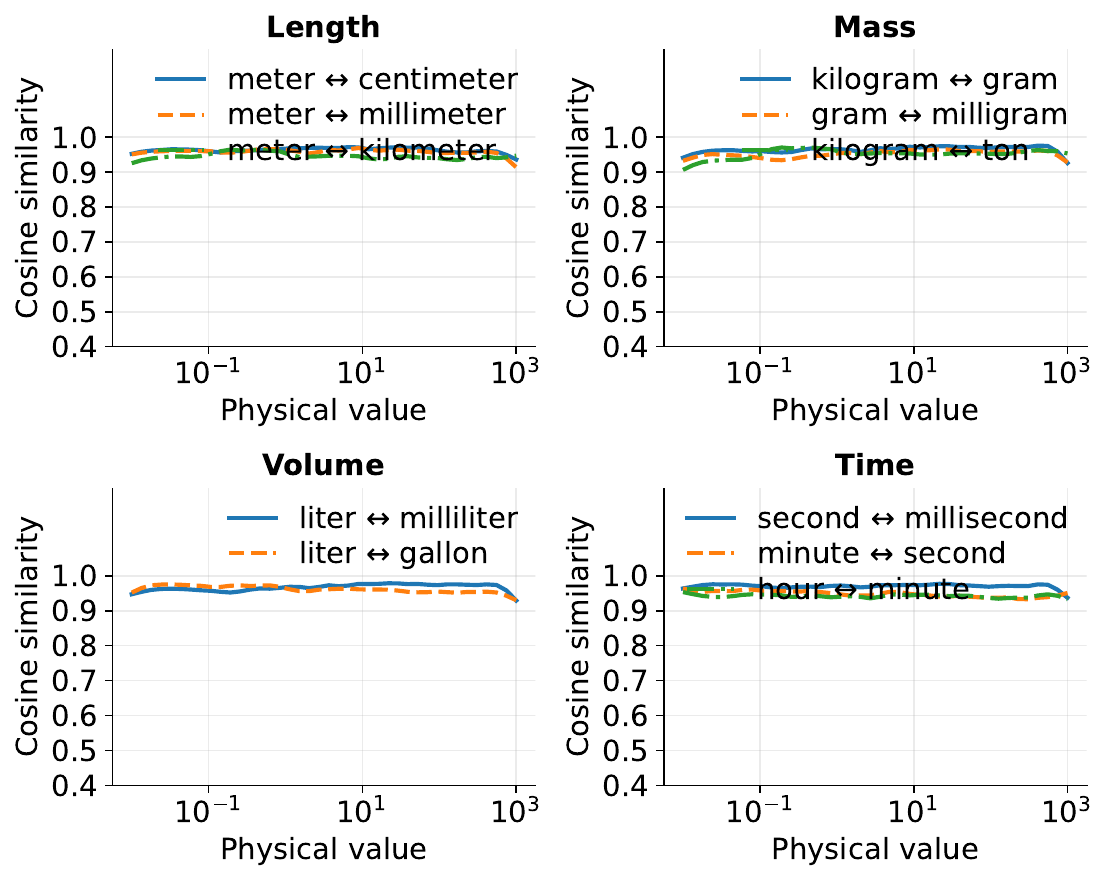}
    \caption{\texttt{multilingual-e5-large-instruct}. For more information see caption in Figure~\ref{fig:alignment}.}
    \label{fig:multilingual_e5_large_instruct-al}
\end{figure}

\begin{figure}
    \centering
    \includegraphics[width=1.0\linewidth]{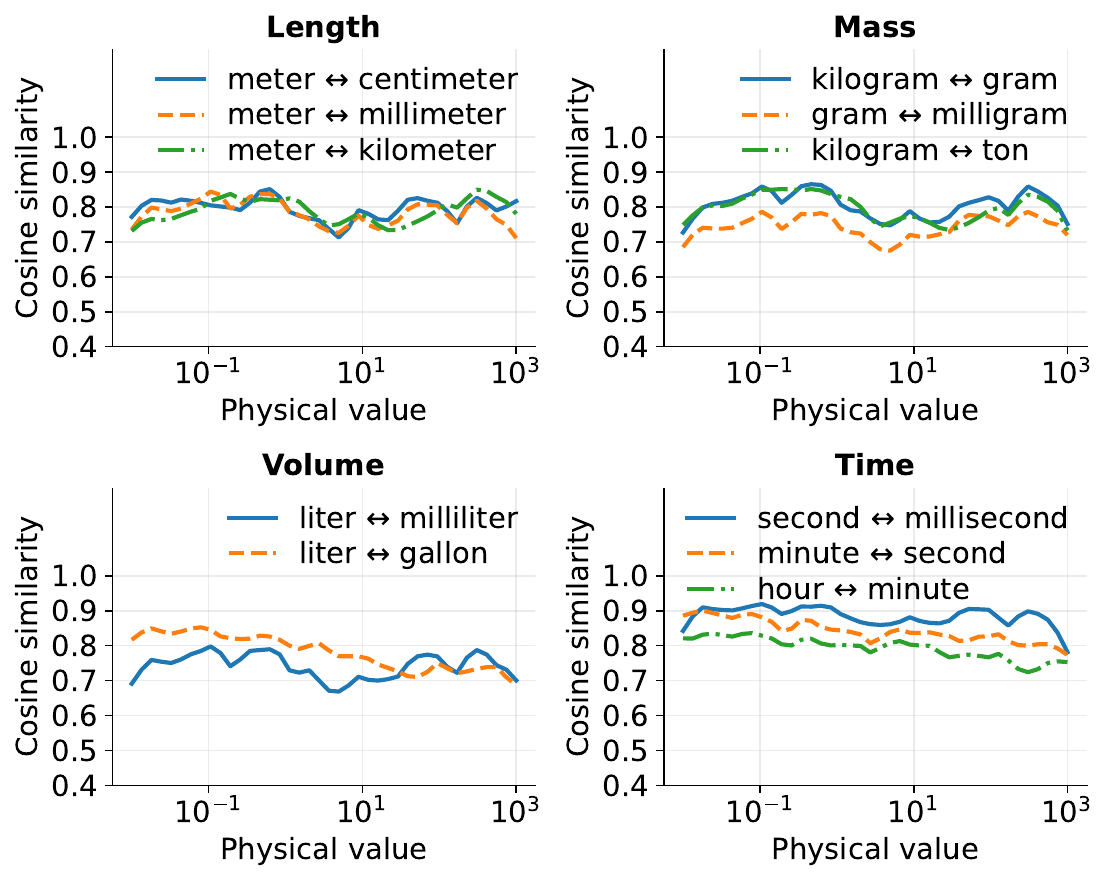}
    \caption{\texttt{bge-large-en-v1.5}. For more information see caption in Figure~\ref{fig:alignment}.}
    \label{fig:bge_large_en_v1.5-al}
\end{figure}

\begin{figure}
    \centering
    \includegraphics[width=1.0\linewidth]{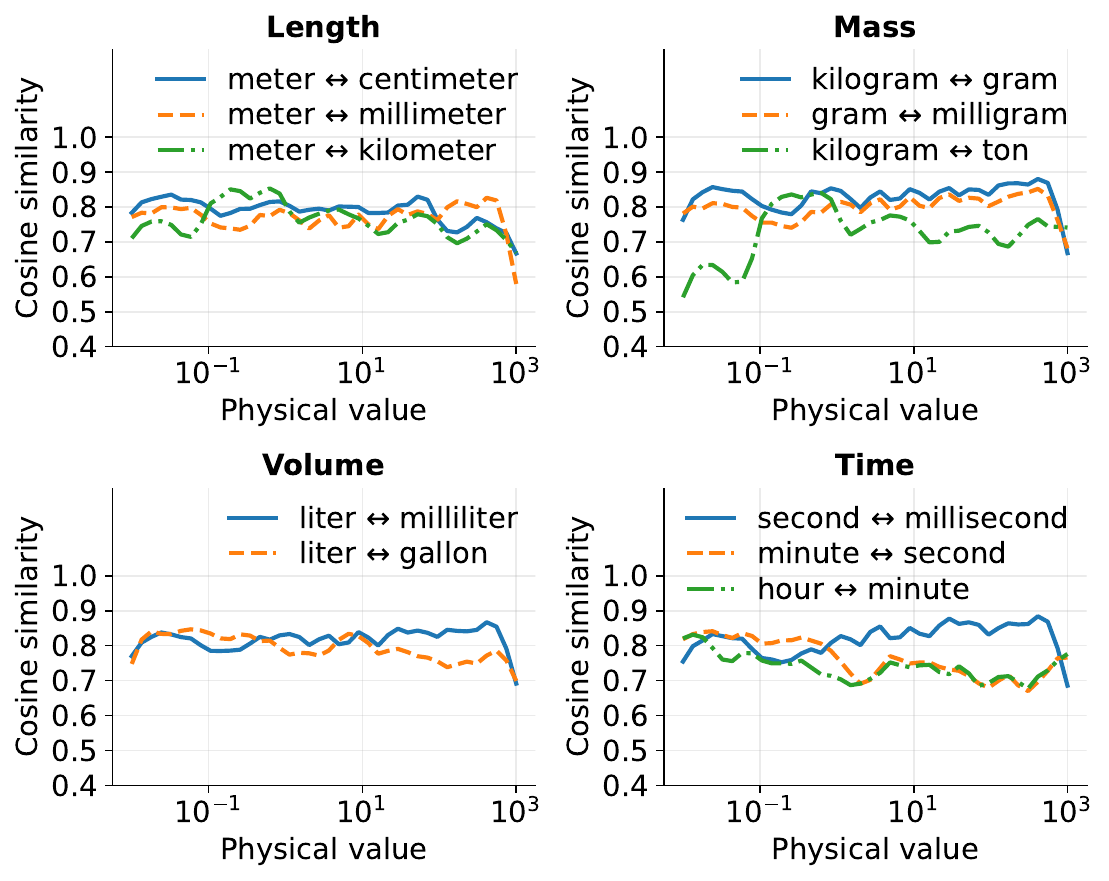}
    \caption{\texttt{bge-m3}. For more information see caption in Figure~\ref{fig:alignment}.}
    \label{fig:bge_m3-al}
\end{figure}

\begin{figure}
    \centering
    \includegraphics[width=1.0\linewidth]{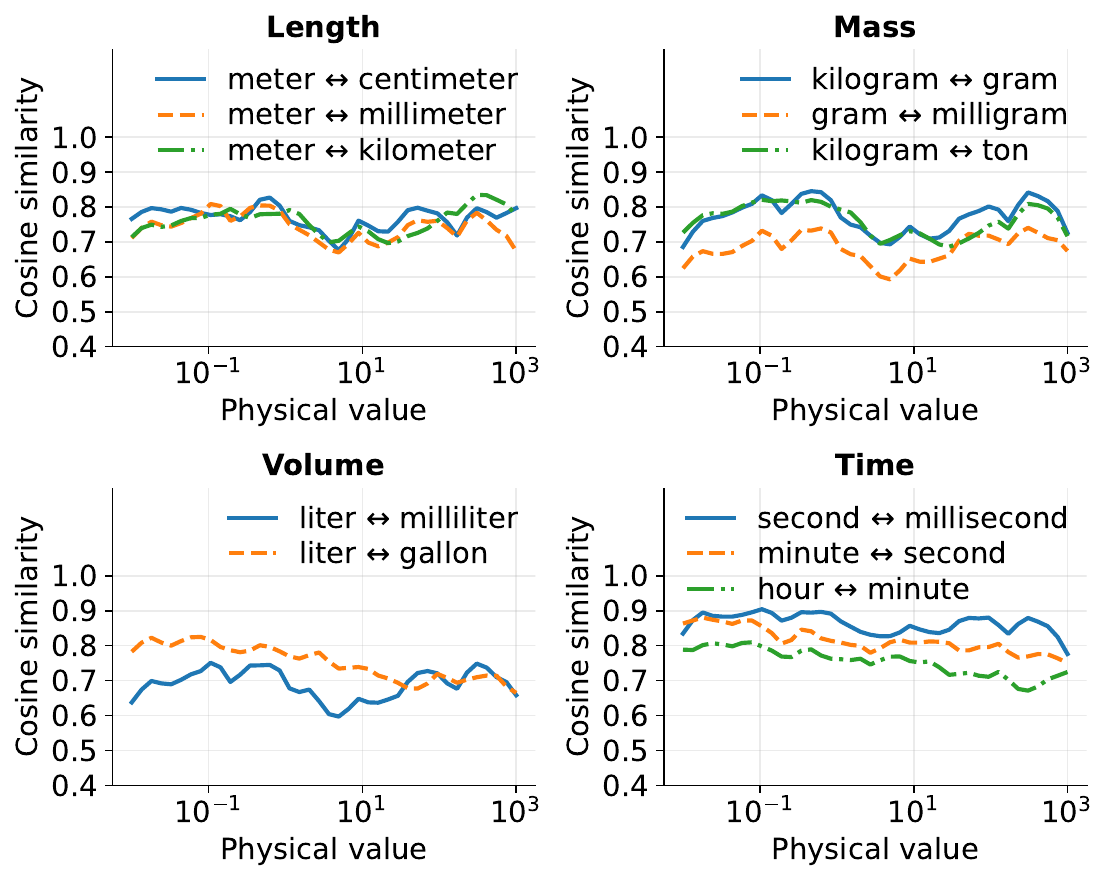}
    \caption{\texttt{mxbai-embed-large-v1}. For more information see caption in Figure~\ref{fig:alignment}.}
    \label{fig:mxbai_embed_large_v1-al}
\end{figure}

\begin{figure}
    \centering
    \includegraphics[width=1.0\linewidth]{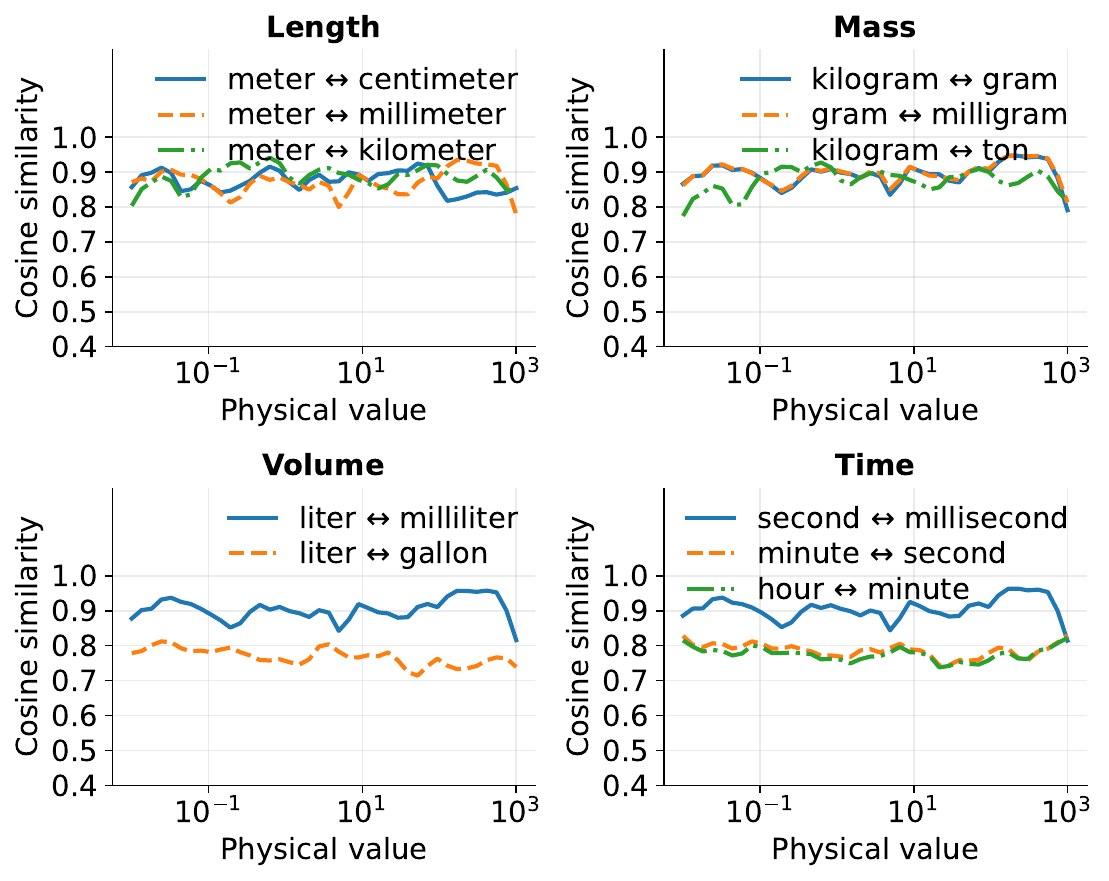}
    \caption{\texttt{granite-embedding-107m-multilingual}. For more information see caption in Figure~\ref{fig:alignment}.}
    \label{fig:granite_embedding_107m_multilingual-al}
\end{figure}

\begin{figure}
    \centering
    \includegraphics[width=1.0\linewidth]{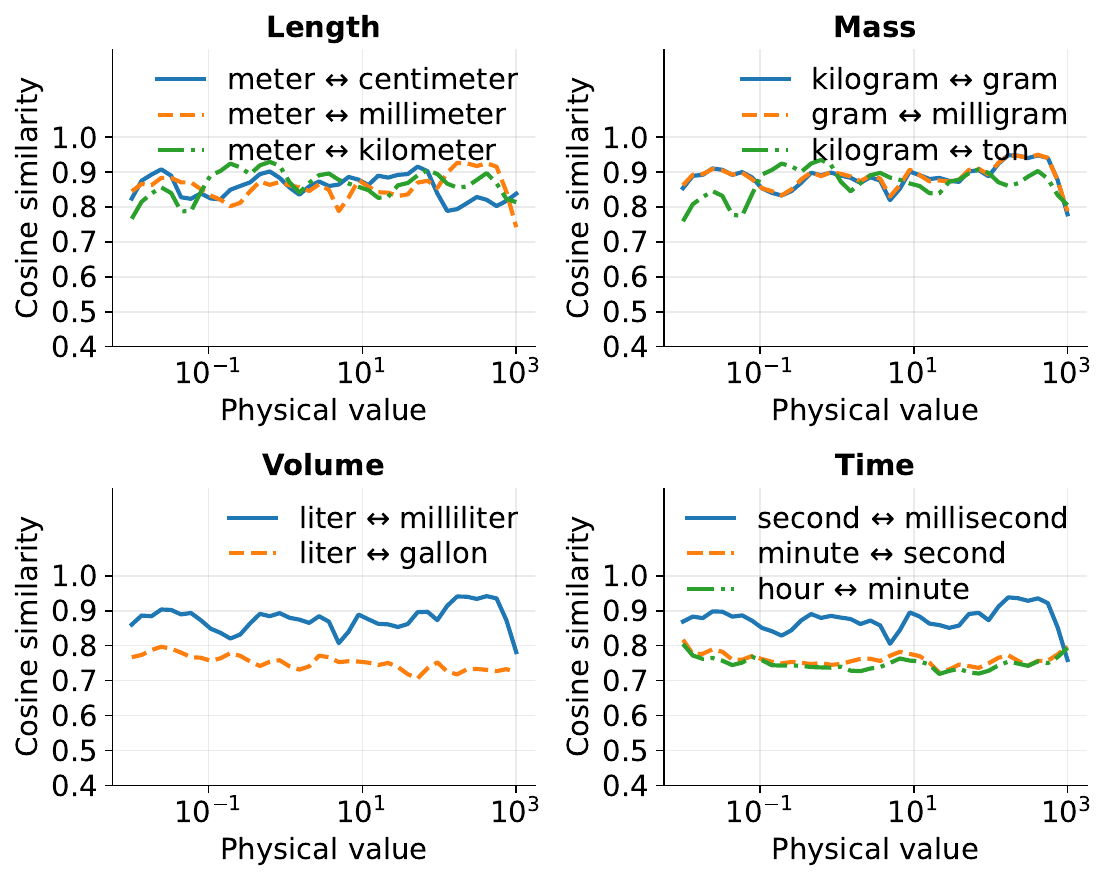}
    \caption{\texttt{granite-embedding-278m-multilingual}. For more information see caption in Figure~\ref{fig:alignment}.}
    \label{fig:granite_embedding_278m_multilingual-al}
\end{figure}

\begin{figure}
    \centering
    \includegraphics[width=1.0\linewidth]{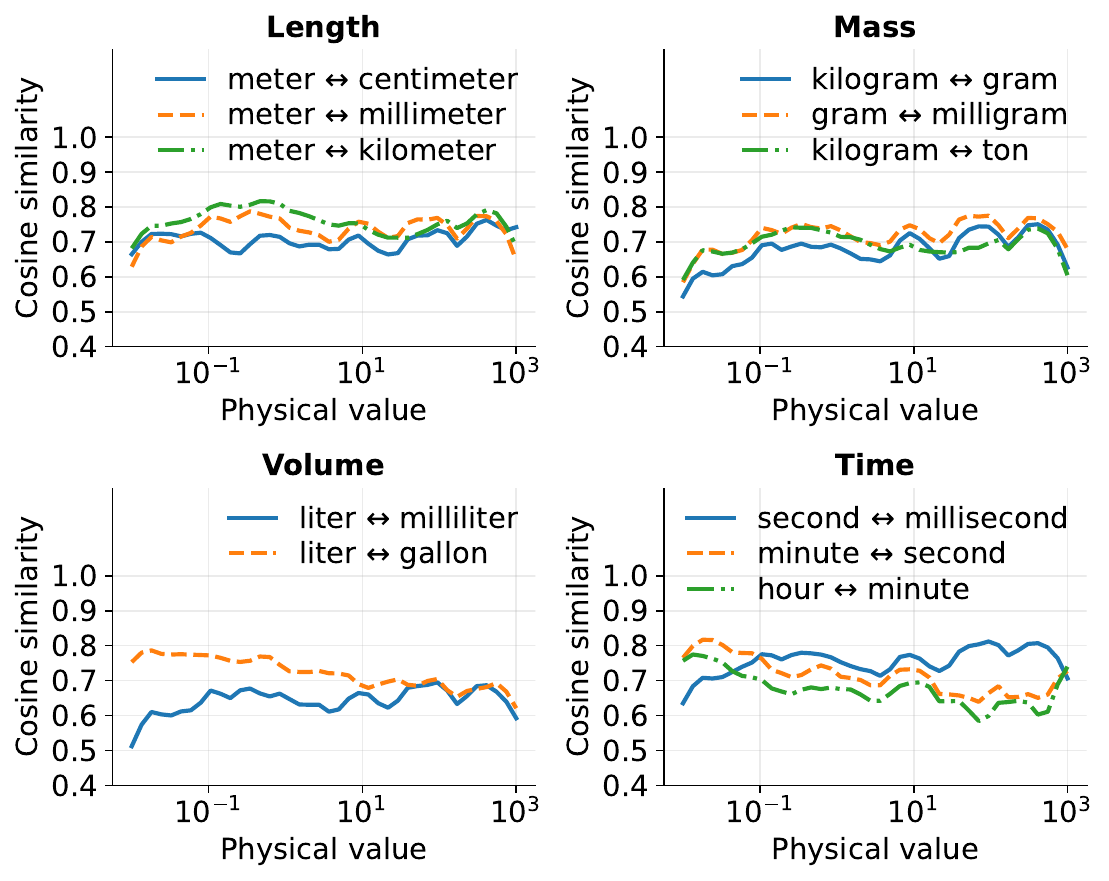}
    \caption{\texttt{nomic-embed-text-v1.5}. For more information see caption in Figure~\ref{fig:alignment}.}
    \label{fig:nomic_embed_text_v1.5-al}
\end{figure}

\begin{figure}
    \centering
    \includegraphics[width=1.0\linewidth]{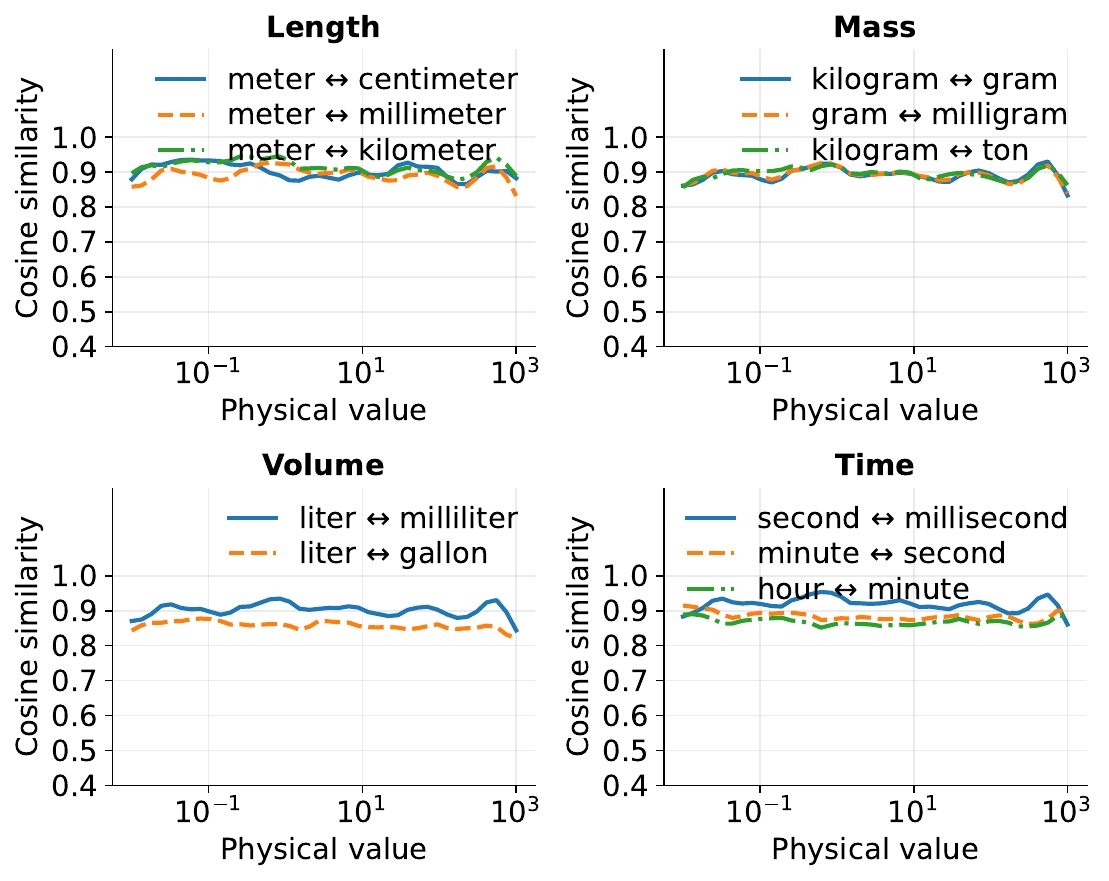}
    \caption{\texttt{granite-embedding-english-r2}. For more information see caption in Figure~\ref{fig:alignment}.}
    \label{fig:granite_embedding_english_r2-al}
\end{figure}

\begin{figure}
    \centering
    \includegraphics[width=1.0\linewidth]{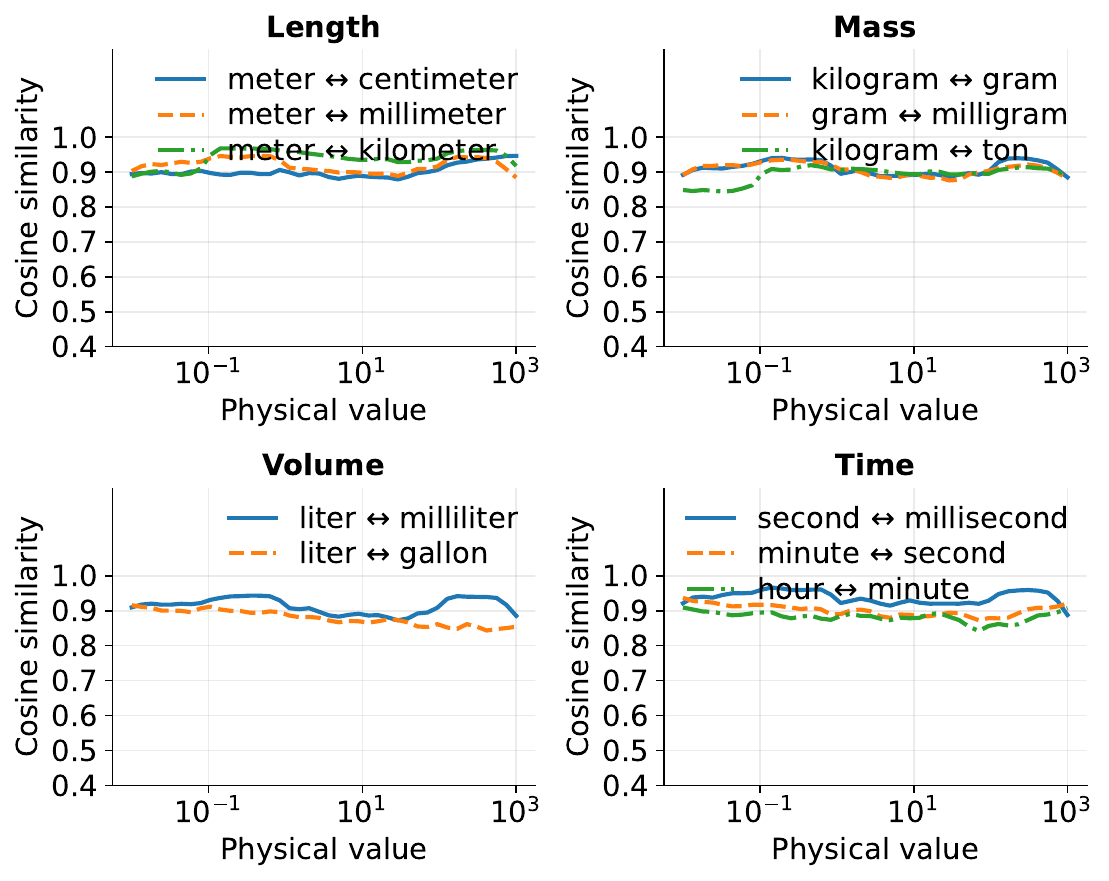}
    \caption{\texttt{granite-embedding-97m-multilingual-r2}. For more information see caption in Figure~\ref{fig:alignment}.}
    \label{fig:granite_embedding_97m_multilingual_r2-al}
\end{figure}

\begin{figure}
    \centering
    \includegraphics[width=1.0\linewidth]{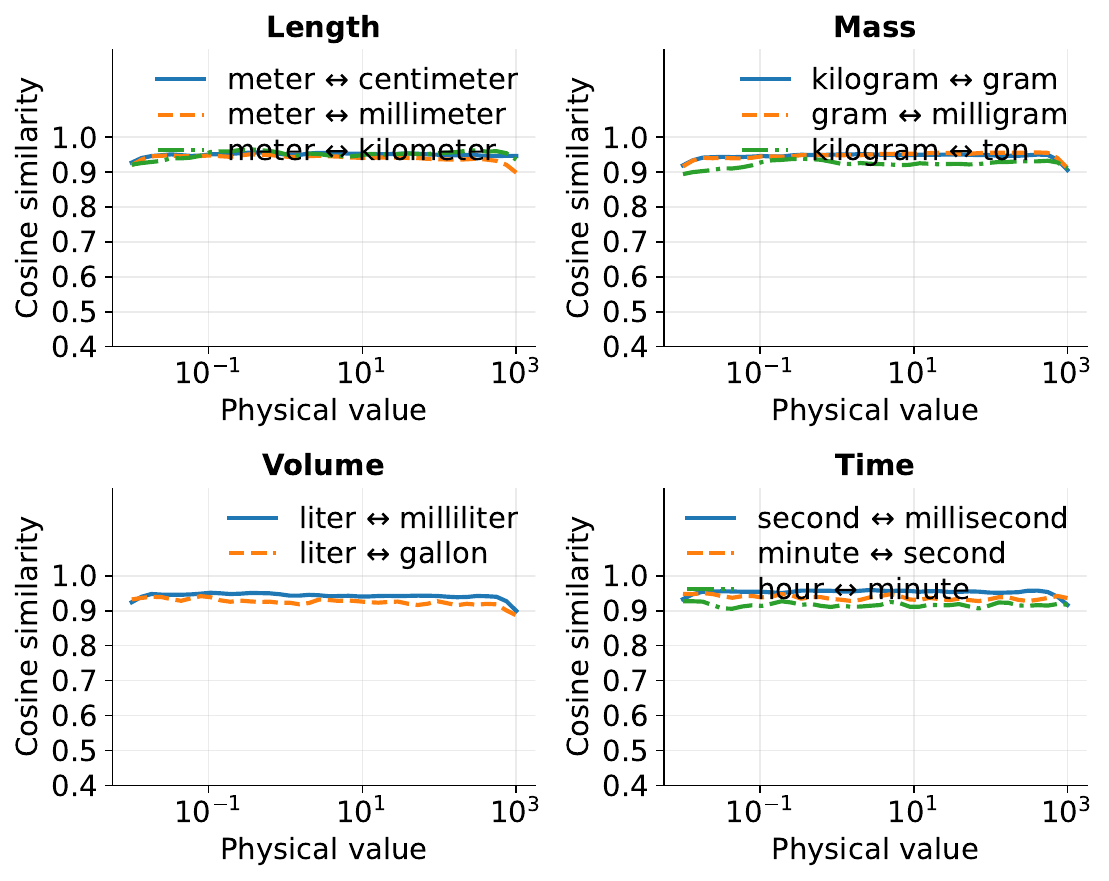}
    \caption{\texttt{granite-embedding-311m-multilingual-r2}. For more information see caption in Figure~\ref{fig:alignment}.}
    \label{fig:granite_embedding_311m_multilingual_r2-al}
\end{figure}

\begin{figure}
    \centering
    \includegraphics[width=1.0\linewidth]{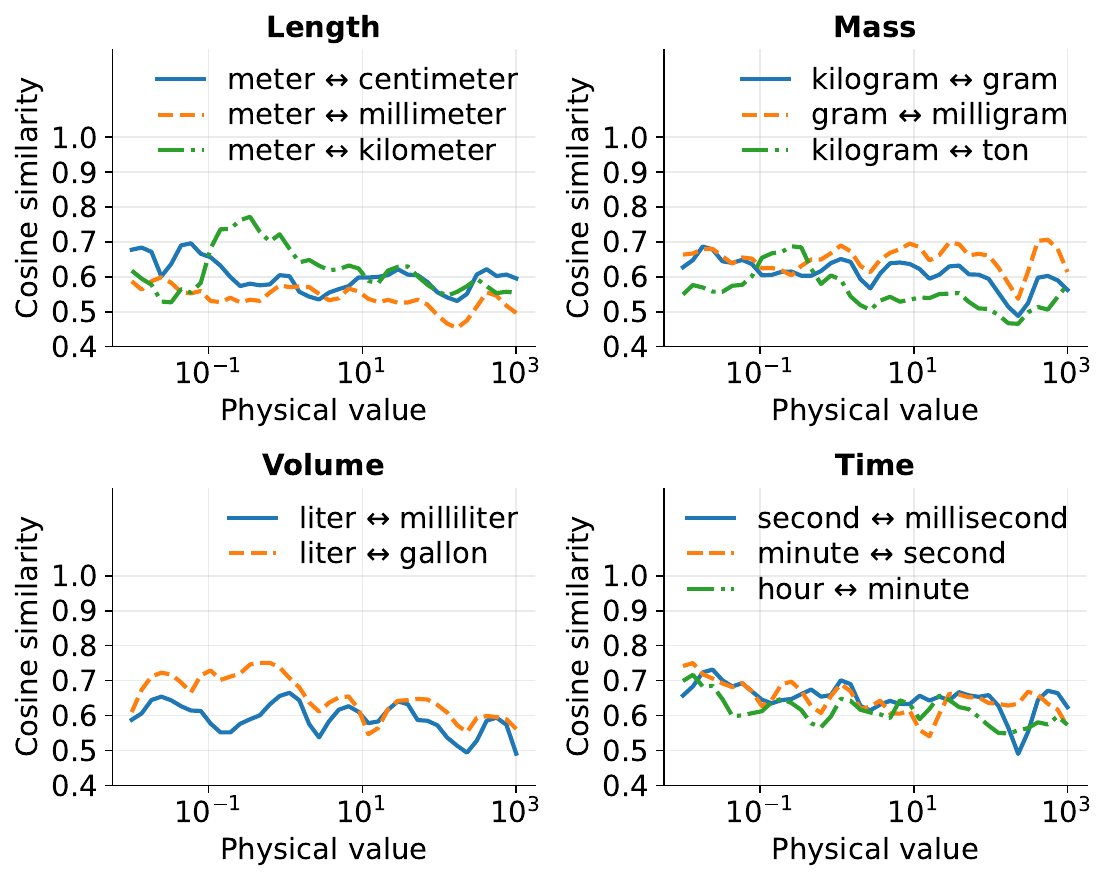}
    \caption{\texttt{DenseOn}. For more information see caption in Figure~\ref{fig:alignment}.}
    \label{fig:DenseOn-al}
\end{figure}

\begin{figure}
    \centering
    \includegraphics[width=1.0\linewidth]{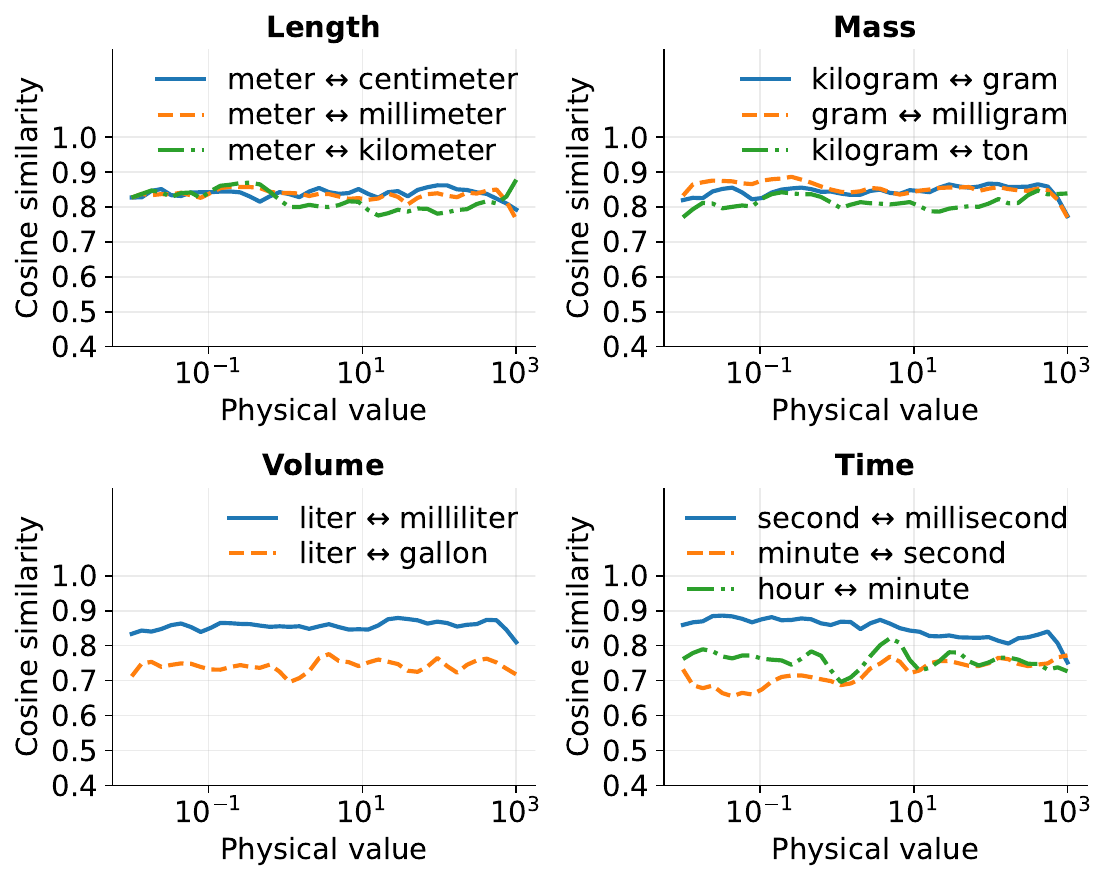}
    \caption{\texttt{Qwen3-Embedding-4B}. For more information see caption in Figure~\ref{fig:alignment}.}
    \label{fig:Qwen3_Embedding_4B-al}
\end{figure}

\begin{figure}
    \centering
    \includegraphics[width=1.0\linewidth]{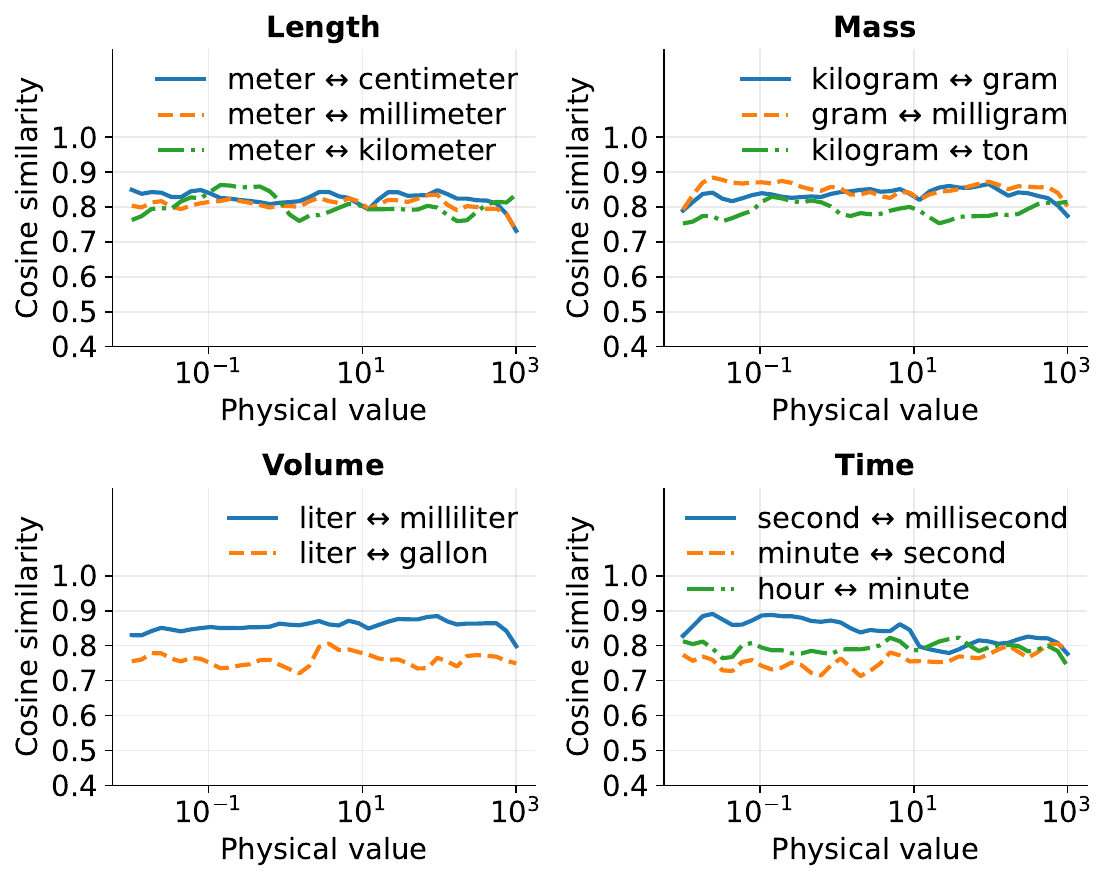}
    \caption{\texttt{Qwen3-Embedding-8B}. For more information see caption in Figure~\ref{fig:alignment}.}
    \label{fig:Qwen3_Embedding_8B-al}
\end{figure}

\begin{figure}
    \centering
    \includegraphics[width=1.0\linewidth]{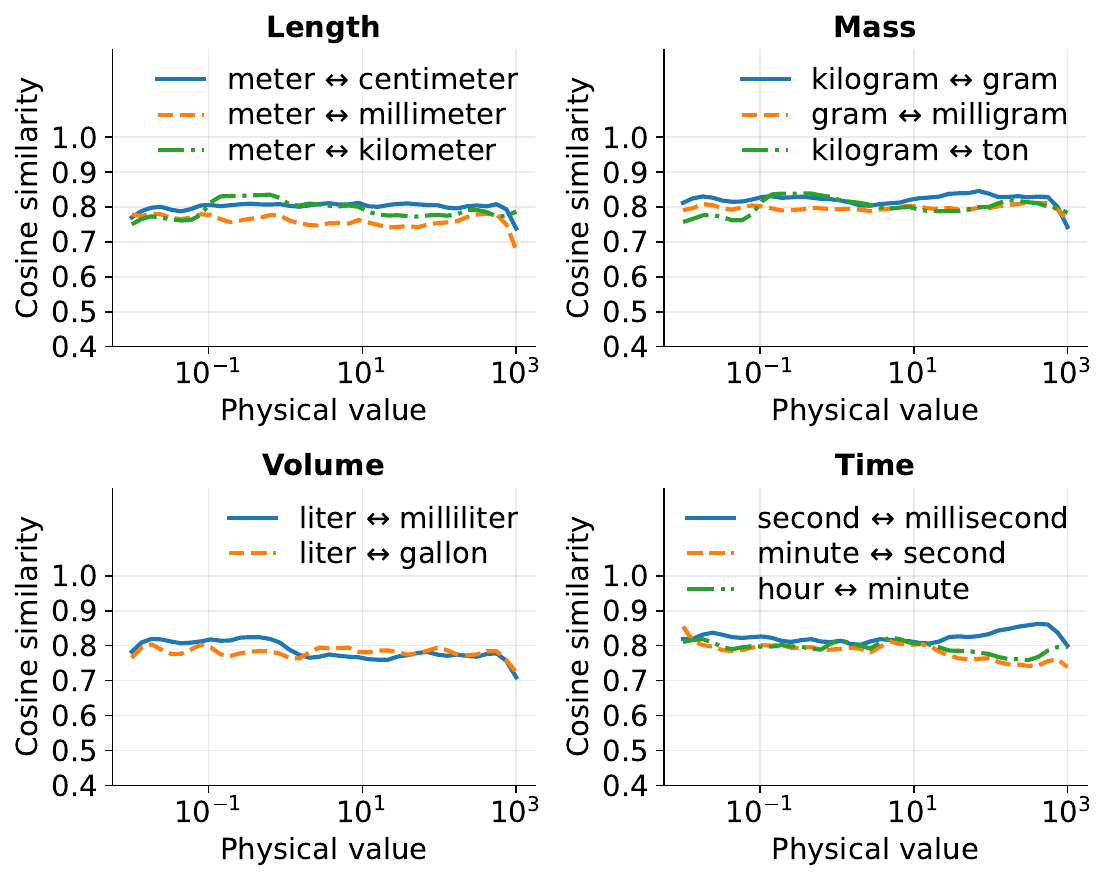}
    \caption{\texttt{harrier-oss-v1-270m}. For more information see caption in Figure~\ref{fig:alignment}.}
    \label{fig:harrier_oss_v1_270m-al}
\end{figure}

\begin{figure}
    \centering
    \includegraphics[width=1.0\linewidth]{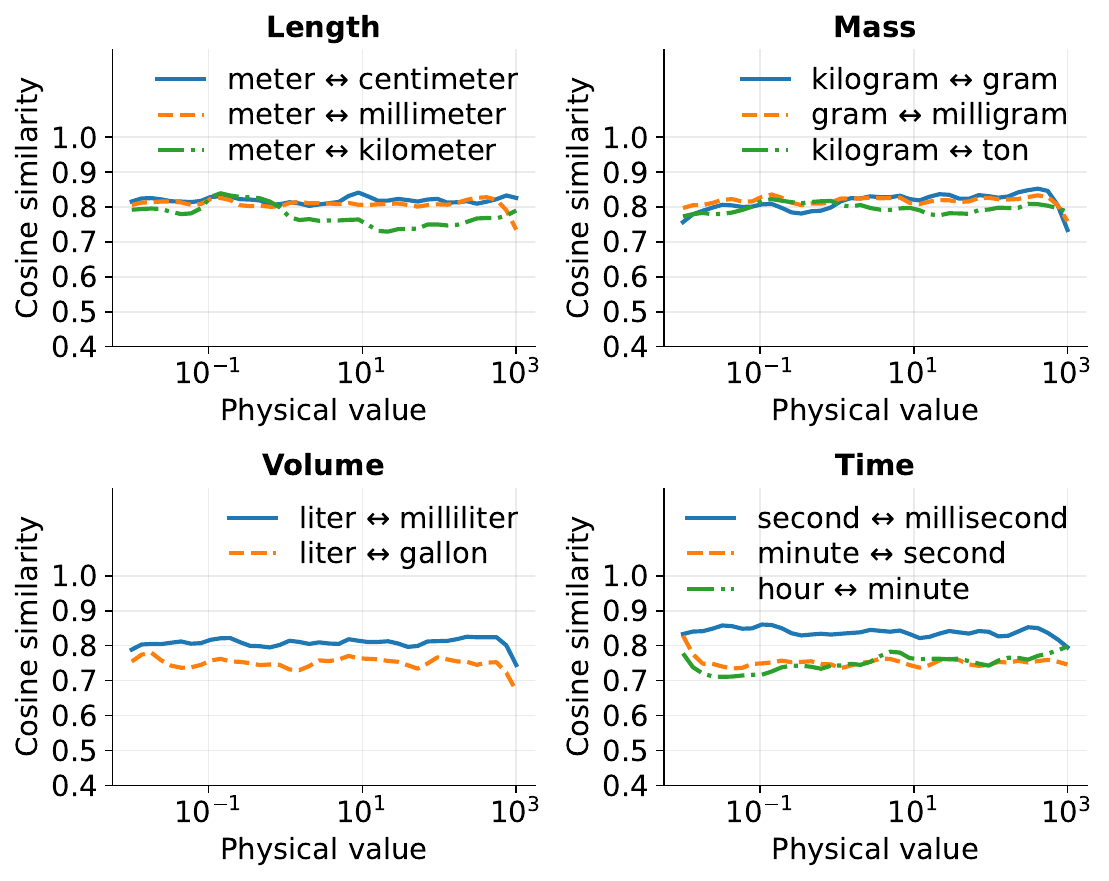}
    \caption{\texttt{harrier-oss-v1-0.6b}. For more information see caption in Figure~\ref{fig:alignment}.}
    \label{fig:harrier_oss_v1_0.6b-al}
\end{figure}

\begin{figure}
    \centering
    \includegraphics[width=1.0\linewidth]{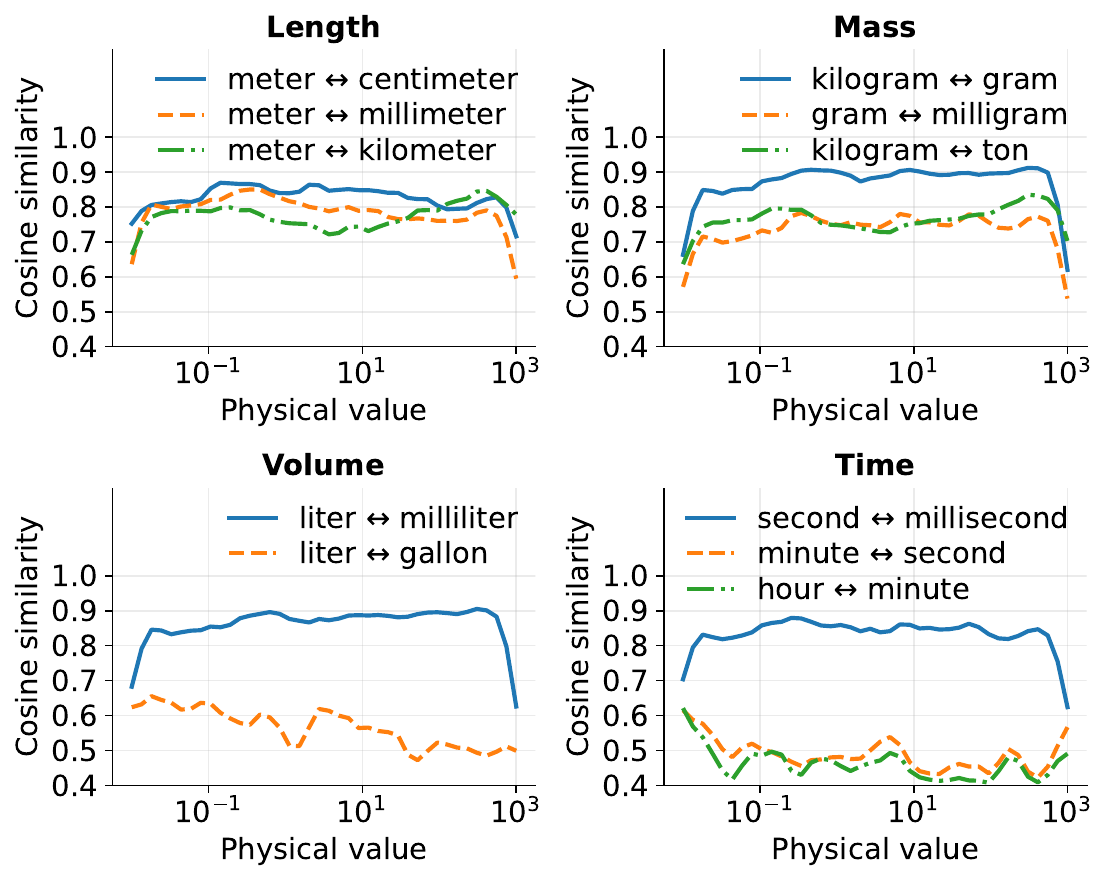}
    \caption{\texttt{embeddinggemma-300m}. For more information see caption in Figure~\ref{fig:alignment}.}
    \label{fig:embeddinggemma_300m-al}
\end{figure}

\end{document}